\documentclass[lettersize,journal]{IEEEtran}

\usepackage{times}
\usepackage{subcaption}
\usepackage{graphicx}
\usepackage{amsmath}
\usepackage{amssymb}
\usepackage{booktabs}
\usepackage{multirow}
\usepackage{arydshln}
\usepackage[super]{nth}
\usepackage[section]{placeins}
\usepackage[noadjust]{cite}
\usepackage{siunitx}
\usepackage[export]{adjustbox}
\usepackage[pagebackref=false,breaklinks=true,letterpaper=true,bookmarks=false]{hyperref}
\usepackage{textcomp}
\usepackage{stfloats}
\usepackage{url}
\usepackage{subcaption}
\usepackage{verbatim}
\def\BibTeX{{\rm B\kern-.05em{\sc i\kern-.025em b}\kern-.08em
    T\kern-.1667em\lower.7ex\hbox{E}\kern-.125emX}}
\usepackage{balance}

\begin{document}

\title{Soft-labelling for budget-constrained semantic segmentation: Bringing coherence to label down-sampling}

\author{Roberto Alcover-Couso,
Marcos Escudero-Viñolo,
Juan C. SanMiguel,
José M. Martinez\\
\thanks{Manuscript created September, 2023; This work has been funded by the SEGA-CV (TED2021-131643A-I00) and the HVD (PID2021-125051OB-I00) projects of the Ministerio de Ciencia e Innovación of the Spanish Government

Roberto Alcover-Couso, Marcos Escudero-Viñolo, Juan C. SanMiguel, José M. Martinez are with the Video Processing and Understanding Lab, Dept. Tecnología electrónica y de las comunicaciones, Escuela Polit\'{e}nica Superior, Universidad Aut\'{o}noma de Madrid, 28049 Madrid, Spain (email: roberto.alcover@uam.es, marcos.escudero@uam.es, juancarlos.sanmiguel@uam.es, josem.martinez@uam.es).}
}

%\markboth{IEEE Transactions on Image Processing}{Alcover-Couso Roberto, \MakeLowercase{\textit{(et al.)}}:Soft-labelling for budget-constrained semantic segmentation}
\maketitle

%%%%%%%%% ABSTRACT
\begin{abstract}

In semantic segmentation, training data down-sampling is commonly performed due to limited resources, the need to adapt image size to the model input, or to improve data augmentation. This down-sampling typically employs different strategies for the image data and the annotated labels. Such discrepancy leads to mismatches between the down-sampled colour and label images. Hence, the training performance significantly decreases as the down-sampling factor increases.
In this paper, we bring together the down-sampling strategies for the image data and the training labels. To that aim, we propose a novel framework for label down-sampling via soft-labelling that better conserves label information after down-sampling. Therefore, fully aligning soft-labels with image data to keep the distribution of the sampled pixels. This proposal also produces reliable annotations for under-represented semantic classes. Altogether,  it allows training competitive models at lower resolutions. Experiments show that the proposal outperforms other down-sampling strategies. Moreover, state-of-the-art performance is achieved for reference benchmarks, but employing significantly fewer computational resources than foremost methods. This proposal enables competitive research for semantic segmentation under resource constraints.

\end{abstract}
%%%%%%%%%%%%%%%%%%%%%%%%%%%%%%%%%%%%%%%%%%%%%%%%%%%%%%%%%%%%%%
\section{Introduction}
\IEEEPARstart{S}{emantic} segmentation is a core component for many core applications (e.g., autonomous driving \cite{Cordts2016Cityscapes} and medicine \cite{9126262}). Semantic segmentation methods assign a class label to each pixel in an image, being typically trained by supervised multi-class learning with training sets composed of colour images and pixel-level multi-class categorical labels.

\begin{figure}[t]
  \centering
  \includegraphics[width=\linewidth]{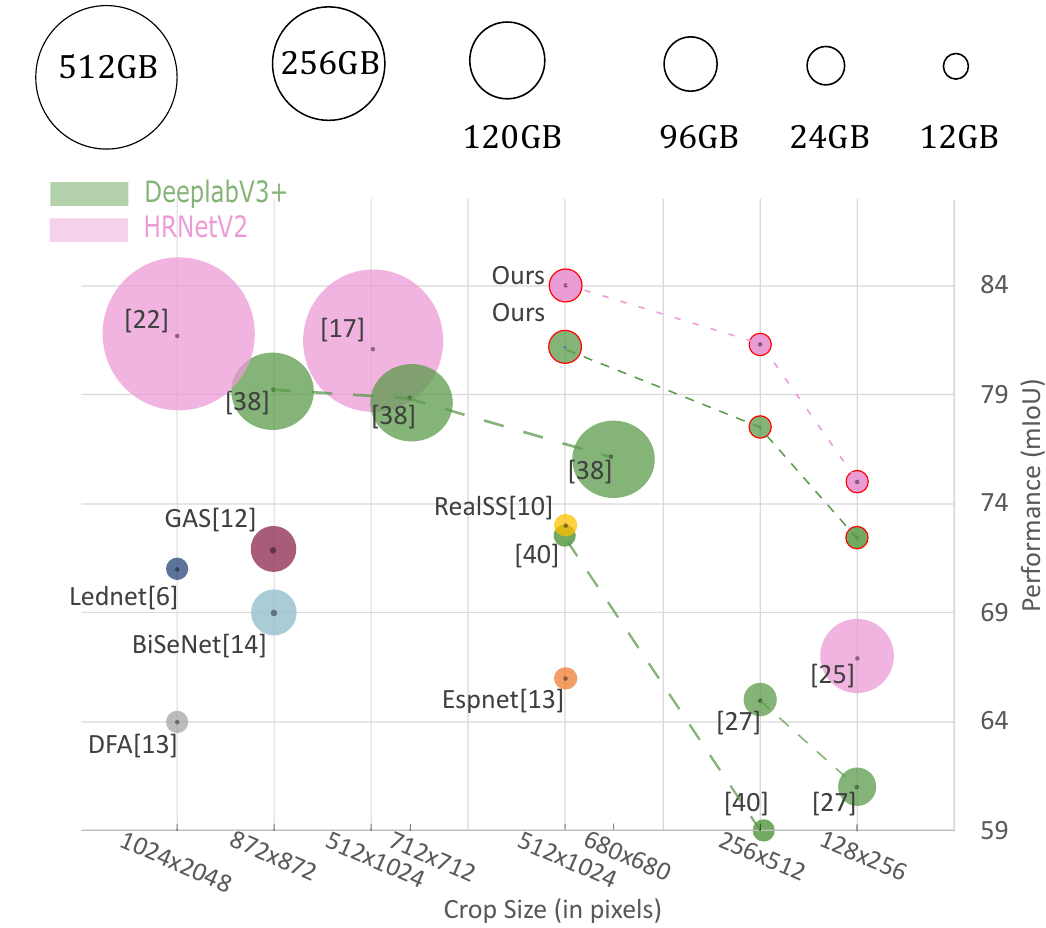}
  \caption{Graphical comparison of performance and resolution of prevalent models trained on the Cityscapes dataset \cite{Cordts2016Cityscapes}. The size of the circle represents the required GPU memory for training. Dashed lines connect equally-setup models trained at different resolutions. For Deeplab and HRNet we consider multiple frameworks which are referenced by the bibliography index. Alternative architectures are specified by their name.}
  %\vspace{-1 em}
  \label{fig:SOTA_COMP_CS}
\end{figure}

Recent semantic segmentation proposals have relied on ever-larger and varied datasets; and high-performance hardware such as Graphical-Processing-Units (GPUs). 
Hardware advances allowed training using full-resolution images with larger batch sizes \cite{NVIDIA_semseg}, and employing resource-expensive regional losses \cite{2019_zhao_rmi}. 
In fact, the former has been shown to be a binding factor in achieving state-of-the-art performance (e.g., the Cityscapes state-of-the-art top-performing HRNetV2 \cite{Wang_2023_CVPR} in Figure \ref{fig:SOTA_COMP_CS} uses a batch size of 16 images for training). However, training with full-resolution images requires high computational and storage resources \cite{8803154}. 
This hindrance creates a huge barrier for budget-constrained research, thus limiting the development of competitive proposals to the availability of large computation facilities. For example, a high-end 32GB GPU is required to train a recent multi-scale method with a batch size of one full-resolution image \cite{NVIDIA_semseg} from the Cityscapes benchmark \cite{Cordts2016Cityscapes}. It is commonly agreed that larger batch sizes are crucial to increase semantic segmentation performance \cite{zhou2019semantic} (e.g., improvements over 300\% are obtained by increasing batch sizes from 2 to 8 \cite{liang2019winter}). The setups for top-performing methods range from 4 \cite{li2022deep} to 50 \cite{deeplabv3plus2018} high-end GPUs, keeping away a large part of the scientific community from competitive research on the topic. 

To operate in resource-limited setups, lightweight semantic segmentation proposals have been proposed to counterbalance the resource requirements by reducing the number of parameters. This enables the use of mid-range GPUs, generally at the expense of performance. Examples of these methods are Lednet \cite{8803154}, DFA \cite{Li2019DFANetDF}, GAS \cite{Lin2020GraphGuidedAS}, Espnet \cite{mehta2018espnetv2}, BiSeNet \cite{Yu_2018_ECCV} and RealSS \cite{hong2021deep}, whose trade-off between performance and resources is depicted in Figure \ref{fig:SOTA_COMP_CS}. However, large batch sizes are still employed to obtain good performance, therefore multiple GPUs are mandatory for their training \cite{Lin2020GraphGuidedAS, xu2022pidnet}.

Moreover, top-tier semantic segmentation methods rely on image resizing data augmentation \cite{li2022deep, zhou2022rethinking}, as employing such techniques typically improves performance due to the small number of images found on semantic segmentation datasets \cite{Cordts2016Cityscapes}. Therefore, re-scaling the input data (i.e., up-sampling or down-sampling images and labels) is transversal to almost every modern semantic segmentation training framework. 
In this regard, an often ignored phenomenon is the \textit{noise} and artefacts introduced by the down-sampling strategy employed for re-scaling. Specifically, a critical issue is the misalignment between the down-sampled colour and label images. On one hand, smooth down-sampling estimates are preferred for the colour images over majority-filter-like strategies such as the Nearest Neighbour (NN), as the latter is prone to create blocking artefacts, jagged edges, and thin structures removals \cite{NN}. On the other hand, Nearest Neighbour is the default down-sampling strategy applied to label images to maintain their integer nature, as labels are categorical values that do not define a metric space.
 %Our proposal manages to obtain results on part with the state-of-the-art with significantly less resources.  

In this paper, we introduce an innovative framework designed for label down-sampling through a soft-labelling approach, which enhances the preservation of label information during down-sampling. Consequently, our approach effectively aligns soft-labels with image data, ensuring the consistent distribution of the sampled pixels. Our best model obtains better performance using down-sampled versions of the trained images to that obtained by state-of-the-art methods using full-resolution images (see models labelled as \textit{Ours} in Figure \ref{fig:SOTA_COMP_CS}), and requiring significantly less computational resources. To our knowledge, we are the first to study this approach. In practice, compared to the model using most resources\textemdash i.e., DeepLabV3+ on half resolution that employs 50 GPUs of 24 GB each to be trained for the Cityscapes dataset \cite{deeplabv3plus2018}, our framework using the same architecture trained using a single Titan RTX GPU is able to outperform this model by a 3\% using $\frac{1}{4}$ of the resolution and requiring less than 12GB of memory, and by a 7\% using the same resolution, less than 24GB of memory, and a batch size of only 3 images.

The contributions of this paper are three-fold: First, we introduce the concept of soft-labels for semantic segmentation, enabling the use of a flexible down-sampling strategy for label images. This approach allows a paired sampling procedure with colour images, thus reducing the noise introduced by the sampling miss-match in semantic segmentation. Second, we demonstrate the advantages of paired sampling by analysing various sampling pairs. In our experiments, we found that the training of our models remains stable even on small batch sizes. This is relevant, as large batch sizes are regarded as necessary for training semantic segmentation models. Third, we present extensive analyses and comparatives with the state-of-the-art methods on three popular semantic segmentation datasets. Our approach not only improves overall performance but also requires significantly fewer resources.

The rest of the paper is organised as follows: First, we review the related work on semantic segmentation, image down-sampling and soft-labelling in Section \ref{sec:RW}. Second, we present our proposed soft-labelling framework for semantic segmentation in Section \ref{sec:SSR}. Third, we showcase experimental results with extensive ablation studies for the Cityscapes \cite{Cordts2016Cityscapes} dataset in Section \ref{sec:EX}. We also perform experiments on other semantic segmentation datasets. Finally,  we conclude the paper and provide directions for future research in Section \ref{sec:CO}.

%%%%%%%%%%%%%%%%%%%%%%%%%%%%%%%%%%%%%%%%%%%%%%%%%%%%%%%%%%%%%%
\section{Related work}
\label{sec:RW}
\begin{figure}[t]
    \centering
    \includegraphics[width=\linewidth]{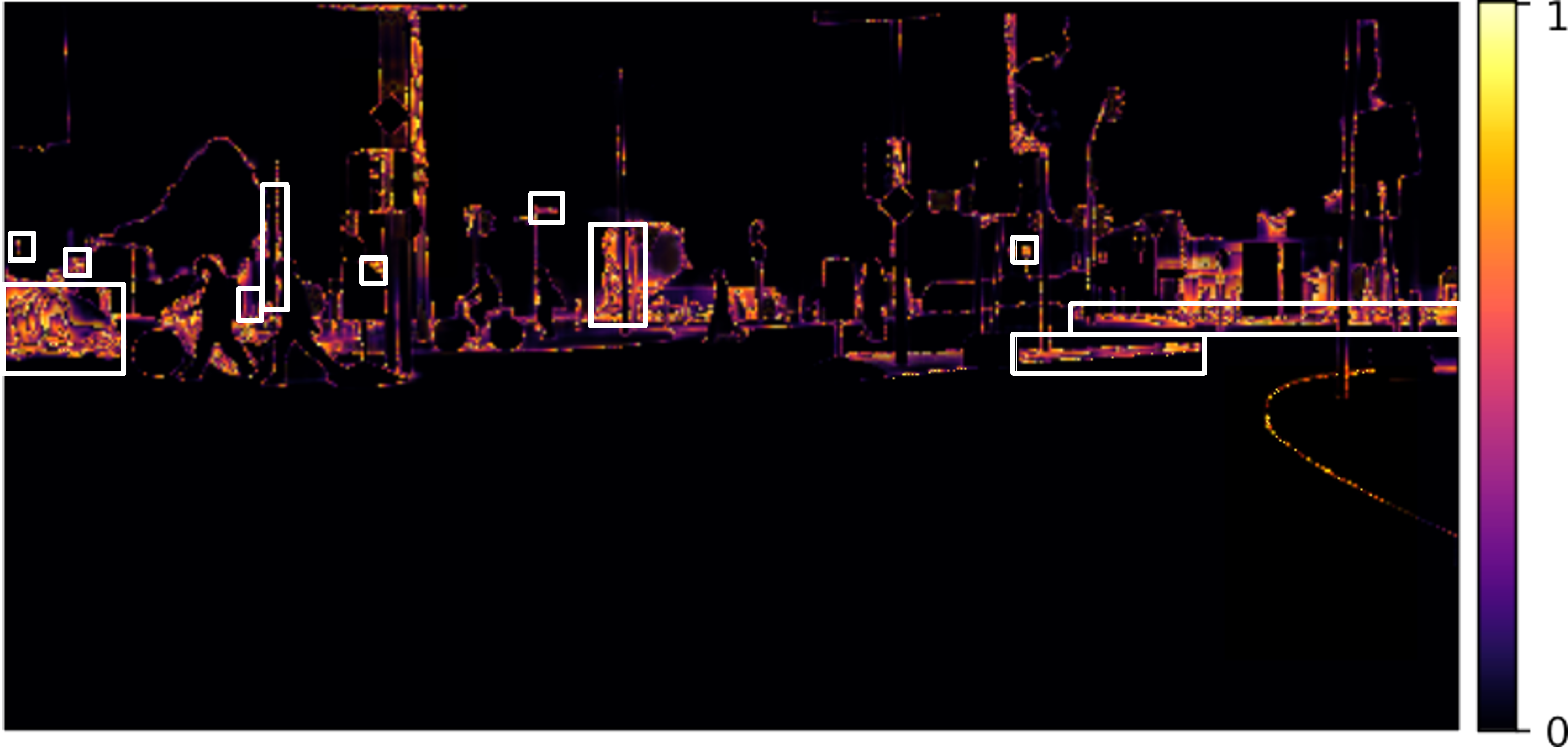}\\
    \includegraphics[width=0.48\linewidth]{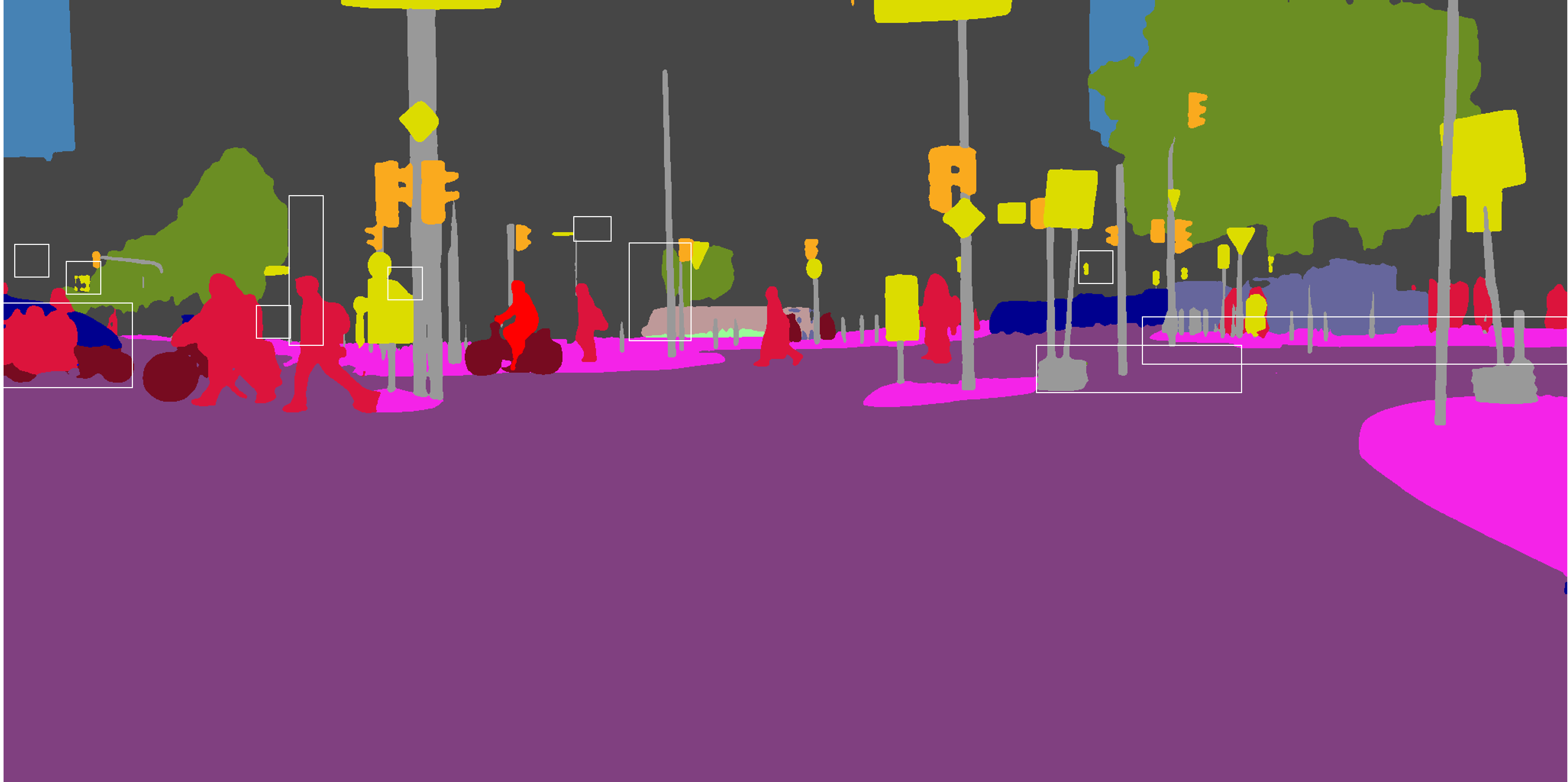}\hfill
    \includegraphics[width=0.48\linewidth]{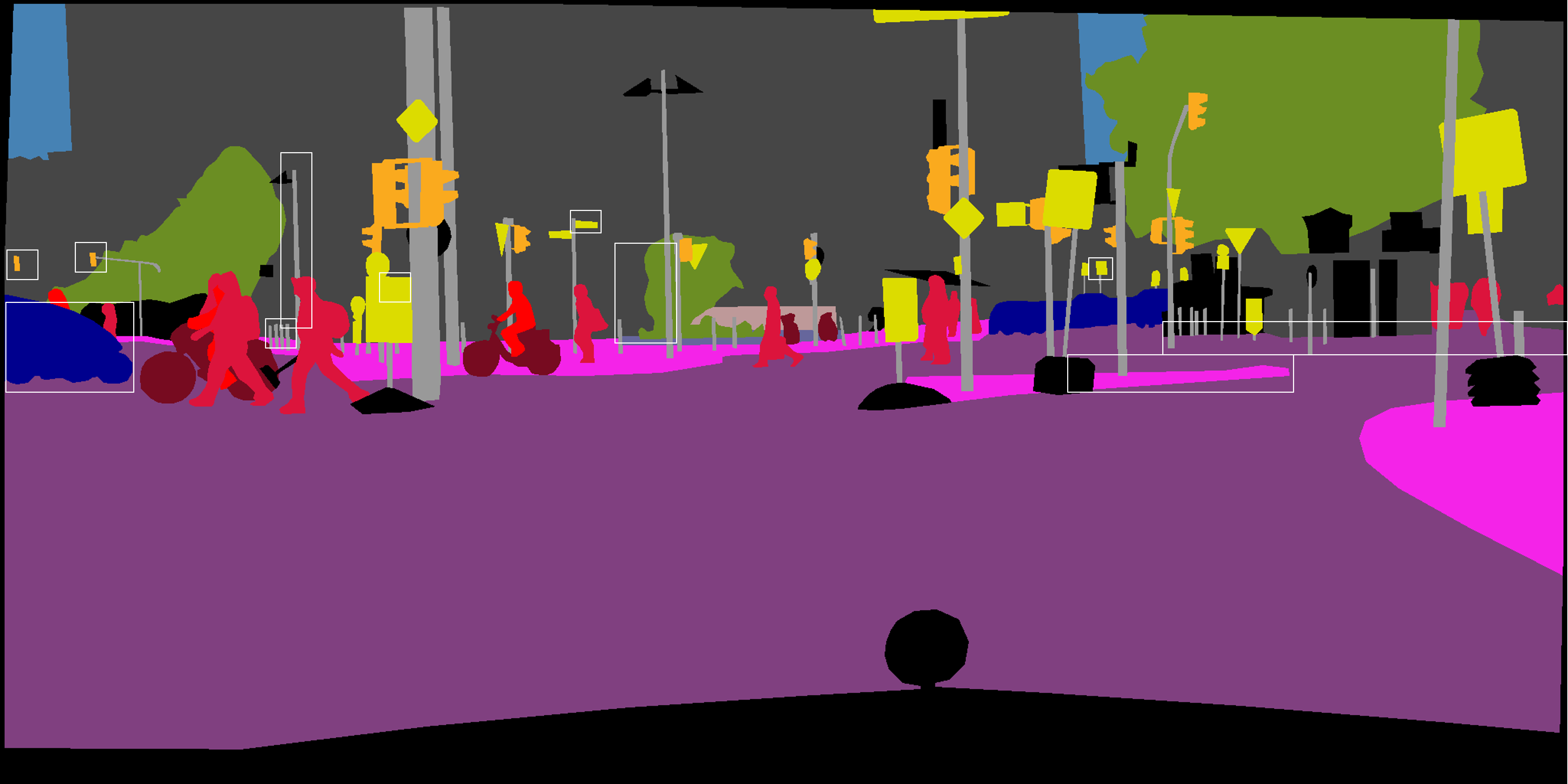}
    \caption{Per-pixel cross-entropy loss maps for the HRNetV2 model trained with a Nearest Neighbour down-sampling \cite{9052469} (top). Loss is normalised to the [0,1] range. In the learning process, the missing of a full structure is equally penalised as edge shifting, as can be observed in the highlighted areas, by comparing the loss values in the areas where the output (bottom, left) and the ground truth (bottom, right) segmentation maps differ. Note that a black colour in the ground truth represents pixels without annotation to be discarded during training.}
    \label{fig:loss}
    
\end{figure}
Building up from Convolutional Networks (CNN) \cite{7298965}, significant advances have been made in semantic segmentation in recent years. These are mainly focused on improving performance \cite{deeplabv3plus2018, 9052469} and inference speed \cite{mehta2018espnetv2,Yang2021RealtimeSS}.

To extract finer details, semantic segmentation methods tend to present low output strides at the cost of reducing the receptive field of the network. To counteract the small receptive fields, Pyramid Pooling is employed to extract information at multiple scales \cite{deeplabv3plus2018}. DeepLab \cite{deeplabv3plus2018} employs atrous convolutions with different levels of dilation to generate features with multiple receptive fields. However, Pyramid Pooling has been recently discarded due to its fixed square context regions (pooling and dilation are typically applied in a symmetric fashion), and replaced by relational context methods, such as HRNetV2 \cite{9052469}, which employs attention to learn the relationships among pixels. Therefore, it is not limited to square regions \cite{9052469} at the cost of greater computational complexity \cite{ borse2021inverseform,NVIDIA_semseg,YuanCW20,  zhou2022rethinking}. These fine details (especially edges) strongly drive the loss of semantic segmentation methods \cite{borse2021inverseform, 2019_zhao_rmi} (see Figure \ref{fig:loss}). We argue that the edge information is usually biased by the down-sampling strategy, producing unstable cues for learning. Edge noise has been tackled through region-based losses \cite{2019_zhao_rmi,10173725}, which handle this instability by aggregating information from both sides of the edges, at the cost of increasing the computational requirements linearly with the size of the neighbourhood employed \cite{2019_zhao_rmi}. However, we argue that these losses do not cover the root of the problem, which is the noise introduced by the unpaired down-sampling strategy.  To that end, our proposal presents smooth transitions between structures, thus, reducing the impact of edges with negligible computational requirements.

\subsection{Budget training semantic segmentation}
In the pursuit of achieving high performance on lightweight architectures while operating within resource-constrained environments, extensive research has explored two primary strategies.
One involves architectural modifications that utilise techniques such as channel shuffling \cite{8803154}, separable convolutions \cite{Lin2020GraphGuidedAS} and group convolutions \cite{mehta2018espnetv2}. These modifications serve to significantly reduce the number of parameters.
Conversely, the alternative approach focuses on the combination of diverse low-resolution information sources, such as different resolutions \cite{Yu_2018_ECCV, Li2019DFANetDF, m_Huynh-etal-CVPR21, hong2021deep}, multiple frames \cite{10112629}, and edge information \cite{xu2022pidnet}. For both strategies, the utilisation of low-resolution images is a pivotal objective, as it exponentially reduces the computational load required for image processing or semantic information fusion.

However, it is worth noting that current efforts to employ low-resolution images involve a trade-off between computational efficiency and segmentation accuracy, with the latter experiencing a significant performance decrease. To the best of our knowledge, our proposal is the first to effectively train with low-resolution images while achieving results on par with the state-of-the-art performance.
 
\subsection{Alignment of colour and label information}
To provide a common sample selection for both colour and label images, down-sampling strategies based on multi-patch division \cite{m_Huynh-etal-CVPR21} and learned sampling \cite{9008795} have been proposed. However, the latter results in the cascading of sampling errors to the segmentation model \cite{9008795}. Multi-patch training frameworks are, in turn, requested to process versions of the image multiple times, with one being a regularly down-sampled version of the image to conserve the scene context and the other crops of different scaled versions of the image to conserve small-scale object details. In practice, this framework allows us to effectively train models with small-resolution input crops. However, 85 crops per image have to be processed, thus multiple GPUs are still needed because of the required large batch sizes. In this paper, we showcase that effective training of semantic segmentation models can be performed even in small resolutions by pairing the colour and label sampling strategies.
%%%%%%%%%%%%%%%%%%%%%%%%%%%%%%%%%%%%%%%%%%%%%%%%%%%%%%%%%%%%%%
\subsection{Soft-label encoding}
Typically in classification tasks, the ground truth data is represented by hard-labelled one-hot encoded vectors with ones in the corresponding class. Label smoothing \cite{10.5555/3454287.3454709} aims at improving the performance of deep learning methods by averaging hard labels following the distribution of labels in a training scope. This smoothing acts as a regularisation by preventing the network from becoming overconfident on the ground truth labels. Müller. et. al, \cite{10.5555/3454287.3454709} study the impact of label smoothing and demonstrate its positive effect on generalisation. Similarly, in label presentation, soft-labels are used to interpolate intermediate labels between predefined categories (e.g., for human age estimation \cite{10.1007/978-3-319-54187-7_14}, depth estimation, or horizon line regression \cite{Diaz_2019_CVPR}). In this paper, soft-labels provide a smooth transition between semantic edges and help conserve the information of classes defined by small objects that are prone to be removed after down-sampling.   

\section{Pairing colour and label down-sampling}
%In this section, we formalize the down-sampling for semantic segmentation and describe our proposal.
\label{sec:SSR}
\subsection{Problem definition: Down-sampling in semantic segmentation}
Semantic segmentation is a multi-class multi-label classification problem generally defined on colour images $\mathbf{x}\in\mathbb{R}^{3, H, W}$\textemdash where $W,H \in \mathbb{N}$ are the width and height of the image and define its resolution. 
%Each colour image has associated a label image $\mathbf{y}\in\mathbb{N}^{C,H,W}$, where each pixel is tagged as a one-hot encoded vector of the $C$ possible semantic classes. 
%CORRECTED VERSION BY JUANCARLOS
Each colour image has associated a label image $\mathbf{y}\in\mathbb{N}^{1,H,W}$, where each pixel is assigned a numerical value corresponding to one of the $C$ semantic classes.
As aforementioned, colour and label images are commonly down-sampled due to resource constraints, image resolution alignment, or data augmentation. Formally, the high-level standard procedure for semantic segmentation, including  down-sampling is defined by:
\begin{equation}
\begin{split}
    &\mathbf{x}\in \mathbb{R}^{3, H, W} \xrightarrow{d_x(\cdot,\boldsymbol{\gamma)}} \mathbb{R}^{3, H', W'} \xrightarrow{G(x; \theta)} \mathbb{R}^{C, H', W'},\\
    &\mathbf{y}\in \mathbb{N}^{ 1, H, W} \xrightarrow{d_y(\cdot,\boldsymbol{\gamma)})} \mathbb{N}^{ 1, H' , W'}, 
\end{split}
\label{eq:seg}
\end{equation}
where  $d_x$ and $d_y$ are the respective down-sampling functions for colour and label images; $\boldsymbol{\gamma}$ is the scale factor, $W'= \gamma_w\cdot W$ and $ \quad H' =\gamma_h\cdot H$ is the target resolution; and $G$ is the semantic segmentation model whose input is each of the down-sampled colour images and outputs a probability map $G(\mathbf{x};\theta)\in\mathbb{R}^{C,H',W'}$, such that $\sum_{c=1}^C G(\mathbf{x};\theta)_{h,w}^c = 1 \hspace{0.25em} \forall \hspace{0.25em} h,w$. %Validation is always performed on full-resolution images.

Typically, $d_x$ is a smooth down-sampling strategy, a common choice is the highly efficient, yet prone to aliasing and shift variant, bilinear down-sampling \cite{Han2013/03}.  In bilinear down-sampling, the sampled pixel is obtained by a weighted average of the pooled pixels. Differently, the categorical nature of the label image does not allow to apply the same strategy, so $d_y$ typically employs the Nearest Neighbour down-sampling \cite{NN}. The lack of semantic relationships between categories precludes the use of smooth down-sampling strategies on the label image \cite{9008795}.

% As these  discrepancies are common and transversal to all the classes, the noise-ignoring learning inductive bias cannot hamper them. 

\begin{figure}[tp]
    \centering
    %\begin{subfigure}[b]{\linewidth}
    \includegraphics[width=\linewidth]{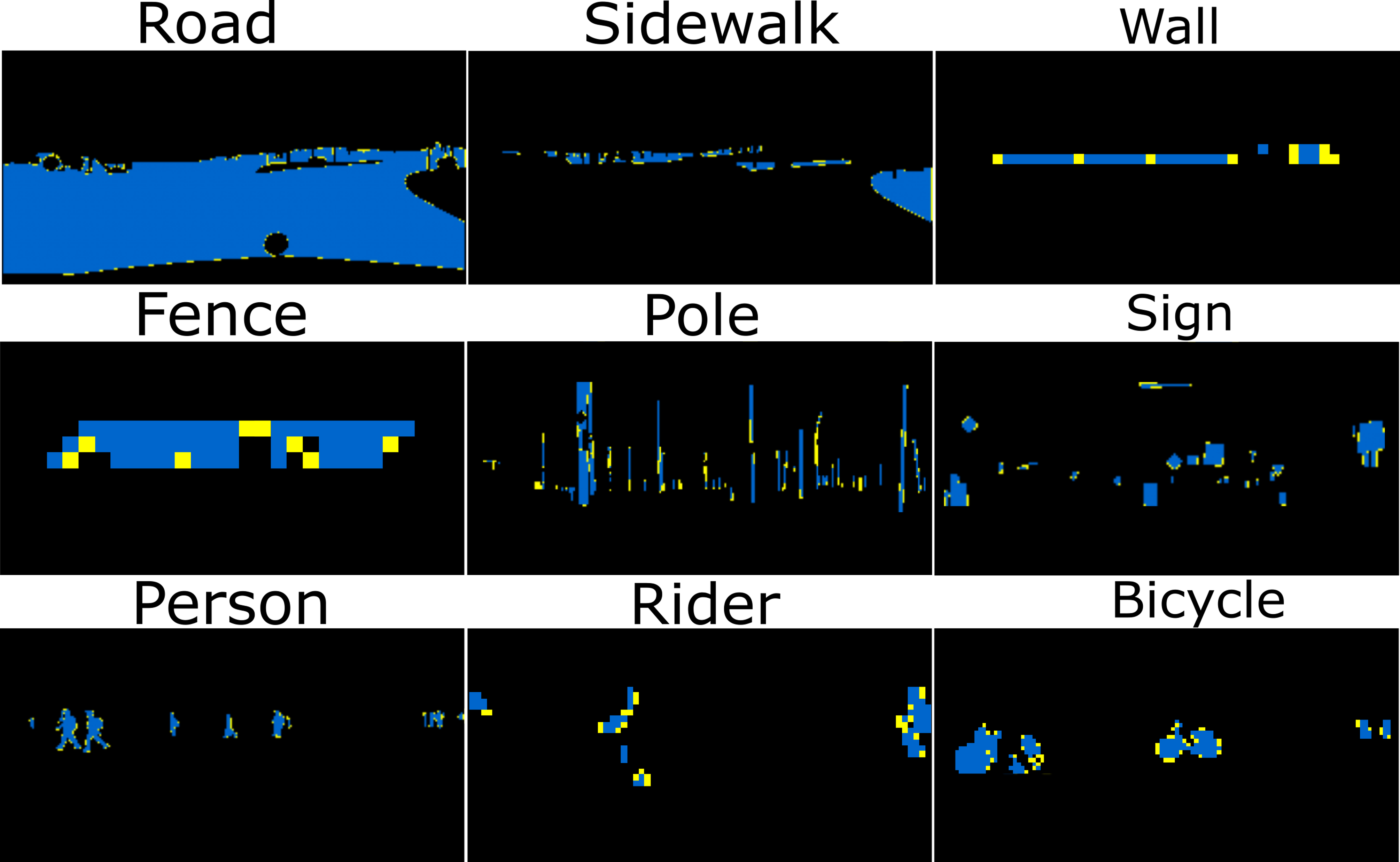}
    \caption{Per-class visual comparison  between Nearest Neighbour (blue), and our soft-label (blue and yellow) after a $\frac{1}{8}$ down-sampling of the label image in Figure \ref{fig:loss}. Note the creation of jagged edges (step-like borders of blue areas) and gaps (discontinuities in blue areas).}
    %\end{subfigure}
    
    % \begin{subfigure}[b]{\linewidth}
    
    % \includegraphics[width=\linewidth]{DiffOrig_100.png}
    % \caption{Comparison of the percentage difference after a $\frac{1}{8}$ label image down-sampling for Nearest Neighbour (Baseline) and soft-label (Ours). Positive (Negative) values indicate label information miss-match by generated artefacts (removal of structures).}
    % \end{subfigure}
    %\caption{Generated artefacts and information conservation of Nearest Neighbour and the proposed down-sampling strategies.}
    \label{fig:label_rescale}
    
\end{figure}

Therefore,  colour and label down-sampling strategies are unpaired. This generates discrepancies that are forced to be learned and memorised by the model. In fact, these discrepancies have been shown to dominate the learning attention in the last critical training epochs \cite{borse2021inverseform}. Remarkable effects of these discrepancies are the shape modification of inter-class boundaries (see \textit{road} and \textit{sidewalk} in Figure \ref{fig:label_rescale}), the hallucination of gaps in continuous structures (see \textit{pole} in Figure \ref{fig:label_rescale}), and the elimination of small structures (see \textit{wall} in Figure \ref{fig:label_rescale}). Together, these effects affect the quantity and quality of the training data, as illustrated in Figure \ref{fig:label_rescale}. As an example, the training of a state-of-the-art model \cite{9052469} with images down-sampled to $\frac{1}{16}$ of the original resolution, results in mIoU performances of 0.3, 3.1, and 0.8 for the \textit{fences, bikes} and \textit{motorcycles} semantic classes. These values represent more than a 98\% drop in performance compared to those obtained when training with half-size or full-resolution images \cite{borse2021inverseform,li2022deep}. 

\subsection{Proposed down-sampling strategy}
To pair the down-sampling strategies for the colour and the label images, we first change the representation of the categorical information in the label image $\mathbf{y}$  to a one-hot encoding version \cite{OHE}, which is shaped into the matrix $\mathbf{\hat{y}}\in\mathbb{R}^{C,H,W}$. This matrix has ones in the spatial coordinates of the \(c\)-planes indicated by the class values at the same spatial coordinates in the label image.  For instance, for a pixel $(h,w) \in [H, W]$ with label $\mathbf{y}_{h,w} = c$, its one-hot encoded representation is:
\begin{equation}
\mathbf{\hat{y}}_{h,w} = \left \{ {\hat{y}}_{h,w}^{k} \right \}_{k=1}^C
\left\{
	\begin{array}{ll}
		{\hat{y}}_{h,w}^{k}= 1, k = c \\
		{\hat{y}}_{h,w}^{k}=0, elsewhere.
	\end{array}
\right.
\end{equation}

By means of this one-hot representation, rather than performing a weighted average of the corresponding sampled pixels as for the colour image, one can estimate the proportion of each class in the set of labels corresponding to the pixels in the down-sampling region. Thereby, defining a vector of soft-labels for each pixel to be down-sampled. 

Formally, the down-sampling on the colour images can be formulated as a regional combination of the pixel values in the image $\mathbf{x}$ according to a weight function $f:\mathbb{N}^2\times\mathbb{N}^2\rightarrow \mathbb{R}$ that is applied to each pixel $(h,w) \in H \times W$ and mapped to each pixel $(u,v) \in H' \times W'$ of the down-sampled (target) image $\mathbf{\overline{x}}$ for each channel $k$:

\begin{equation}
        \mathbf{\overline{x}}_{u,v}^k=d_x(\mathbf{x},\boldsymbol{\gamma}) = \sum_{h,w}^{H,W} f(h,w; u,v) \cdot \mathbf{x}_{h,w}^k .
       % \overunderset{c=1}{3} \left\{ \sum_{h,w}^{H,W} \left\( f(h,w; h',w') \times \mathbf{x}_{h,w} \right\) \right\},     
\end{equation}
For instance, for bilinear down-sampling, $f$ is an even, decreasing function, that only depends on the distance of the target down-sampled pixel $\mathbf{\overline{x}}_{u,v}$ and the sampled pixel $\mathbf{x}_{h,w}$ in the original resolution colour image. For each down-sampled pixel, its corresponding multi-class label $\mathbf{\overline{y}}_{h,w}$ is obtained by combining the one-hot encodings $\mathbf{\hat{y}}_{h,w}$ according to the employed down-sampling strategy $f(h,w; u,v)$.% For example, for  $\gamma=\frac{1}{2}$ bilinear down-sampling, the vector of soft-labels for pixel $(u,v)$ is obtained by:

\begin{figure}[btp]
    \centering
    \includegraphics[width=\linewidth]{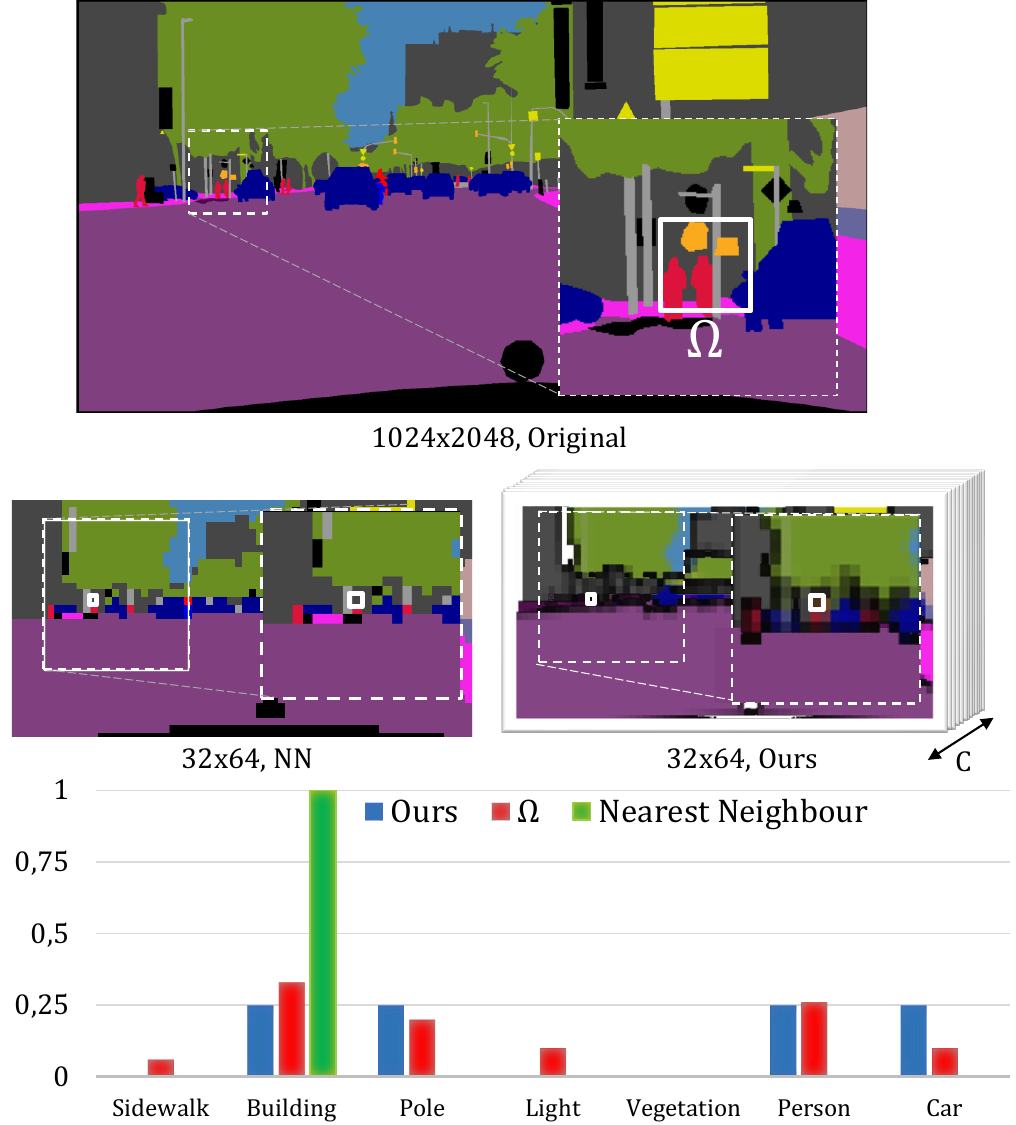}
    \caption{Visual comparison of the information conserved by down-sampling strategies for a Cityscapes image where $C=19$ classes and the down-sampling factor is $\gamma = (1/32, 1/32)$. Top-row image shows the original-resolution labels. The middle row represents the down-sampled labels for our proposal (right) and the Nearest Neighbour (left). The bottom row depicts the distribution of present semantic classes of the highlighted region $\Omega$ in the original label image (blue) and the ones obtained for the down-sampling of $\Omega$ with Nearest Neighbour (green) and the proposed one (red). Note that $\Omega$ is resized to a single pixel for the down-sampled versions (highlighted in white).}
    %As our proposal generates a $C$ dimensional provability vector per pixel, to visualize it, we represent the per-pixel entropy.}
    % \vspace{-1 em}
    \label{fig:soflabels}
\end{figure}

% \begin{equation}
% \begin{split}
%        \mathbf{\overline{y}}_{u,v}^k =& 0.25 \cdot \mathbf{\hat{y}}_{h,w}^k + 0.25 \cdot \mathbf{\hat{y}}_{h+1,w}^k + \\
%        & 0.25 \cdot \mathbf{\hat{y}}_{h,w+1}^k + 0.25 \cdot \mathbf{\hat{y}}_{h+1,w+1}^k,      
% \end{split}
% \end{equation}
%where $h =  u\cdot 2$ and $w = v\cdot 2$.
Besides being aligned with the one used for the colour image, the proposed down-sampling strategy enlarges the amount of information maintained from the original label image after down-sampling. In Figure \ref{fig:soflabels}, we can observe that the widely applied Nearest Neighbour is worse at maintaining label information after down-sampling is performed.
\subsection{Controlling the impact of multi-class pixels in the training loss}
Typically, semantic segmentation training is driven by the minimization of the Cross-Entropy (CE) loss between the one-hot encoding of the label image $\mathbf{y}$, and the output probability map $G(\mathbf{x};\theta)$. However, CE does not return a zero loss for soft-labels with non-zero entropy (i.e., $\forall \hspace{.5mm} x \hspace{1.5mm}  CE(x,x)= H(x)$, where $H(x)$ is the entropy of the vector $x$) even if the predicted labels are perfectly aligned with the ground truth soft-label. Therefore, the Kullback-Leibler (KL) divergence loss \cite{kullback1951information} is often preferred for soft-label classification \cite{Diaz_2019_CVPR,10.1007/978-3-319-54187-7_14,DBLP:conf/iccv/TzengHDS15,Wang2020}:
\begin{equation}
    \mathcal{L}_{KL} = \sum_{k,u,v}^{C,H',W'}\mathbf{\overline{y}}^{k}_{u,v}\cdot log\left(\frac{\overline{\mathbf{y}}^{k}_{u,v}}{G(\mathbf{\overline{x}};\theta)^{k}_{u,v}}\right),
    \label{eq:CE}
\end{equation}
where $k$ is the class index. The employment of soft-labels gives rise to a new concept of single ($\exists c | \overline{\mathbf{y}}^c = 1$) and Multi-Class labels (MC). In general, one can assume that the soft-labels after down-sampling are more common to be single-class instances (when corresponding pixels are inner to the class region) than multi-class ones (when corresponding pixels are close to the region contour). Therefore, single-class instances are prone to be easier to learn and their learning may help that of multi-class ones. However, due to the imposed Nearest Neighbour sampling on the ground truth labels, semantic edges have been analysed to dumper semantic segmentation performance \cite{2019_zhao_rmi}. As a counter-measure, large batch sizes must be employed \cite{liang2019winter} and regional losses have been proposed to reduce the impact of the loss of edges \cite{borse2021inverseform, 2019_zhao_rmi}. We analyse the impact of MC pixels in the experimental section \ref{sec:EX}.

\section{Experimental results}
\label{sec:EX}
%In this section, we present our experimental procedure and report and discuss the experimental results of the proposed method. First, we introduce the datasets, evaluation metrics and implementations details. Then, we follow by presenting the results comparing the training requirements in terms of memory and semantic-segmentation performance. Finally, we include a comparison with the state-of-the-art and discuss on the impact and differences of using the KL and the CE losses during training. Qualitative examples can be found in the supplementary material.

\subsection{Setup}
\paragraph{Datasets}
We mainly evaluate our proposal using the \textbf{Cityscapes} dataset \cite{Cordts2016Cityscapes}, an urban scene dataset with $1024\times 2048$ images (3K for training and 0.5K for validation) and 19 semantic classes. Moreover, we consider other popular datasets to validate the generality of our proposal: Mapilliary \cite{MVD2017}  and ADE20K \cite{zhou2019semantic}. The \textbf{Mapilliary} dataset is another real-images urban scene dataset for benchmarking semantic segmentation methods.  It comprises 18K and 2K images for training and validation. It defines 65 semantic classes in a variety of image resolutions, ranging from $1024\times 768$ up to over $4000 \times 6000$. The \textbf{ADE20K} dataset is a general-purpose dataset for evaluating semantic segmentation methods. It is composed of 20K images in the training set and 2K images for validation. It defines 150 semantic classes in images of various resolutions ranging from $256\times256$ to $2100\times2100$.

\paragraph{Evaluation metrics}
We adopt the default semantic segmentation performance measure, the PASCAL VOC per-class Intersection over Union (IoU) \cite{Everingham10thepascal}, to quantify the similarity between the model prediction and the annotations. It is defined as $IoU = {TP}/{(TP+FP+FN)}$, where TP, FP, and FN stand for, respectively, True Positives, False Positives, and False Negatives. As a global performance evaluation, we use the overall mean IoU averaging IoU values for all classes (mIoU). All the reported models have been evaluated using the full-resolution images of the respectively validation sets.

\paragraph{Explored semantic segmentation architectures}
For the Cityscapes and Mapilliary datasets, we employ two high-performance architectures: HRNetV2-48 \cite{9052469} and DeeplabV3+ \cite{deeplabv3plus2018} with a backbone of WideResNet-38 aligned with \cite{rotabulo2017place}. For the ADE20K dataset, we employ SegFormer (MiT-B2) \cite{Xie2021SegFormerSA}.
 We compare our strategy against 23 state-of-the-art methods, 22 of them employing Nearest Neighbour down-sampling, and also we also consider \cite{9008795}  that explicitly learns a down-sampling alternative to Nearest Neighbour.

\paragraph{Implementation details}
We use PyTorch as our deep-learning framework.
As GPU hardware, experiments are carried out on a constrained setup with a single GPU (either a GeForce Titan 12GB,  Titan RTX 24GB, or A40 48GB). We employ the maximum batch size for each GPU: 6 and 20 for the 12GB GPU (input resolution of $256 \times 512$ and $128 \times 256$, respectively), 3 and 12 for the 24GB GPU (input resolution of $512 \times 1024$ and $256 \times 512$, respectively), and 6 and 3 for the 48GB GPU (input resolution of $512 \times 1024$ and $1024 \times 2048$, respectively). Our models are trained using SGD optimiser with momentum of $0.9$ and weight decay of $10^{-4}$ as in \cite{NVIDIA_semseg}.  No additional data is used, relying solely on the train and validation sets provided for each dataset. The code and setup for reproducibility are available at \href{https://github.com/vpulab/soft-labels-SS}{https://github.com/vpulab/soft-labels-SS}.

\subsection{The paired down-sampling enables effective and efficient semantic segmentation training }
We validate our proposal for the crop-based data augmentation widely used in semantic segmentation \cite{borse2021inverseform, li2022deep, NVIDIA_semseg}.
This augmentation defines an input resolution for training (i.e., crop size), performs random resizing of the original image and label data for all batch samples, and later extracts training crops from resized images. This resizing ranges from half to twice the defined crop size. We apply our down-sampling strategy to the resizing of label images, and we consider three down-sampling factors relative to the original training data resolution: $\frac{1}{8},\frac{1}{4}$ and $\frac{1}{2}$ (i.e., crop size). Therefore, we train models for each explored architecture (DeeplabV3+ and HRNetV2), considering the three down-sampling factors and the original resolution. As the baseline for comparison, we use the default unpaired down-sampling strategies for colour (bilinear) and label images (Nearest Neighbour).

\begin{figure}
    \centering
    \begin{subfigure}[b]{0.3\linewidth}
    \includegraphics[width=\textwidth]{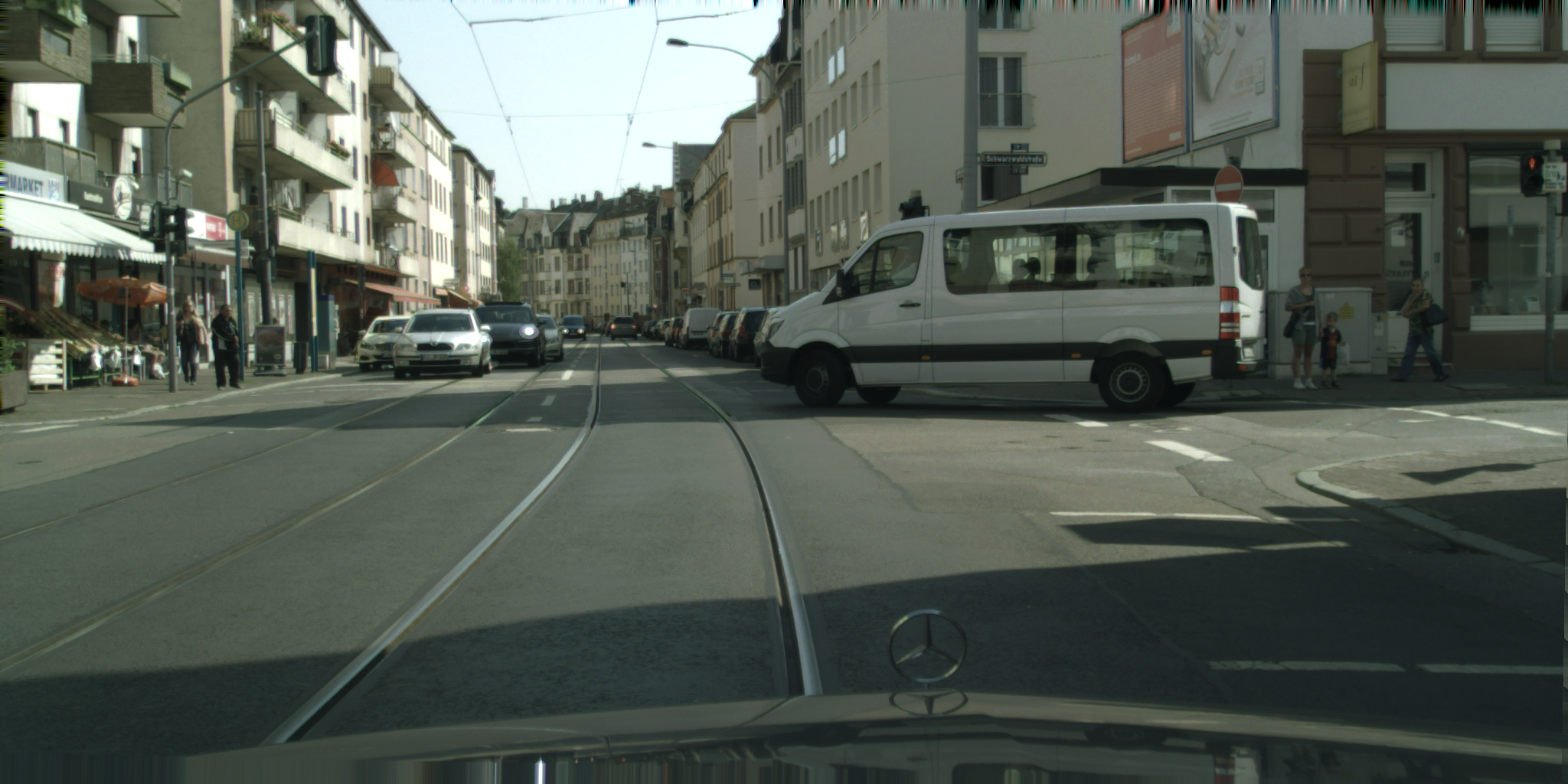}
    \end{subfigure}\hfill
    \begin{subfigure}[b]{0.3\linewidth}
    \includegraphics[width=\textwidth]{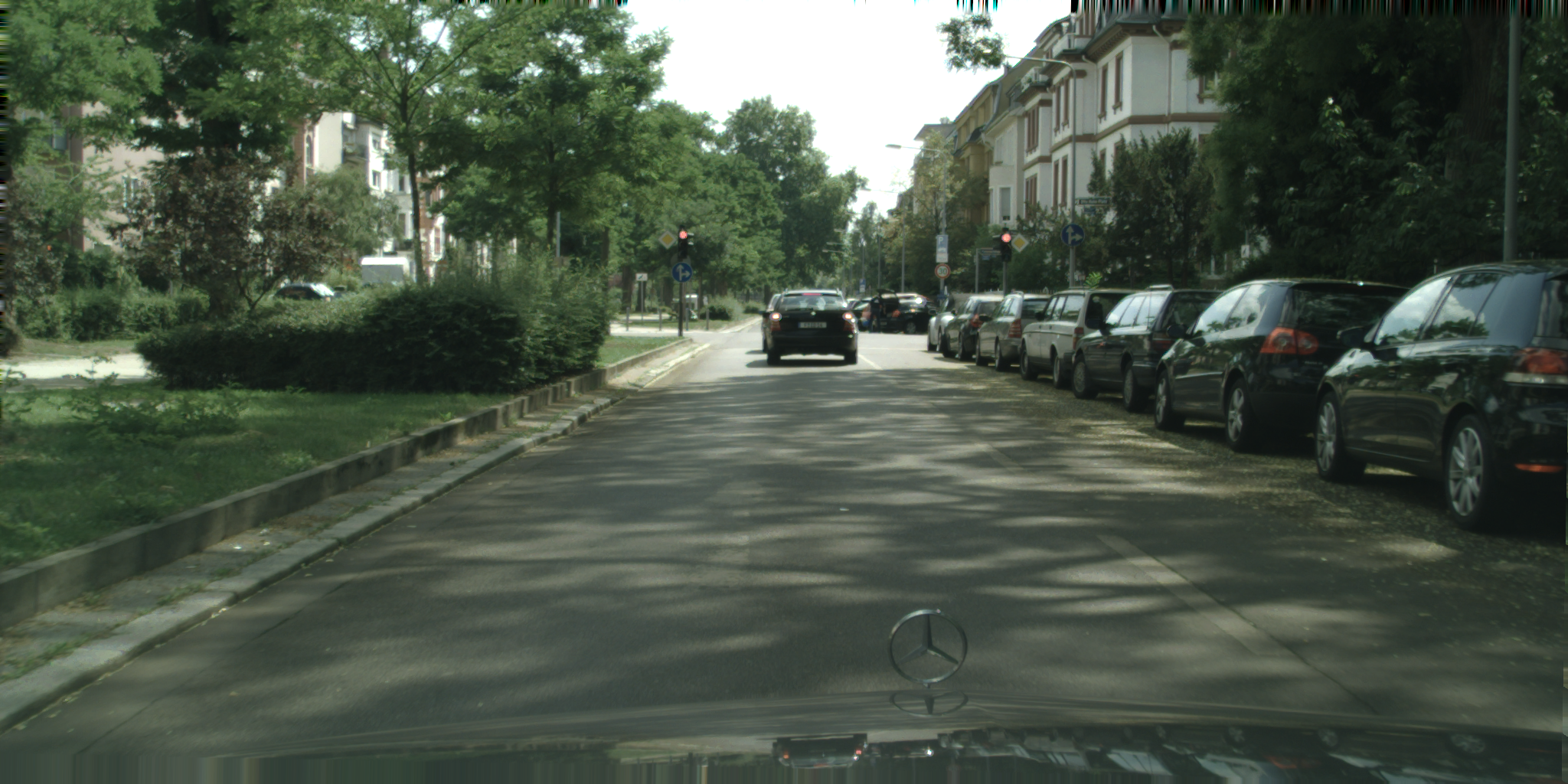}
    \end{subfigure}\hfill
    \begin{subfigure}[b]{0.3\linewidth}
    \includegraphics[width=\textwidth]{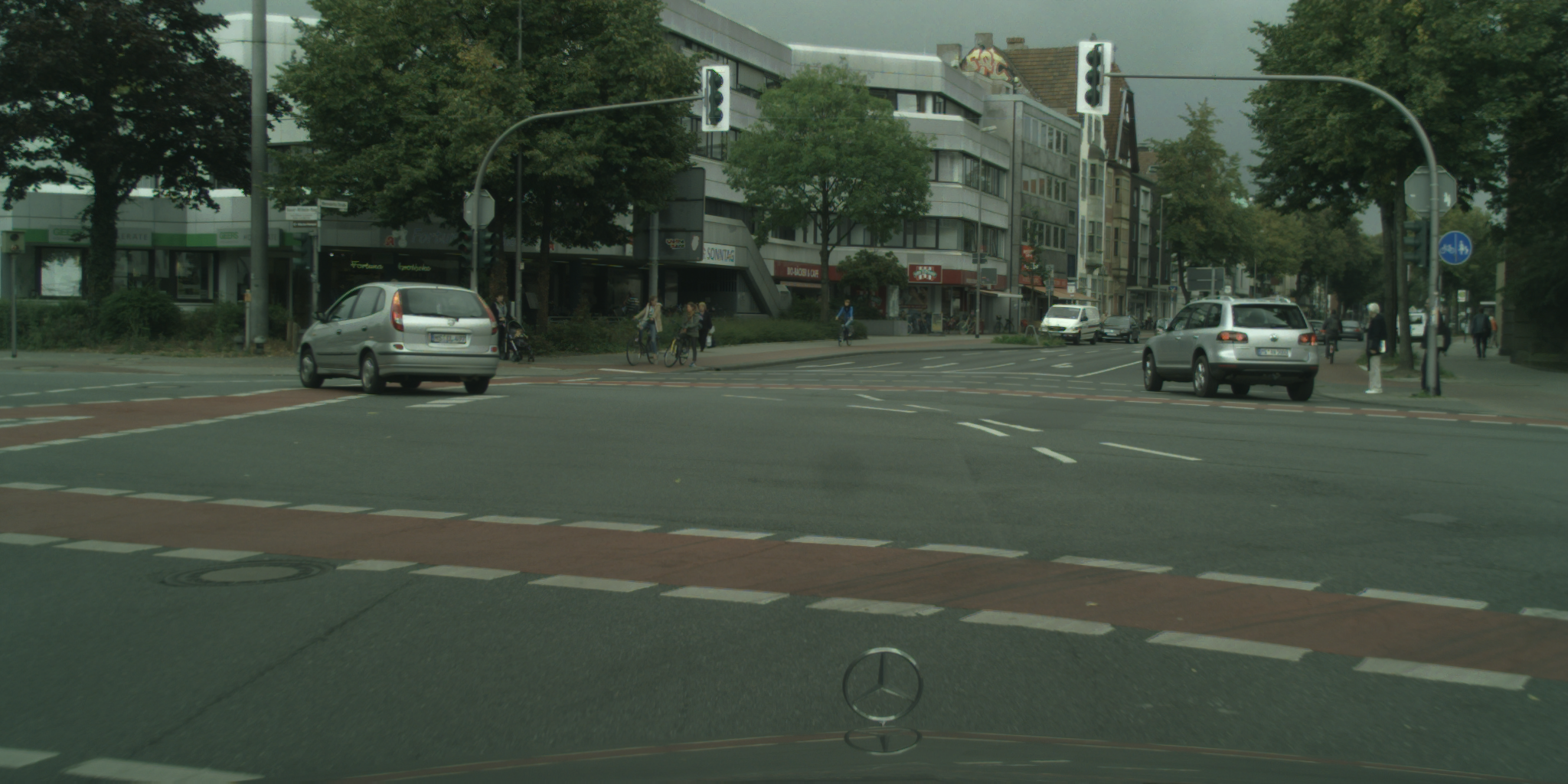}
    \end{subfigure}\hfill
    \\
    \begin{subfigure}[b]{0.3\linewidth}
    \includegraphics[width=\textwidth]{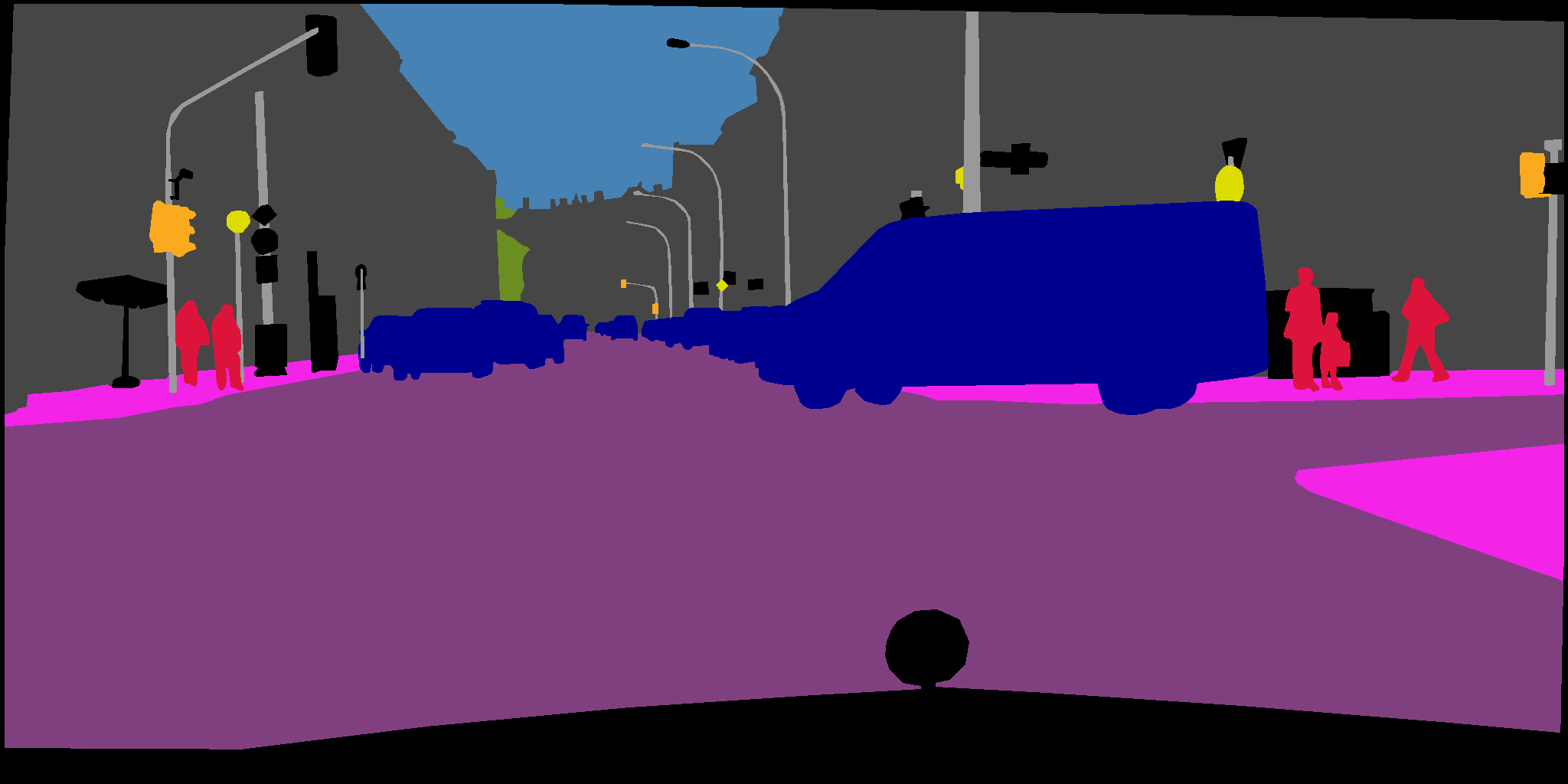}
    \end{subfigure}\hfill
     \begin{subfigure}[b]{0.3\linewidth}
    \includegraphics[width=\textwidth]{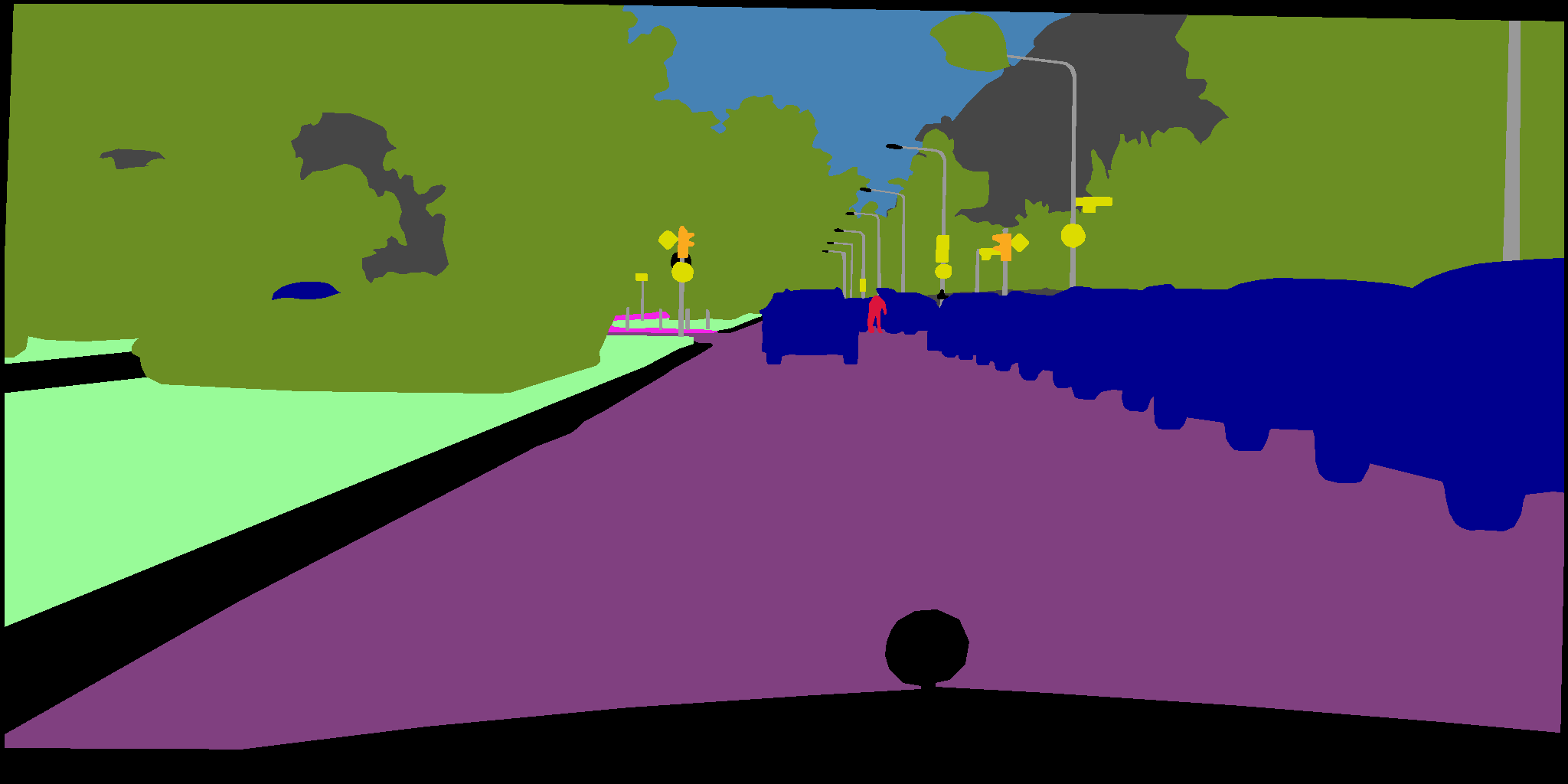}
    \end{subfigure}\hfill
    \begin{subfigure}[b]{0.3\linewidth}
    \includegraphics[width=\textwidth]{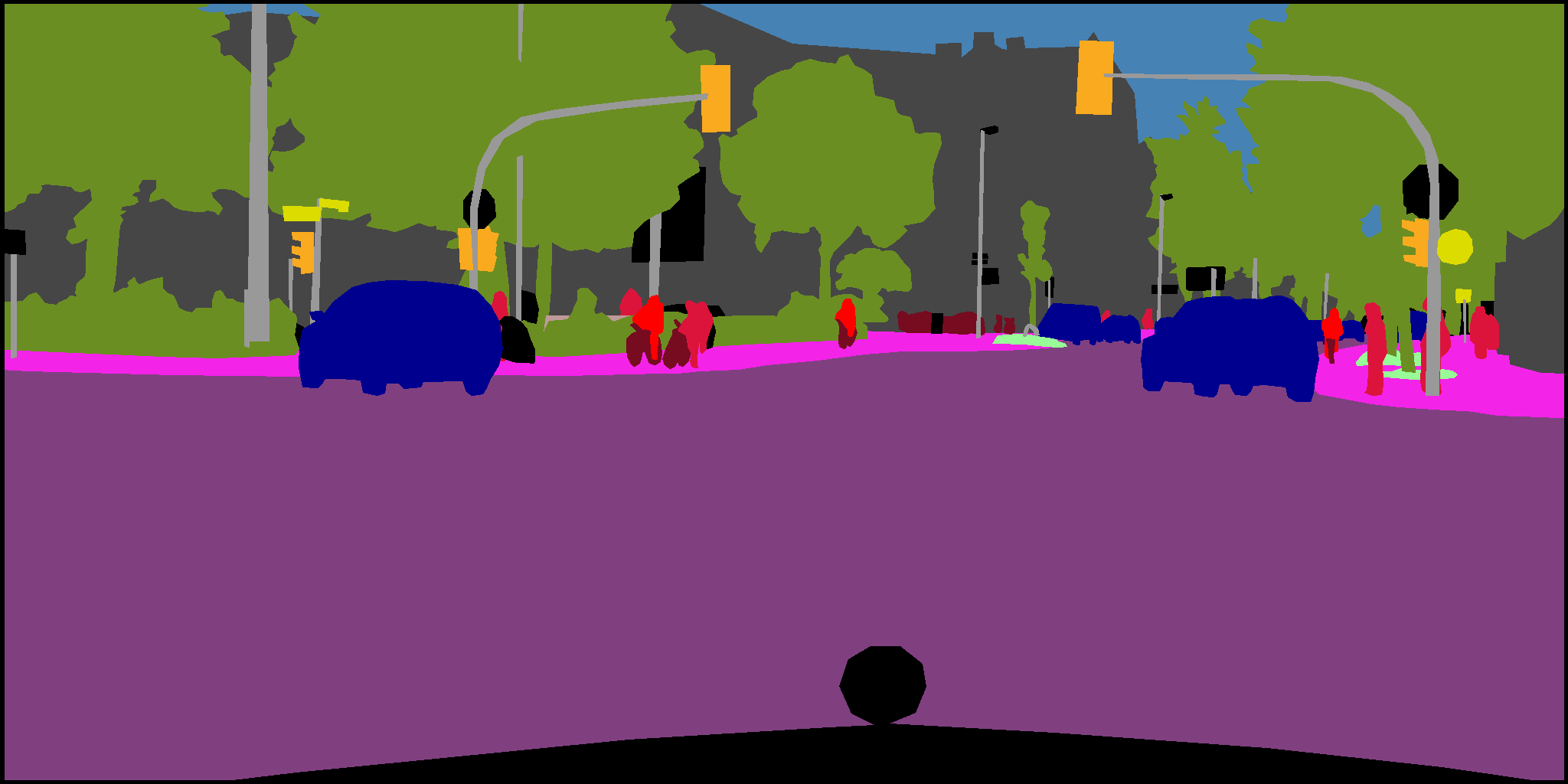}
    \end{subfigure}\hfill
    \\
    \begin{subfigure}[b]{0.3\linewidth}
    \includegraphics[width=\textwidth]{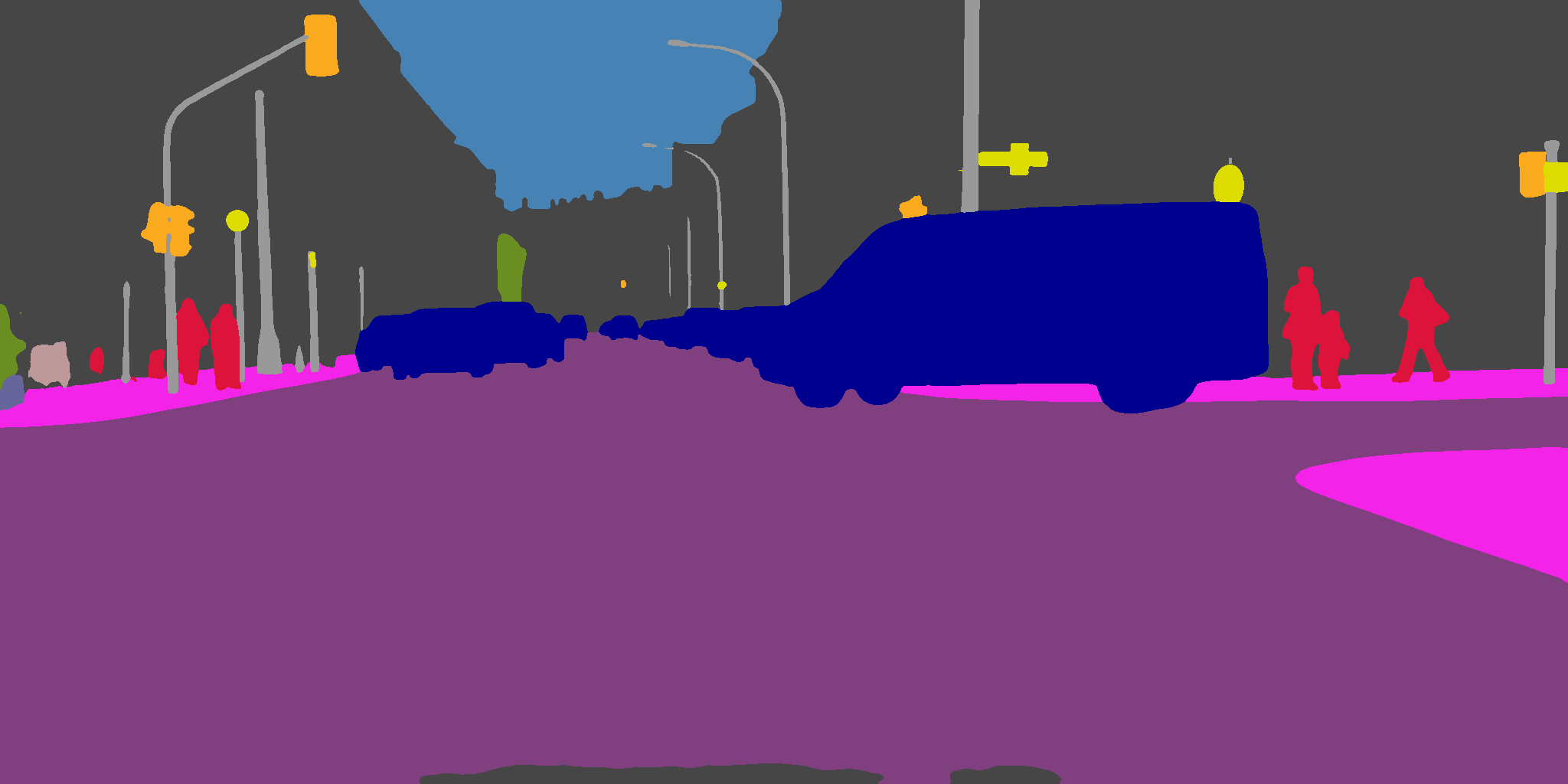}
    \end{subfigure}\hfill
    \begin{subfigure}[b]{0.3\linewidth}
    \includegraphics[width=\textwidth]{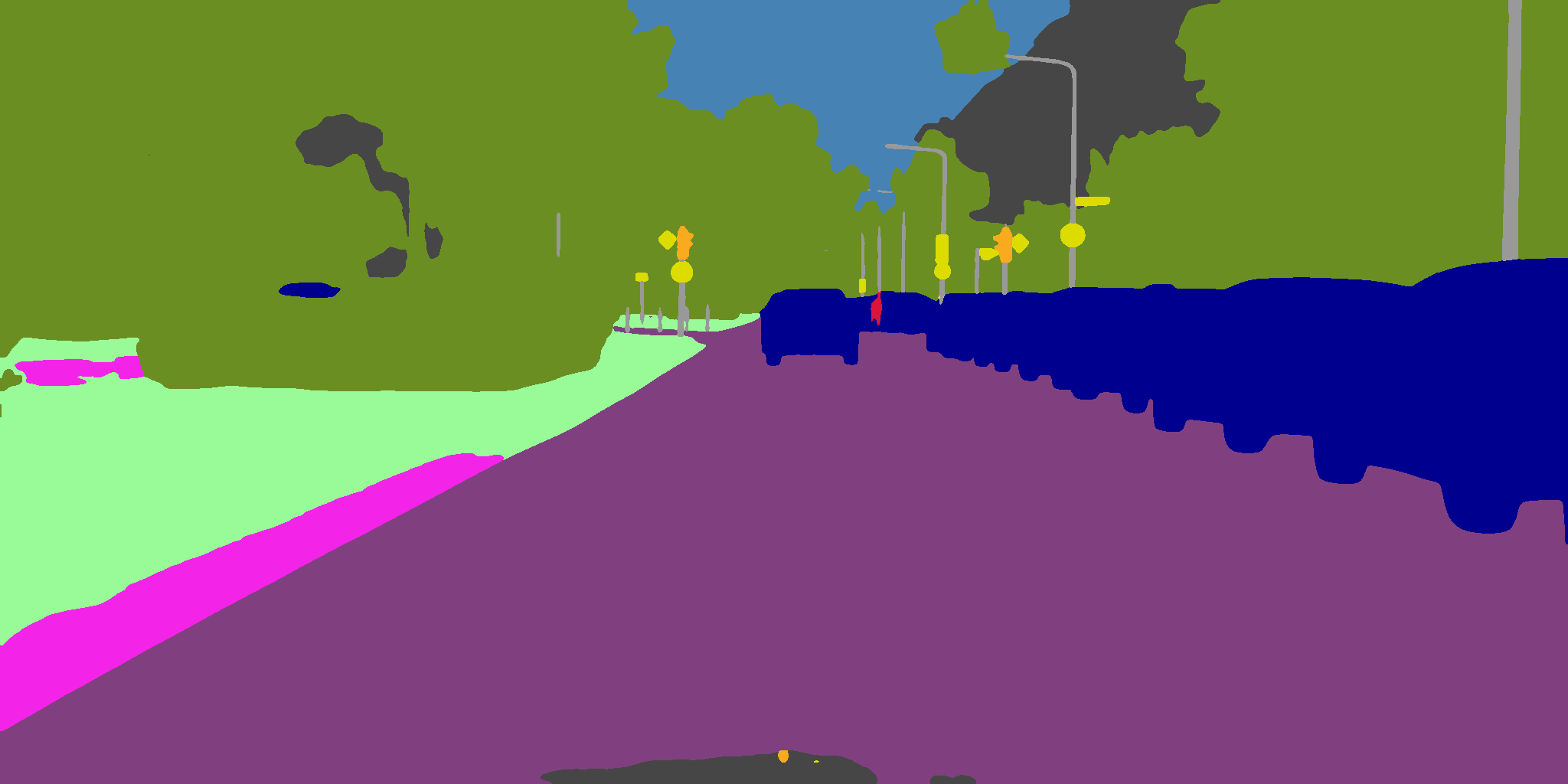}
    \end{subfigure}\hfill
     \begin{subfigure}[b]{0.3\linewidth}
    \includegraphics[width=\textwidth]{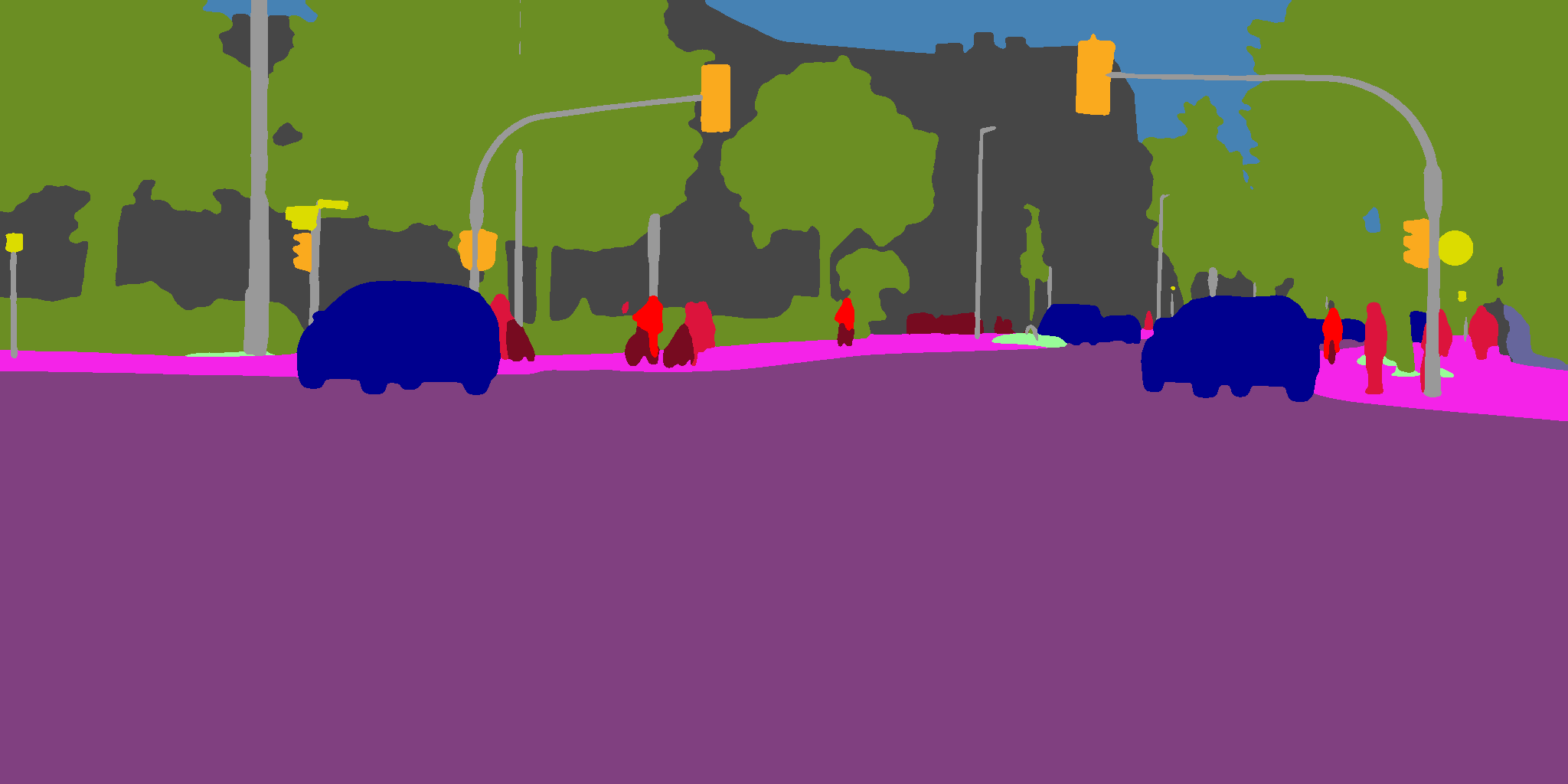}
    \end{subfigure}\hfill
    \\
    \begin{subfigure}[b]{0.3\linewidth}
    \includegraphics[width=\textwidth]{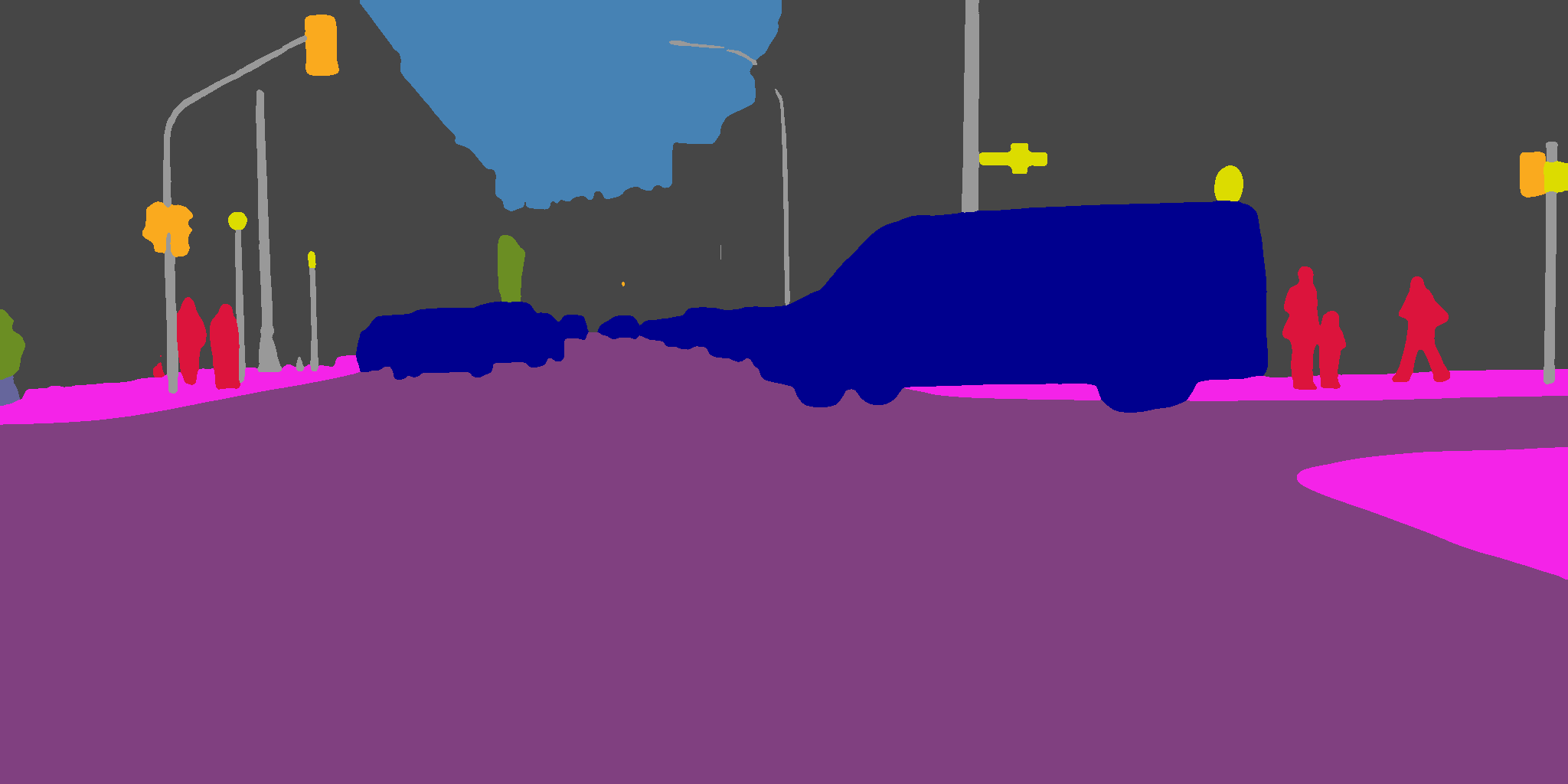}
    \end{subfigure}\hfill
    \begin{subfigure}[b]{0.3\linewidth}
    \includegraphics[width=\textwidth]{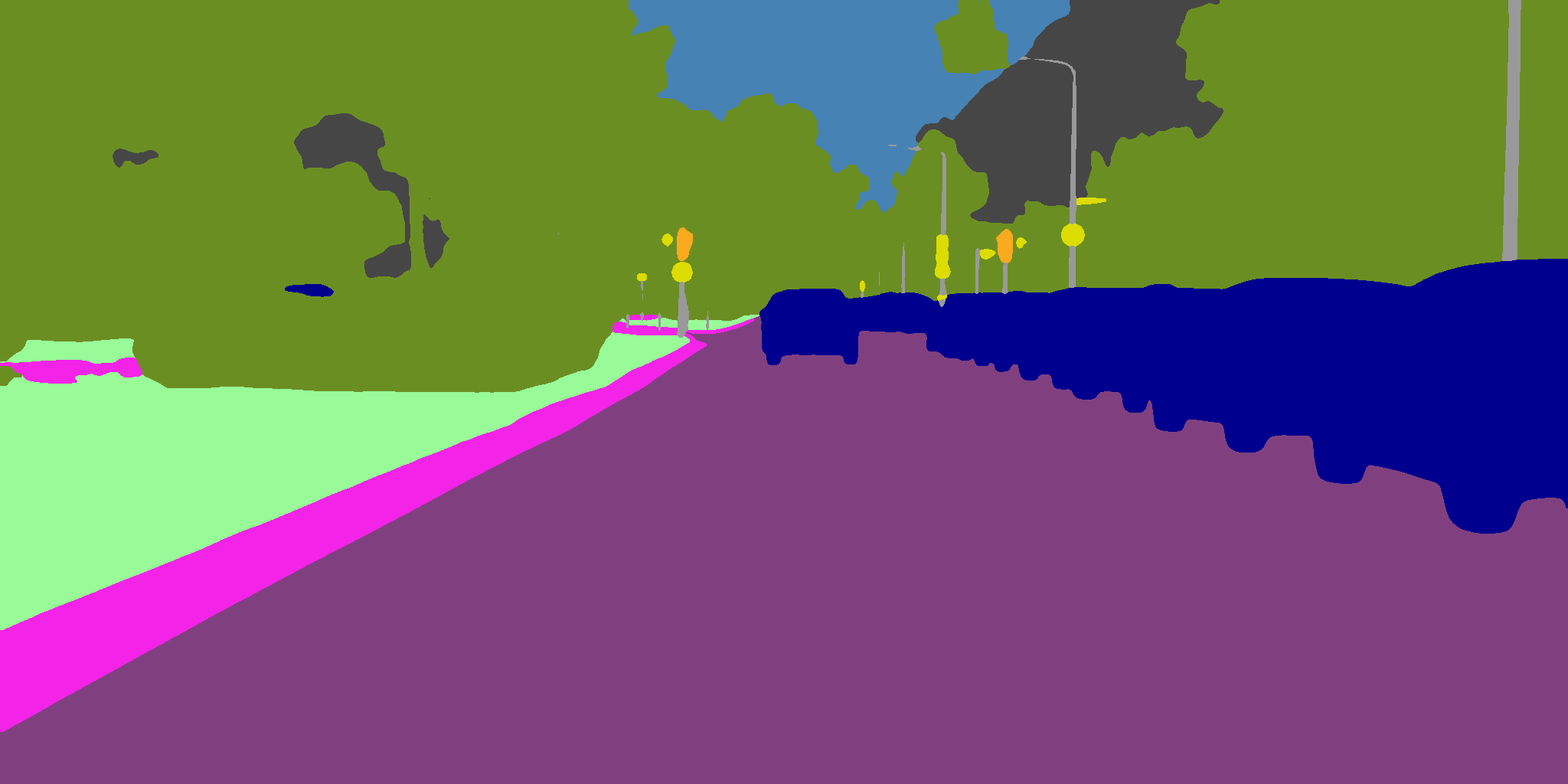}
    \end{subfigure}\hfill
    \begin{subfigure}[b]{0.3\linewidth}
    \includegraphics[width=\textwidth]{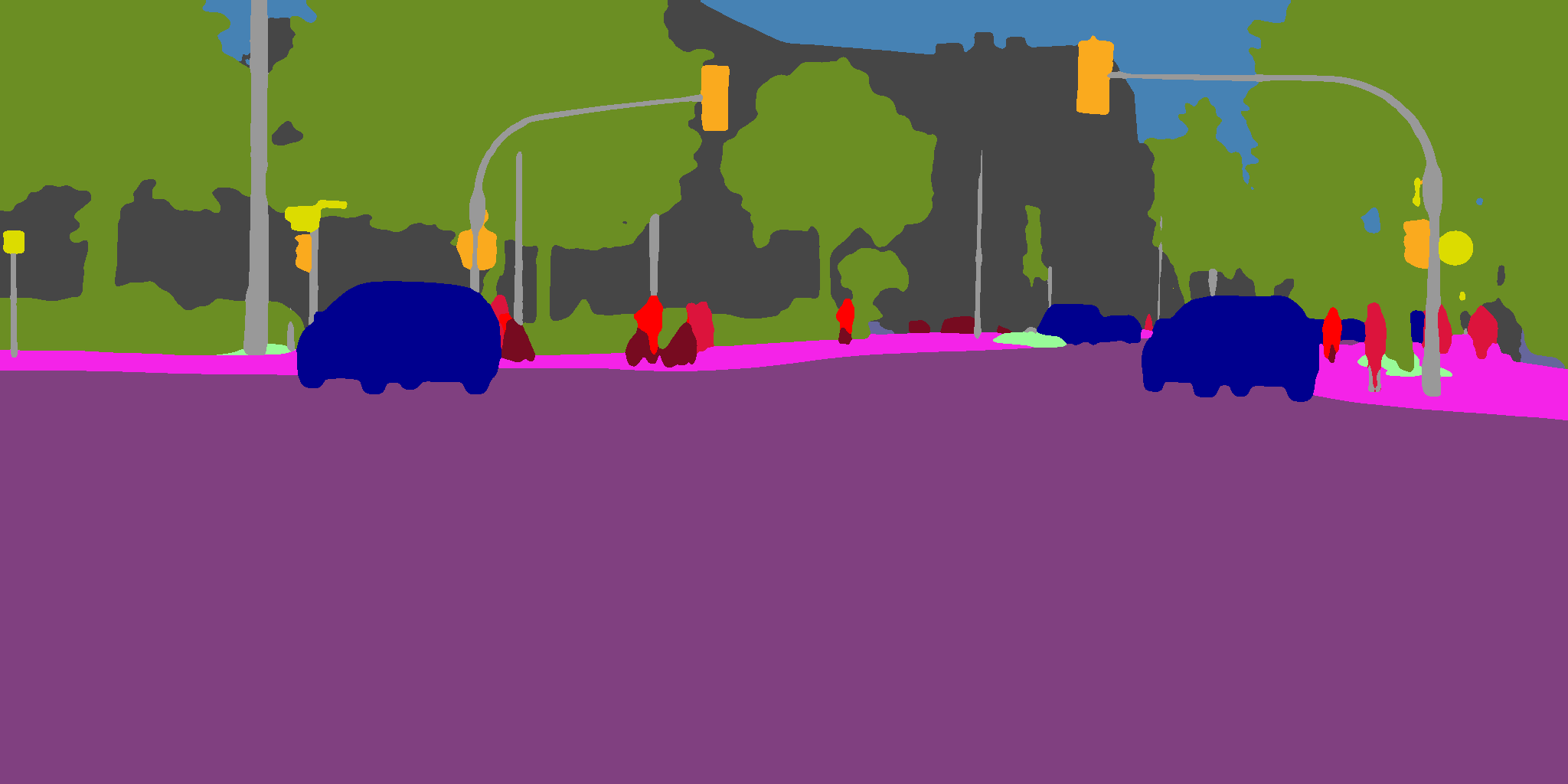}
    \end{subfigure}\hfill
    \\
    \begin{subfigure}[b]{0.3\linewidth}
    \includegraphics[width=\textwidth]{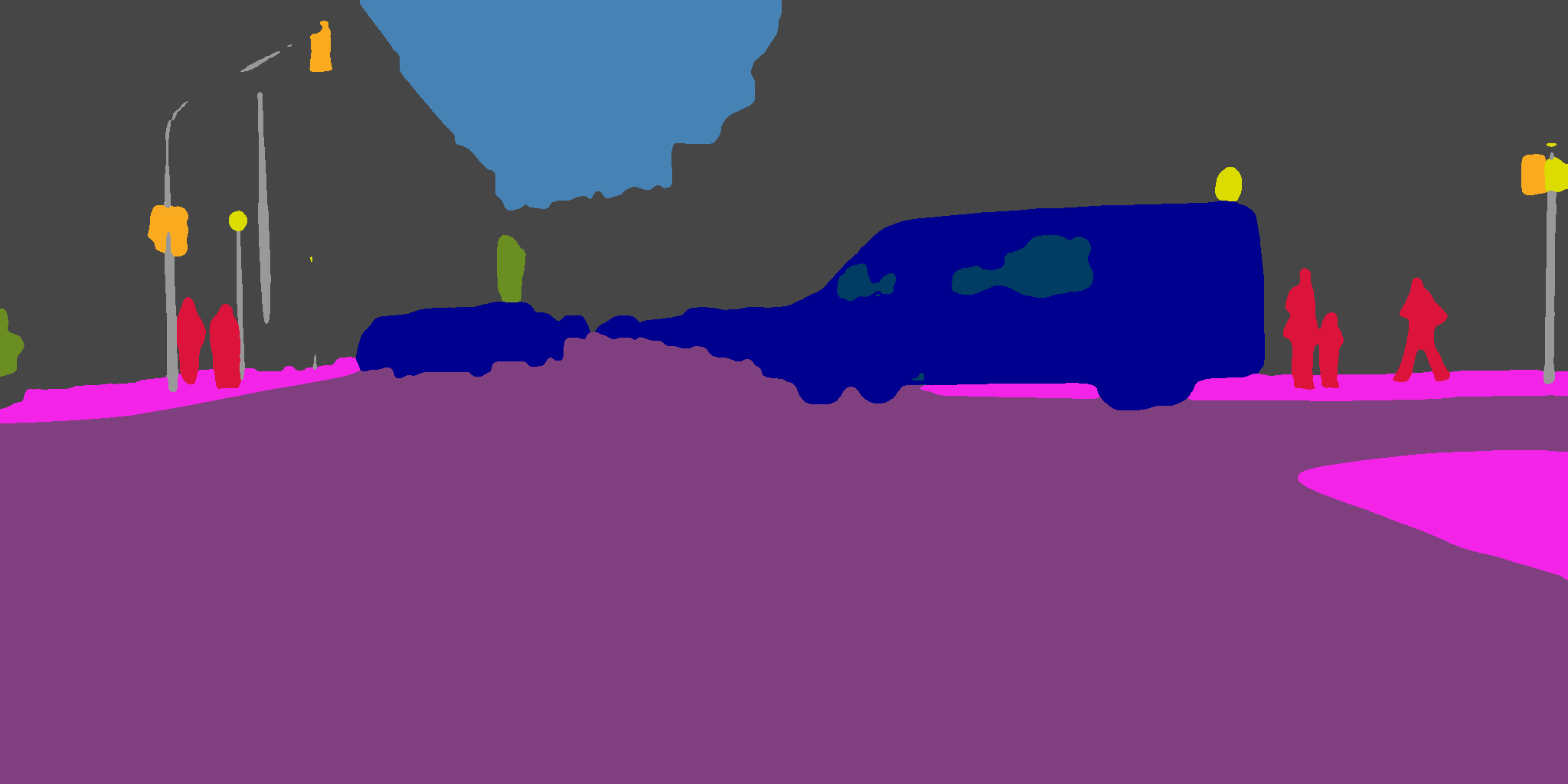}
    \end{subfigure}\hfill
    \begin{subfigure}[b]{0.3\linewidth}
    \includegraphics[width=\textwidth]{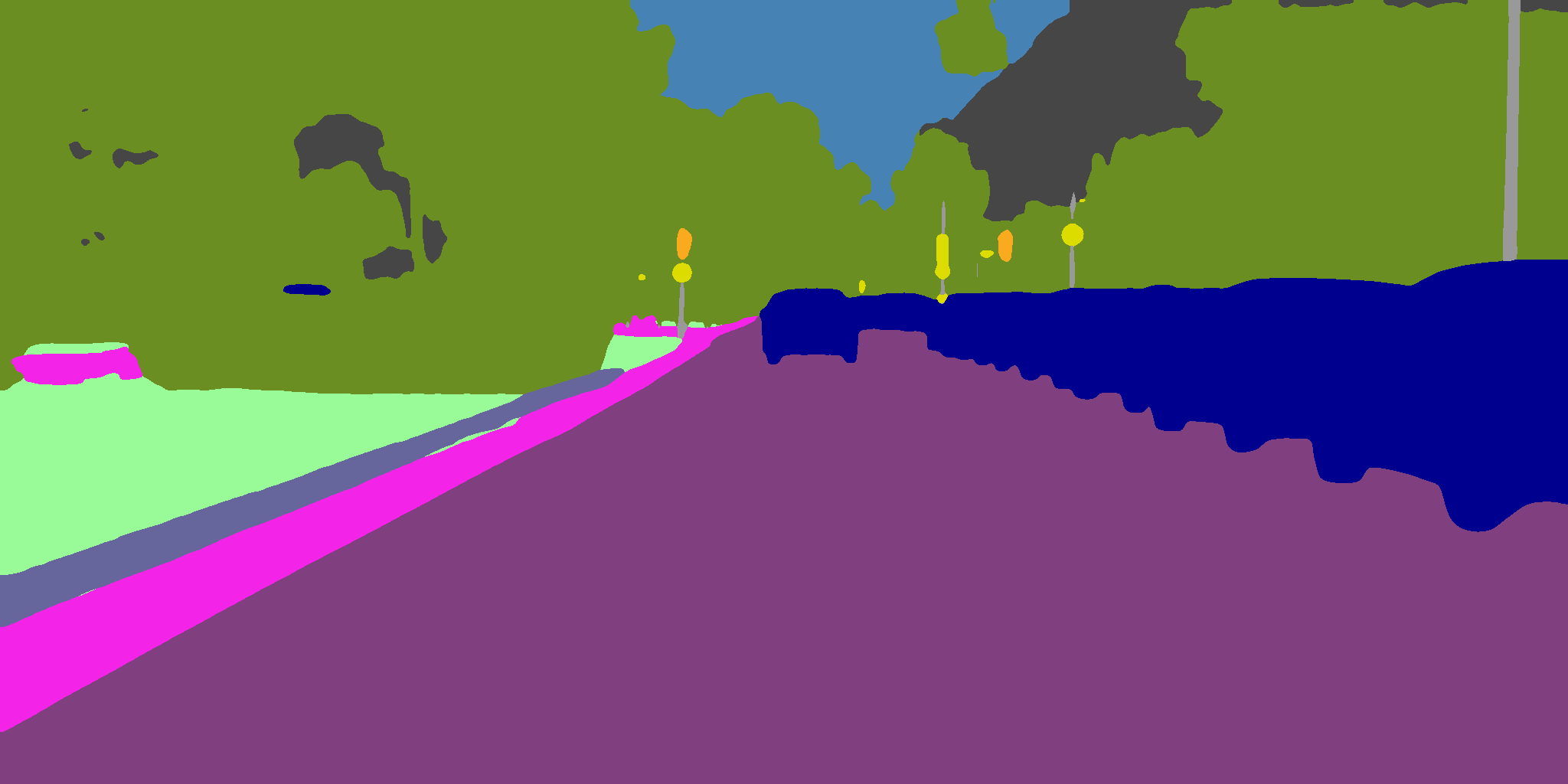}
    \end{subfigure}\hfill
    \begin{subfigure}[b]{0.3\linewidth}
    \includegraphics[width=\textwidth]{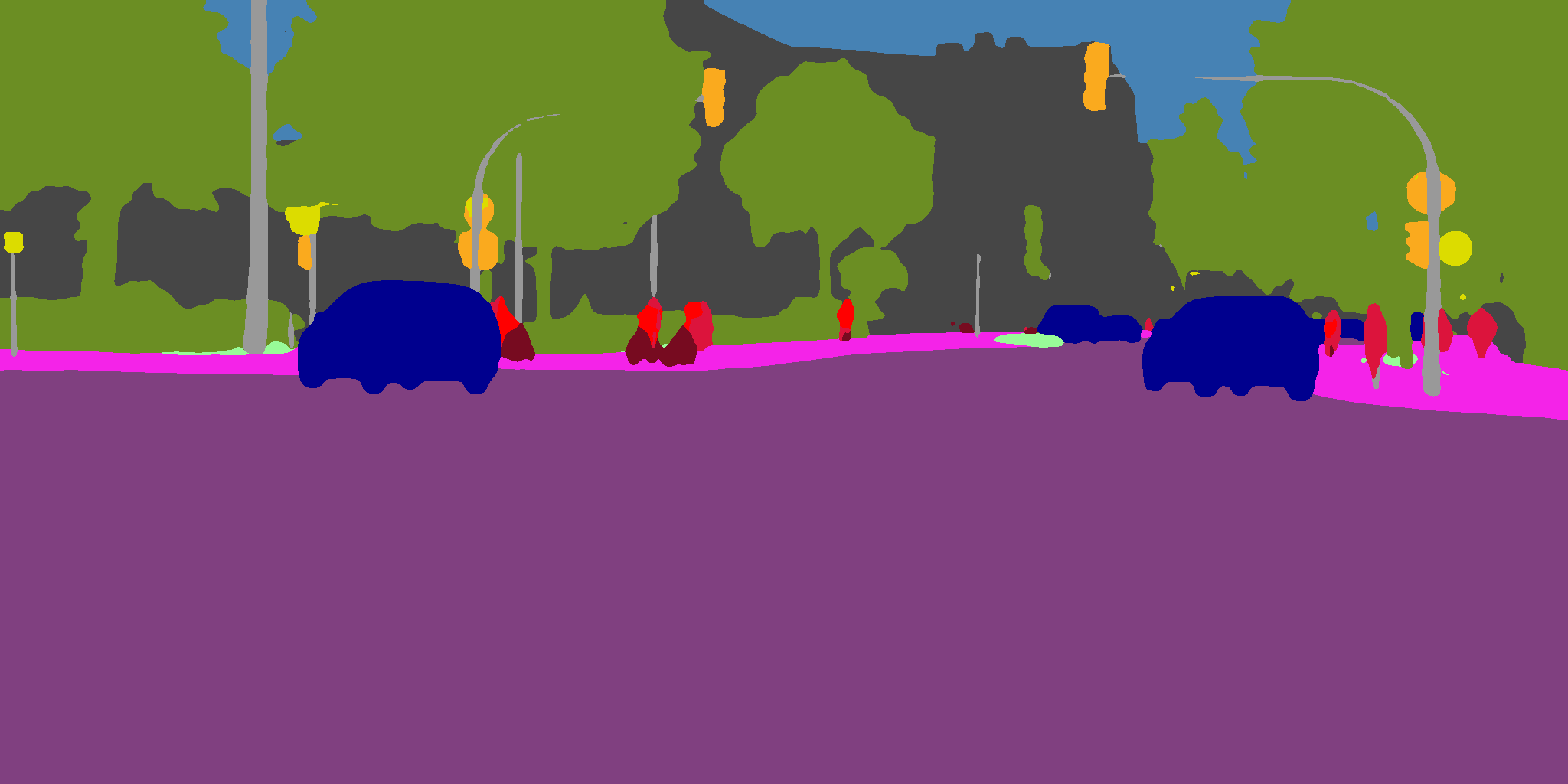}
    \end{subfigure}\hfill
    \caption{Qualitative comparison of our models trained on different input resolutions. First and second rows represent the colour and label images. The third, fourth and fifth rows illustrate the outputs of HRNetV2 models trained with input resolutions of $\frac{1}{2},\frac{1}{4},\frac{1}{8}$ respectively.}
    \label{fig:Qualitative}
\end{figure}

\paragraph{Performance comparison}
\begin{table*}[tp]
\centering
    \resizebox{\textwidth}{!}{%
    \begin{tabular}{c c c c c c c c c c c c c c c c c c c c c c c }
        \toprule
         &Down-sampling & &\multicolumn{19}{ c }{IoU per class} &  \tabularnewline\cline{4-22}
         
         %\rotatebox[origin=c]{90}{Architecture}
         & Strategy & Resolution &\rotatebox[origin=c]{90}{\textit{road} } & \rotatebox[origin=c]{90}{\textit{sidewalk} } & \rotatebox[origin=c]{90}{\textit{building} } & \rotatebox[origin=c]{90}{\textit{wall}} &  \rotatebox[origin=c]{90}{\textit{fence}}& \rotatebox[origin=c]{90}{\textit{pole} } &\rotatebox[origin=c]{90}{\textit{light}}&\rotatebox[origin=c]{90}{\textit{sign} } & \rotatebox[origin=c]{90}{\textit{vegetation}} & \rotatebox[origin=c]{90}{\textit{terrain}}& \rotatebox[origin=c]{90}{\textit{sky}}& \rotatebox[origin=c]{90}{\textit{pedestrian}}& \rotatebox[origin=c]{90}{\textit{rider}} & \rotatebox[origin=c]{90}{\textit{car}}& \rotatebox[origin=c]{90}{\textit{truck}} & \rotatebox[origin=c]{90}{\textit{bus}}& \rotatebox[origin=c]{90}{\textit{train}} & \rotatebox[origin=c]{90}{\textit{motorcycle}} & \rotatebox[origin=c]{90}{\textit{bicycle}}& mIoU\tabularnewline\midrule
         \multirow{6}{*}{\rotatebox[origin=c]{90}{DeeplabV3+}}&Nearest Neighbour & $1/8$& 81.5&43.9&72.4&6.5&7.8&45.2&21.2&36.4&81.8&26.9&42.7&65.6&16.3&74.5&15.3&13.0&9.7&19.2&51.4&38.5\tabularnewline
         &Ours & $1/8$& 98.0& 82.8&91.0& 54.0& 50.1& 56.0& 56.2& 70.4& 91.6& 58.8& 94.3& 75.8& 48.3& 94.0&77.2& 83.9& 75.5& 50.7& 72.2& 72.7\tabularnewline
         %\cline{2-23}
         &Nearest Neighbour & $1/4$&  93.7& 76.7&87.13&33.9&43.3&62.1&61.9&70.8&89.4&51.2&89.8&79.2&53.7&92.0&41.6&69.9&47.4&47.9&73.1&66.6\tabularnewline
         &Ours & $1/4$& 98.3& 85.9&92.7& 58.0& 60.2& 64.3& 67.3& 76.8& 92.6& 64.1& 94.9& 80.9& 60.3& 95.2&80.4& 86.8& 78.4& 62.8& 76.2& 77.7\tabularnewline
         %\cline{2-23}
         &Nearest Neighbour & $1/2$& 97.5& 82.2&91.5& 43.6& 58.6& 65.9& 69.8& 77.9& 92.1& 58.5& 94.2& 82.1& 62.4& 94.8&71.7& 87.5& 75.5& 64.1& 76.4& 76.1\tabularnewline
         &Ours & $1/2$& 98.3& 86.8&93.4& 61.6& 66.4& 70.0& 73.5& 81.5& 93.2& 64.2& 95.4& 84.8& 68.3& 95.4&84.8& 91.7& 83.5& 71.2& 79.9& 81.3\tabularnewline\midrule
         %&Nearest Neighbour & $1/16$& 20.0&10.13&32.9&1.4&0.3&19.5&3.4&3.7&38.1&3.4&14.4&7.5&1.8&17.4&1.2&2.8&0.1&0.8&3.1&9.58\tabularnewline
         %&Proposed & $1/16$& 95.8&65.57&82.4&31.2&20.2&18.5&9.4&31.9&84.3&48.6&83.5&51.2&12.1&83.1&40.9&36.8&36.22&7.2&42.03&46.4\tabularnewline
         \multirow{6}{*}{\rotatebox[origin=c]{90}{HRNet}}&Nearest Neighbour & $1/8$& 88.5&59.5&82.9&18.6&14.7&50.7&56.5&73.1&85.0&36.0&63.4&65.6&32.1&88.0&31.5&57.3&37.4&31.7&54.9&54.1\tabularnewline
        &Ours & $1/8$& 98.3& 85.6&91.6& 54.5& 52.5& 58.1& 65.8& 75.5& 91.9& 66.8& 93.6& 78.0& 55.8& 94.8&82.3& 85.0& 65.8& 61.5& 73.2& 75.3\tabularnewline
        %\cline{2-23}
         &Nearest Neighbour & $1/4$&  96.7&78.6&91.0&42.1&50.4&62.7&68.9&79.0&91.3&59.9&92.9&77.0&54.0&92.1&60.1&72.5&68.7&43.3&72.7&71.3\tabularnewline
         &Ours & $1/4$& 98.7& 88.3&93.8& 61.4& 72.2& 70.1& 74.1& 82.6& 93.2& 68.7& 95.4& 83.4& 66.0& 96.0&89.5& 93.0 & 82.7& 68.8& 79.3& 82.0\tabularnewline
         %\cline{2-23}
         &Nearest Neighbour& $1/2$& 98.3& 86.9&93.2& 66.9& 66.4& 68.8& 76.3& 83.3& 92.3& 59.3& 92.8& 83.6& 65.8& 95.6&84.5& 88.7 & 78.4& 63.4& 79.3& 80.5\tabularnewline
         &Ours & $1/2$& 98.8& 89.7&94.4& 70.0& 74.4& 73.5& 78.1& 85.4& 93.5& 70.5& 95.8& 85.9& 71.0& 96.3&86.2& 93.6& 87.9& 76.6& 81.5& 84.4\tabularnewline\bottomrule
    \end{tabular}}
    \caption{Per-class and overall performance comparison for the Cityscapes validation set. Two architectures have been trained (DeepLabV3+ and HRNetV2) with the label down-sampling (Nearest Neighbour) and the proposed one using soft-labels. The batch size employed are of 20, 12 and 3 for the input resolutions of $1/8$, $1/4$ and $1/2$ respectively. }
    \label{tab:class_perf}
\end{table*}
Table \ref{tab:class_perf} compiles the per-class performances for all the trained models. Both the global (mIoU) and all the per-class performances (IoU) of the models trained using the proposed down-sampling strategy are higher than those of their respective baselines. It is noteworthy that the performance of the $\frac{1}{4}$ models using the proposed down-sampling strategy is higher than those of their respective $\frac{1}{2}$ baseline models. Gains in performance are larger for thin classes (\textit{fence, pole, traffic light}) and low-represented classes (e.g., \textit{wall}). Figure \ref{fig:Qualitative} depicts qualitative results for different training input resolutions. Note how as the training input decreases its resolution, so does the model's capability of discerning further finer details. %In the supplementary material, we include an statistical analysis on the correlation between the nature of the classes and the gains in performance.   %Finally, we would like to remark how our proposal is capable of over-performing the respective reported results for each architecture employing a scale of only $\frac{1}{4}$.

\paragraph{Memory requirements}

\begin{table}[tbp]
    \centering
    %\resizebox{\linewidth}{!}{.8%
    \begin{tabular}{c l l c c l}
         %Architecture
         & Resolution & TTT & Memory&BS & mIoU  \\\midrule
         & $128\times256$ & 1 & 12GB&20&72.7 \\%(0.84\/batch)
         DeeplabV3+& $256\times512$ & 3.4 & 24GB &12& 77.7\\
         & $512\times1024$ & 15.5 & 24GB & 3&81.3 \\
         & $1024\times2048$ & 15.5 & 48GB &3& 81.5 \\\midrule
         & $128\times256$ & 1 & 12GB& 20& 75.3\\%(2.59\/batch)
         HRNetV2& $256\times512$ & 3.4 & 24GB & 12&82.0\\
         %& $256\times512$ & 1.7 & 24GB & 82.01\\
         & $512\times1024$ & 7 & 24GB & 3&84.4\\
         & $1024\times2048$ & 7 & 48GB &3& 84.7\\
         %HRNetv2\cite{YuanCW20} &$512\times1024$&1.75 & $(24\times 4)$GB & 81.6 \\\hline
         \bottomrule
         %DeeplabV3+\cite{rotabulo2017place}& $512\times1024$ & 1.55 &$(24\times 10)$GB & 75.82\\\hline
         %Espnetv2 (Ours)& $128\times256$ & ? & 12GB& \\
         %& $256\times512$ & ? & 12GB & \\
         %& $512\times1024$ & ? & 24GB & \\\hdashline
         %Espnetv2\cite{mehta2018espnetv2} &$256\times512$& ?& 12GB & 54.6\\
         %Espnetv2\cite{mehta2018espnetv2} &$512\times1024$& ?& 12GB & 66.2\\\hline
    \end{tabular}%}
    \caption{Relative total training time (TTT) comparison of the image resolutions used for training. mIoU is computed on the Cityscapes validation set. Batch size (BS) mainly drives the total training time, as the processing time for one image is relatively comparable to employing the maximum BS possible due to parallelization. The dataset's original resolution is $1024\times 2048$.}
    \label{tab:memory}
\end{table}
The highest performance of semantic segmentation models is usually obtained by training with large resolutions, which leads to high computational resources in terms of memory consumption and computation time. In Table \ref{tab:memory}, we present the time and memory requirements to train the models reported in Table \ref{tab:class_perf}. The advantages of the proposed strategy is clearly seen as resources are significantly decreased. For instance, for the $\frac{1}{8}$ resolution models, the training is over 7 times faster than that of the $\frac{1}{2}$ resolution ones, while maintaining more than 90\% of the mIoU performance at the $\frac{1}{2}$ resolution. Note that for full-resolution, we needed to employ twice the memory required for $\frac{1}{2}$ resolution and present very similar performance. This suggests that our soft-labels are capable of conserving all information on $\frac{1}{2}$ half-resolution images.

\paragraph{On the advantages of using multi-class pixels}
\label{sec:KLVSCE}

\begin{table}[tbp]
    \centering
    %\resizebox{.8\linewidth}{!}{%
    \begin{tabular}{c c c c c}
         &Resolution & Down-sampling  &MC& mIoU \\\midrule
         &$128 \times 256$ & Nearest Neighbour&$\times$ &38.5\\ 
         \multirow{6}{*}{\rotatebox[origin=c]{90}{Deeplab}}&& Ours& $\times$ & 70.0\\
%         & Proposed& CE&\checkmark & 72.0\\
         && Ours& \checkmark & \textbf{72.7}
         \\%\hdashline
         &$256 \times 512$ & Nearest Neighbour& $\times$&66.6\\
         && Ours&$\times$ & 73.7\\
%         & Proposed& CE&\checkmark & 81.5\\
         && Ours& \checkmark & \textbf{77.7}\\% \hdashline
         &$512 \times 1024$ & Nearest Neighbour& $\times$&76.1\\ 
         && Ours& $\times$ & 80.1\\
%         & Proposed& CE&\checkmark & 84.2\\
         && Ours& \checkmark & \textbf{81.5}\\ \midrule
         &$128 \times 256$ & Nearest Neighbour&$\times$ &54.1\\ 
         \multirow{6}{*}{\rotatebox[origin=c]{90}{HRNet}}&& Ours& $\times$ & \textbf{76.0}\\
%         & Proposed& CE&\checkmark & 72.0\\
         && Ours& \checkmark & 75.3\\ %\hdashline
         &$256 \times 512$ & Nearest Neighbour& $\times$&71.3\\
         && Ours&$\times$ & 80.7\\
%         & Proposed& CE&\checkmark & 81.5\\
         && Ours& \checkmark & \textbf{82.0}\\% \hdashline
         &$512 \times 1024$ & Nearest Neighbour& $\times$&80.5\\ 
         && Ours& $\times$ & 82.1\\
%         & Proposed& CE&\checkmark & 84.2\\
         && Ours& \checkmark & \textbf{84.4}\\ \bottomrule
    \end{tabular}%}
    \caption{Performance comparison of the down-sampling strategy employed on the semantic maps and the employed loss for training.  Results from training on the Cityscapes validation set with a batch size of 20, 12 and 3 for the input resolutions of $128 \times 256$, $256 \times 512$ and $512 \times 1024$ respectively. MC indicates if the loss takes into account multi-class pixels. Bold indicates best results. Note that the original $1024\times 2048$ resolution is excluded, as down-sampling is unnecessary.}
    \label{tab:abl}
    
\end{table}
%\paragraph{Study on the loss}
We quantify the effect of considering multi-class pixels by only computing the loss over the single-class pixels or both the single and multi-class pixels. Note that Nearest Neighbour only generates single-class pixels. We focus only on the most competitive architecture (HRNetV2) for the following analyses. Reported results in Table \ref{tab:abl} suggest that the main benefits of the proposed strategy emerge from removing the introduced sampling noise: as the gain obtained by MC is narrower than the overall gain obtained when the down-sampling strategy is changed. However, empirical results suggest that MC pixels do have a positive impact on the model performance for bigger input resolutions. This is particularly noteworthy at the highest resolution ($\frac{1}{2}$ resolution), where our soft-labels preserve all label information.

\begin{figure}[]
    \centering
    \begin{subfigure}[b]{.5\linewidth}
    \includegraphics[width=\linewidth]{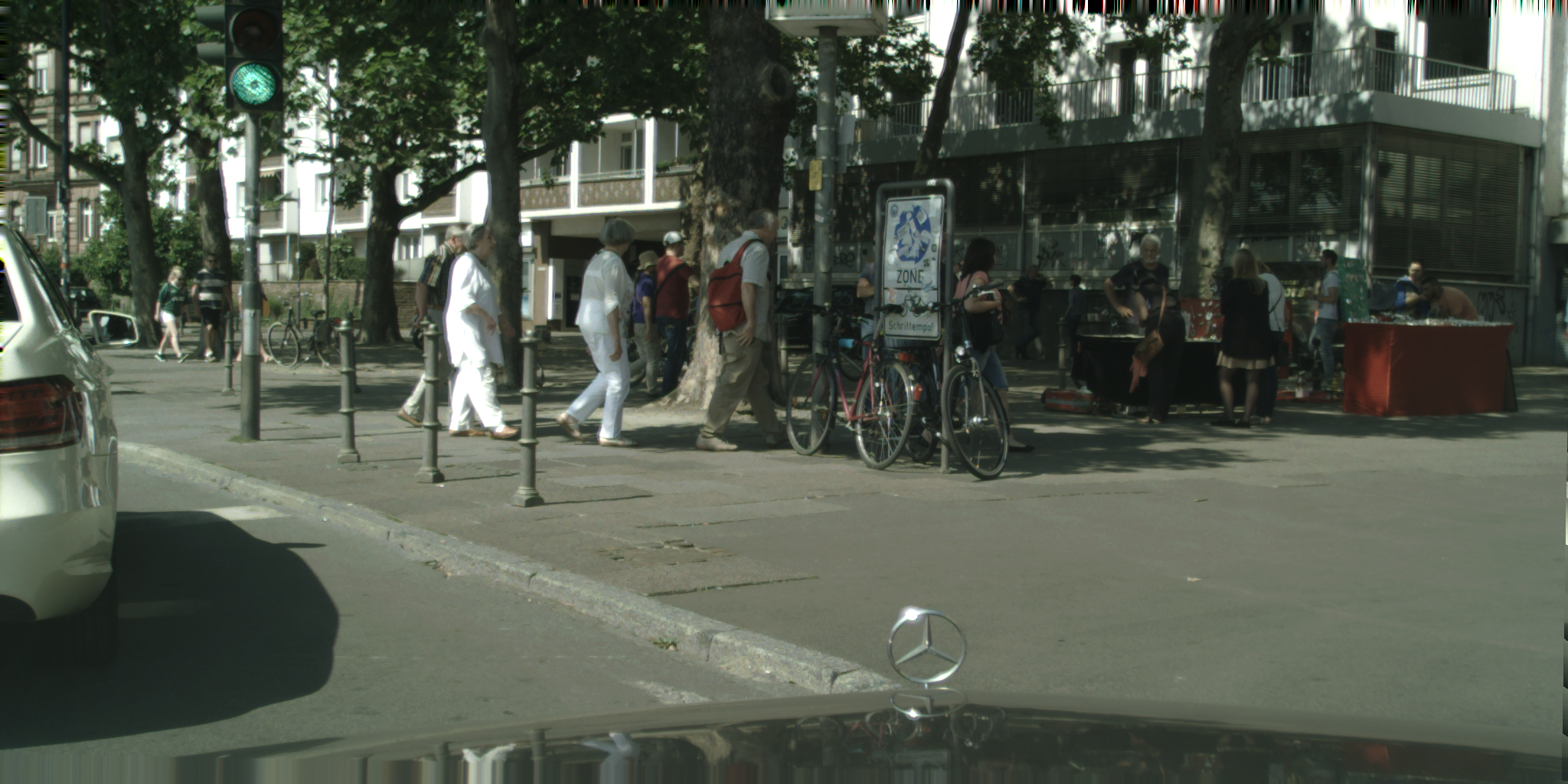}
    \end{subfigure}\hfill
    \begin{subfigure}[b]{.5\linewidth}
    \includegraphics[width=\linewidth]{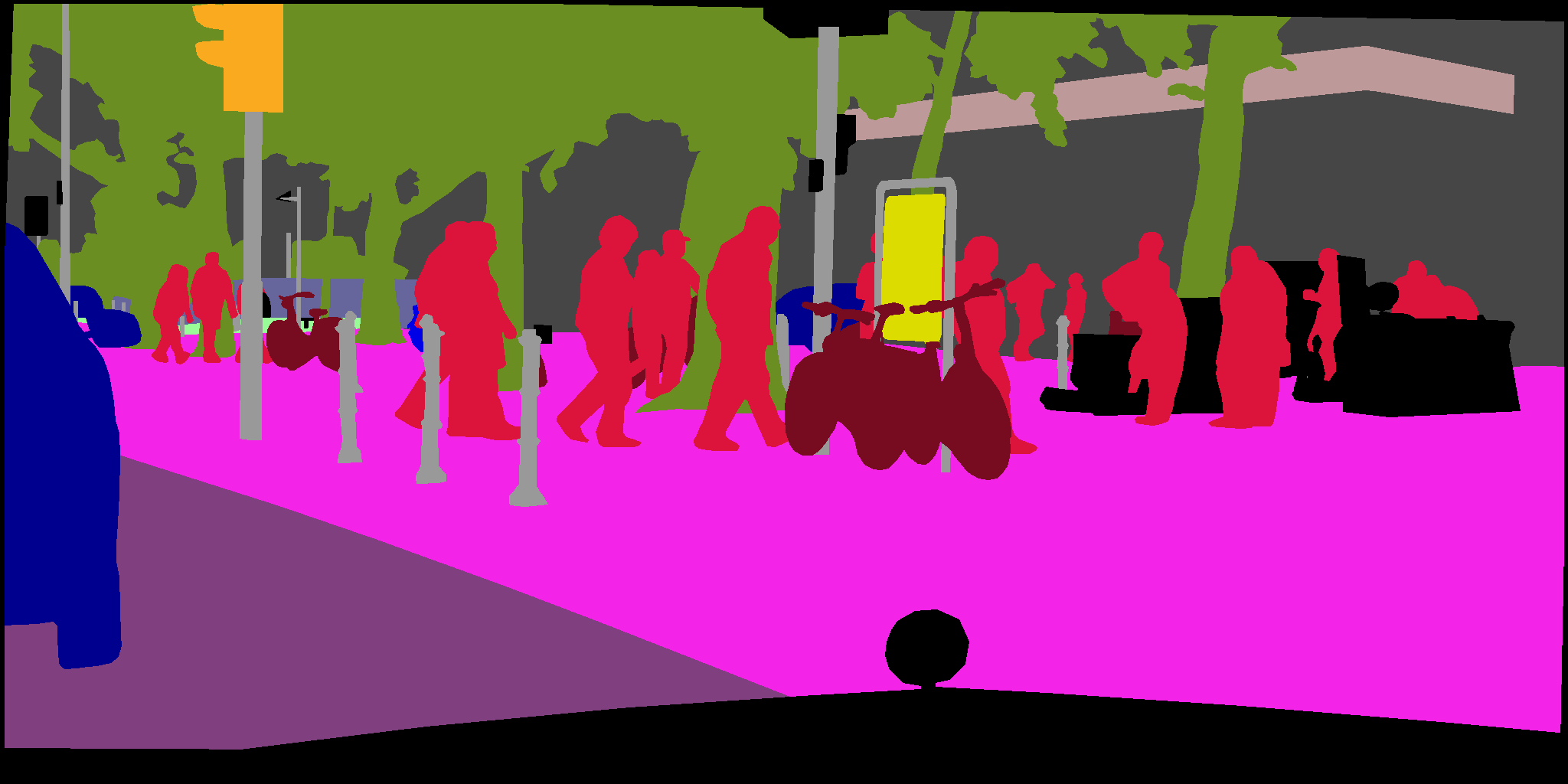}
    \end{subfigure}\\
    \begin{subfigure}[b]{0.5\linewidth}
    \includegraphics[width=\linewidth]{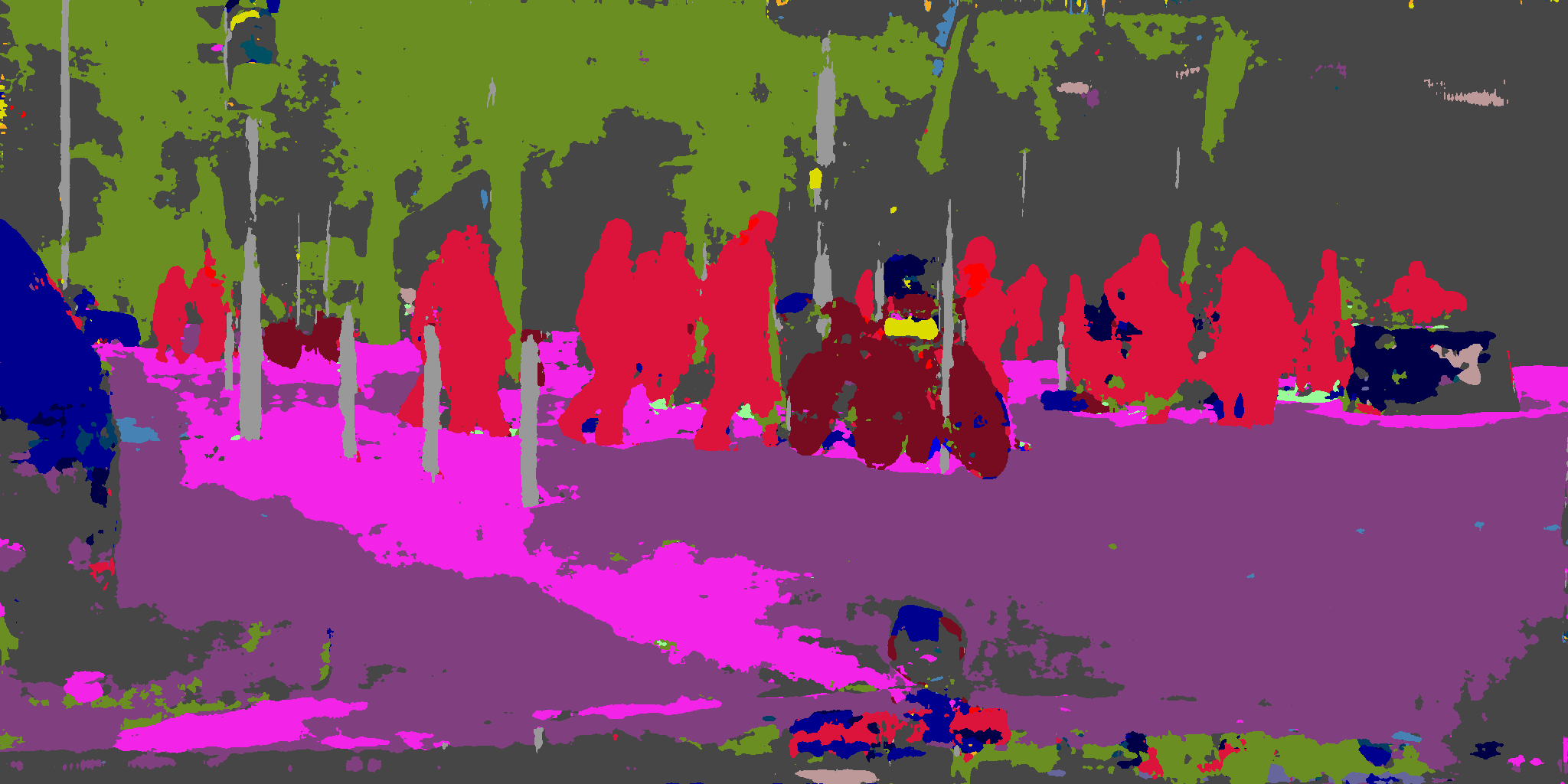}
    \end{subfigure}\hfill
    \begin{subfigure}[b]{0.5\linewidth}
    \includegraphics[width=\linewidth]{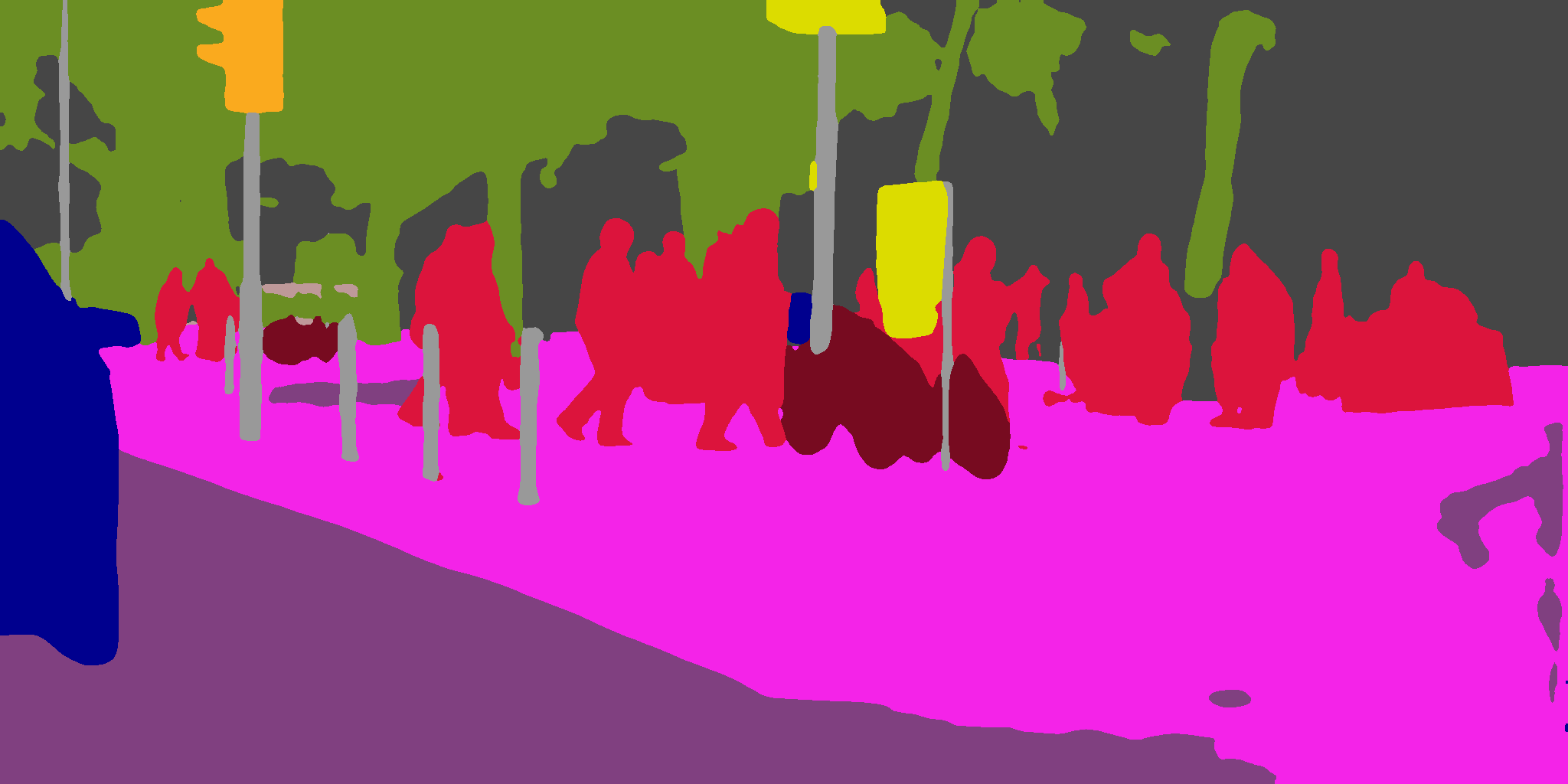}
    \end{subfigure}\\
    \begin{subfigure}[b]{0.5\linewidth}
    \includegraphics[width=\linewidth]{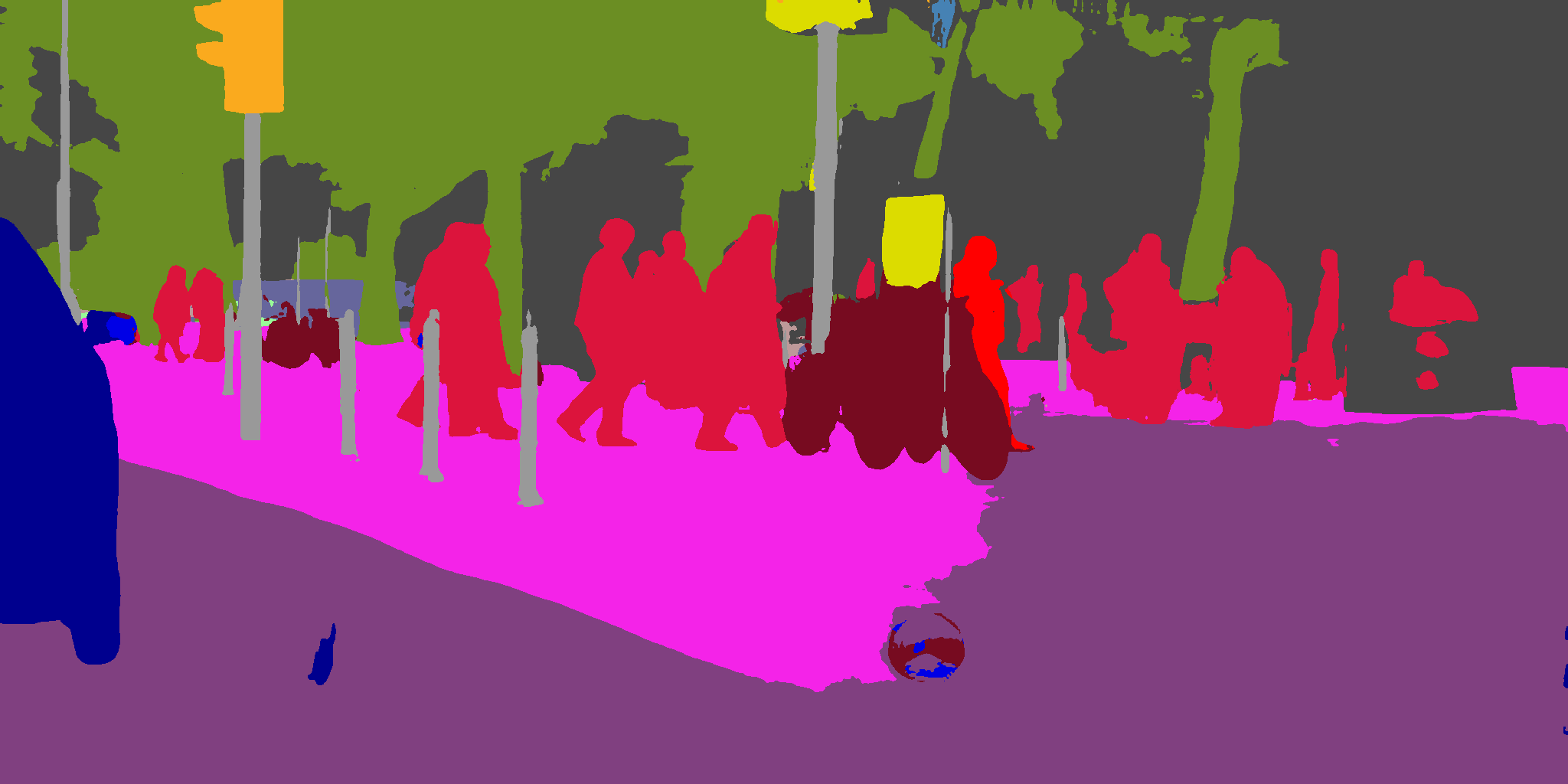}
    \end{subfigure}\hfill
    \begin{subfigure}[b]{0.5\linewidth}
    \includegraphics[width=\linewidth]{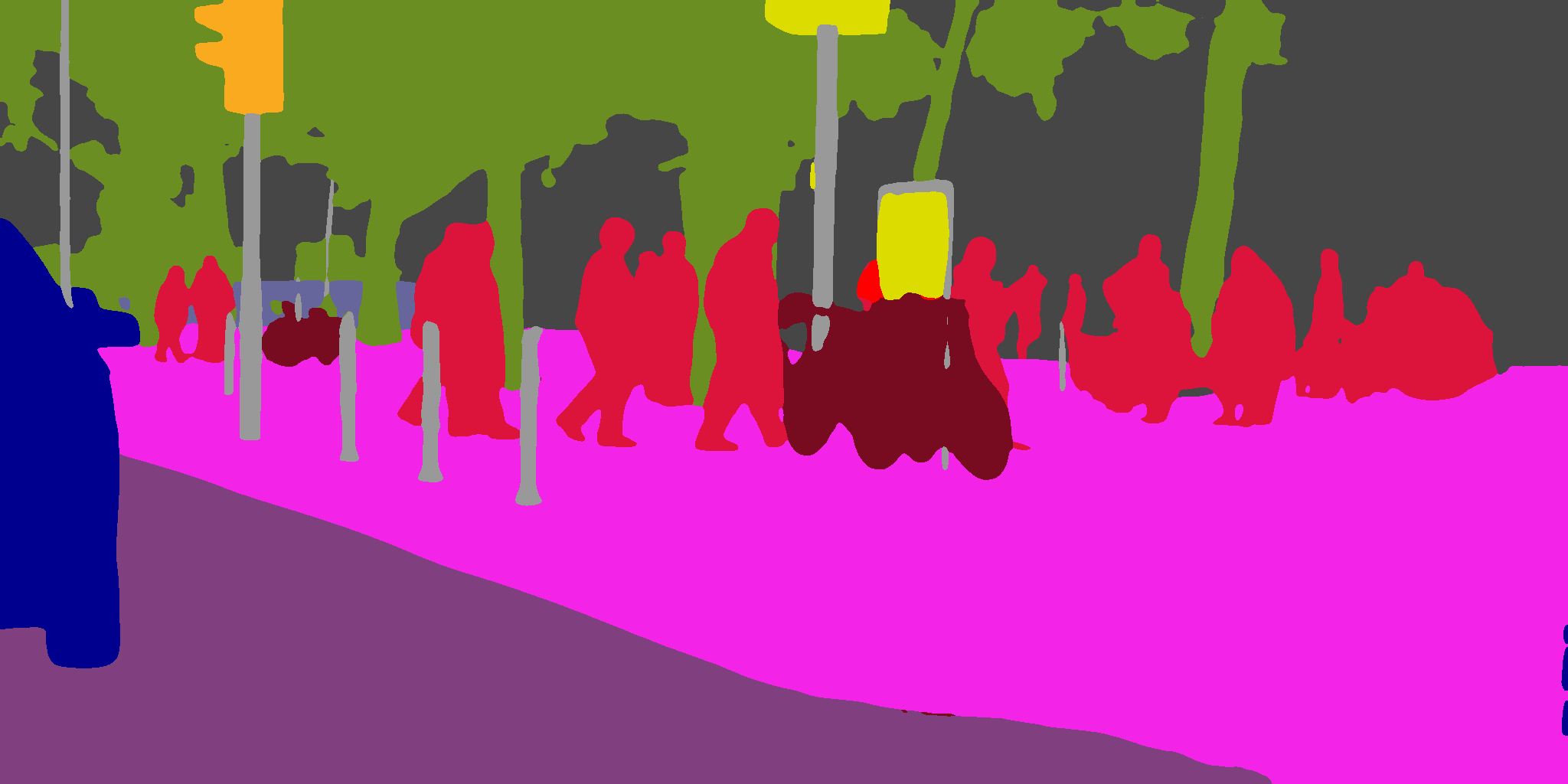}
    \end{subfigure}\\
    \caption{Qualitative comparison of our models to the ones trained employing Nearest Neighbour resize on the Cityscapes validation set with an HRNet architecture. First row presents the colour and label images respectively. Second and third rows represent the prediction of the NN model (left) and our model (right) both trained with 12GB of memory and an input resolution of $128\times256$ and $256\times512$ respectively.}
    \label{fig:baseline}
    %\vspace{-5mm}
\end{figure}

Figure \ref{fig:baseline} compares models trained employing Nearest Neighbour with models trained using the proposed soft-label strategy for different input resolutions. As depicted for the $\frac{1}{8}$ resolution (i.e., $128\times256$), employing Nearest Neighbour seems to preclude the learning of some classes due to information lost and the number of artefacts introduced in the down-sampling.

%These loss setups are used to drive training processes on the Cityscapes training images at different resolutions using the proposed strategy or the Nearest Neighbour one. %is the impact of multi-class pixels in the loss slightly benefits the final performance for smaller input resolutions. This is specially remarkable for the lowest resolution explored, where the model trained completely ignoring the multi-class pixels outperforms the other models.

\paragraph{The paired down-sampling alleviates  batch size constraints in semantic segmentation}
\begin{figure}
    \centering
    \includegraphics[trim=5 5 5 5,clip,width=.9\linewidth]{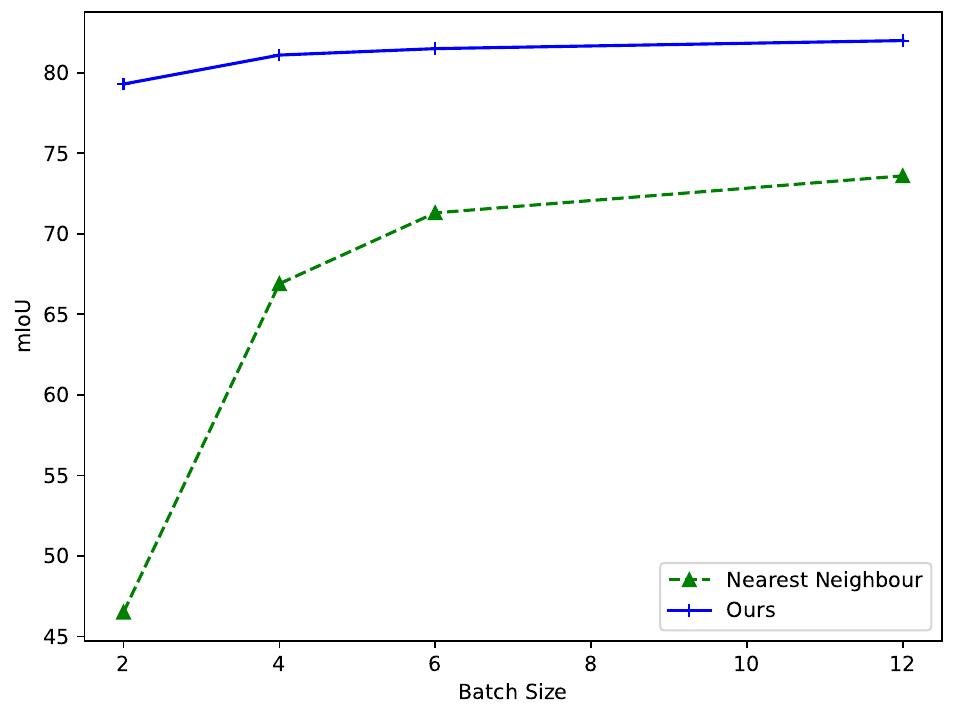}
    \caption{Comparative analysis on the impact of different batch sizes for training an HRNetV2 architecture on the Cityscapes dataset, with an input training size of $256\times 512$ to the final performance of the model measured on mIoU.}
    \label{fig:BS}
\end{figure}
% \begin{table}[tp]
% \centering
% \resizebox{.75\linewidth}{!}{%
%     \begin{tabular}{c c c}
%     %\toprule
%         BS & Sampling& mIoU \\\hline
%         2& Nearest Neighbour& 46.5\\
%         2&bilinear& 79.3\\
%         4&Nearest Neighbour&66.9\\
%         4&bilinear& 81.1\\
%         6&Nearest Neighbour& 71.3\\
%         6&bilinear& 81.5\\
%         12&Nearest Neighbour& 73.6\\
%         12&bilinear&82.0\\\bottomrule
%        %$512\times1024$ &3&bilinear & 84.4\\
%        %  &6&bilinear& 84.8\\\bottomrule
%     \end{tabular}}
%     \caption{Comparative  analysis on the impact of different batch sizes for training an HRNetV2 architecture with an input training size of $256\times 512$ to the final performance of the model measured on mIoU.}
%     \label{tab:BS}
    
% \end{table}
It is commonly agreed that large batch sizes are crucial for training semantic segmentation models \cite{zhou2019semantic, liang2019winter}, where performance can increase up to 300\% by increasing the batch size \cite{liang2019winter}. We argue that this is to compensate for the noise introduced into the training from the Nearest Neighbour down-sampling. Figure \ref{fig:BS} compares batch sizes and the performance obtained, which clearly illustrates how our models are significantly less dependant on the batch size compared to the ones trained with Nearest Neighbour. 
\paragraph{On the importance on sampling selection}
\begin{table}[]
    \centering
    %\resizebox{.8\linewidth}{!}{%
        \begin{tabular}{c l l}
         Colour  & Label & mIoU \\\midrule
         bilinear & Nearest Neighbour& 71.3\\
         bicubic & Nearest Neighbour& 71.4\\
         bilinear & ours-bicubic& 80.6\\
         bilinear & ours-bilinear & 81.5\\
         bicubic & ours-bilinear & 81.3\\
         bicubic & ours-bicubic & 82.6\\\bottomrule
    \end{tabular}%}
    
    \caption{Ablation study on the strategies for sampling colour and labels for HRNetV2 trained on the Cityscapes dataset ($\frac{1}{4}$ of original's resolution, $256\times 512$) with a batch size of 6.}
    \label{tab:Sampling}
\end{table}
Our proposal enables any combination of samplings for the colour and label training images. In Table \ref{tab:Sampling}, we present the results of an study on commonly employed sampling procedures for colour and label images. Notably, the utilization of Nearest Neighbour sampling significantly diminishes the model's overall performance. We attribute this decline to the emergence of perceptible hard edges in the sampled images.

At our selected resolution ($256\times 512$) for this experiment, which is $\frac{1}{4}$ of the original resolution, we observe that employing bicubic interpolation for both colour and label images yields the most favourable outcomes, as it employs more pixels for the down-sampling. Therefore, generating richer soft-labels. However, for the purpose of benchmarking against state-of-the-art methods, along with the rest of the paper, we exclusively compare Nearest Neighbour against bilinear down-sampling. This decision aligns with the prevailing practice in the field, where bilinear colour sampling predominates. Moreover, adopting bicubic sampling would introduce additional computational overhead, potentially skewing the fairness of our comparisons.

\subsection{On conserving all the information}
\label{sec:Corr}

\begin{table*}[]
\centering
    \resizebox{\textwidth}{!}{%
    \begin{tabular}{c c c c c c c c c c c c c c c c c c c c c}
        \toprule
         Down-sampling & &\multicolumn{18}{ c }{Per-class prior difference $D_{d,c}$} &  \tabularnewline\cline{3-21}
         
         %\rotatebox[origin=c]{90}{Architecture}
         Strategy & Resolution &\rotatebox[origin=c]{90}{\textit{road} } & \rotatebox[origin=c]{90}{\textit{sidewalk} } & \rotatebox[origin=c]{90}{\textit{building} } & \rotatebox[origin=c]{90}{\textit{wall}} &  \rotatebox[origin=c]{90}{\textit{fence}}& \rotatebox[origin=c]{90}{\textit{pole} } &\rotatebox[origin=c]{90}{\textit{light}}&\rotatebox[origin=c]{90}{\textit{sign} } & \rotatebox[origin=c]{90}{\textit{vegetation}} & \rotatebox[origin=c]{90}{\textit{terrain}}& \rotatebox[origin=c]{90}{\textit{sky}}& \rotatebox[origin=c]{90}{\textit{pedestrian}}& \rotatebox[origin=c]{90}{\textit{rider}} & \rotatebox[origin=c]{90}{\textit{car}}& \rotatebox[origin=c]{90}{\textit{truck}} & \rotatebox[origin=c]{90}{\textit{bus}}& \rotatebox[origin=c]{90}{\textit{train}} & \rotatebox[origin=c]{90}{\textit{motorcycle}} & \rotatebox[origin=c]{90}{\textit{bicycle}}\tabularnewline\midrule
         Nearest Neighbour& 1/8 & -8.52&-2.53&-15.84&-0.25&-0.34&-0.47&-0.10&-0.21&-11.08&-0.50&-6.40&-0.22&-0.01&-2.36&-0.06&-0.07&-0.08&-0.04&-0.13\\
         Ours & 1/8 & -8.25 & -2.77 & -14.48 & -0.27 & -0.35 & -0.44 & -0.09 & -0.18 & -9.82 & -0.48 & -5.47 & -0.23 & -0.02 & -2.37& -0.06 & -0.06 & -0.06 & -0.04 & -0.12 \\
         Nearest Neighbour& 1/4 & 3.93 & 1.44 & 4.89 & 0.16 & 0.21 & 0.19 & 0.02 & 0.05 & 2.78 & 0.29 & 1.00 & 0.13 & 0.12 & 1.40 & 0.04 & 0.02 & 0.04 & 0.02 & 0.07 \\
         Ours & 1/4 & 1.31 & 0.63 & 4.27 & 0.07 & 0.08 & 0.12 & 0.02 & 0.06 & 2.93 & 0.08 & 1.80 & 0.05 & 0.01 & 0.49 & 0.02 & 0.02 & 0.01 & 0.01 & 0.02 \\
         Nearest Neighbour& 1/2 & -1.26 & -0.50 & 0.33 & -0.05 & -0.07 & -0.01 & 0.01 & 0.01 & 0.01 & 0.65 &-0.09& 0.09 & -0.04& -0.00 & -0.45 & -0.01 & -0.01 & -0.01& -0.02 \\
         Ours & 1/2 & 0 & 0 & 0 & 0 & 0 & 0 & 0 & 0 & 0 & 0 & 0 & 0 & 0 & 0 & 0 & 0 & 0 & 0 & 0 \\\bottomrule
        \end{tabular}}
        
        \caption{Prior discrepancy $D_{d,c}$ of each down-sampled version of the Cityscapes dataset with respect to the full-resolution priors. For visualisation purposes, the values are multiplied by $10^4$. The closer to zero, the better the similarity of the results for down-sampling strategy $d$ with the original full-resolution priors.}
        \label{tab:priors}
\end{table*}
\begin{figure*}

    \centering
    \begin{subfigure}[b]{\linewidth}
    \centering
        \includegraphics[width=.9\linewidth,fbox]{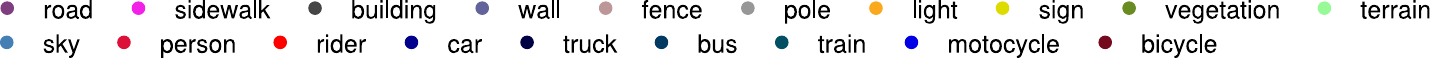}
    \end{subfigure}\\\vspace{1mm}
    \begin{subfigure}[b]{0.32\linewidth}
        \includegraphics[width=\linewidth]{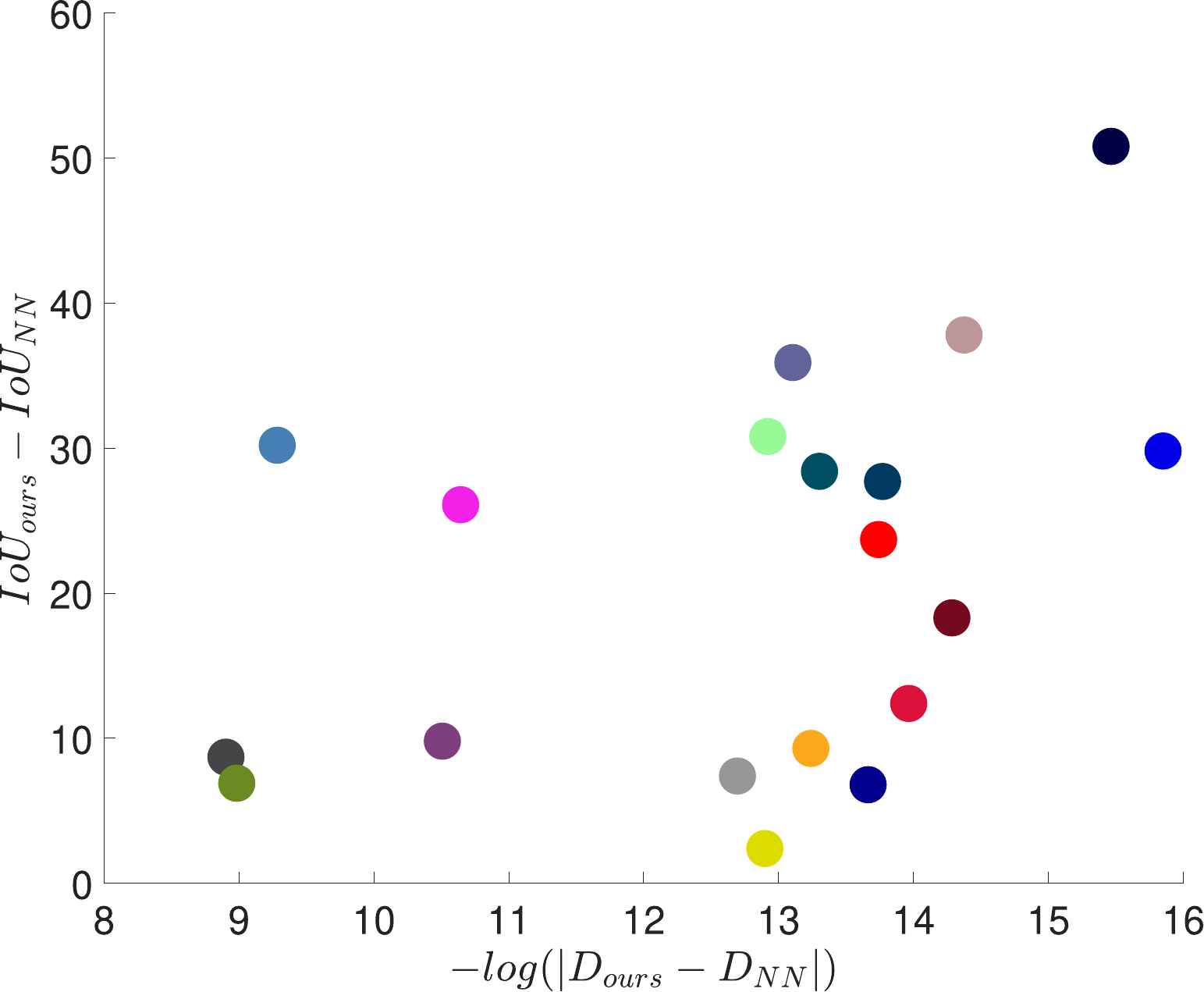}
        \caption{1/8 scale}
    \end{subfigure}\hfill
    \begin{subfigure}[b]{0.32\linewidth}
        \includegraphics[width=\linewidth]{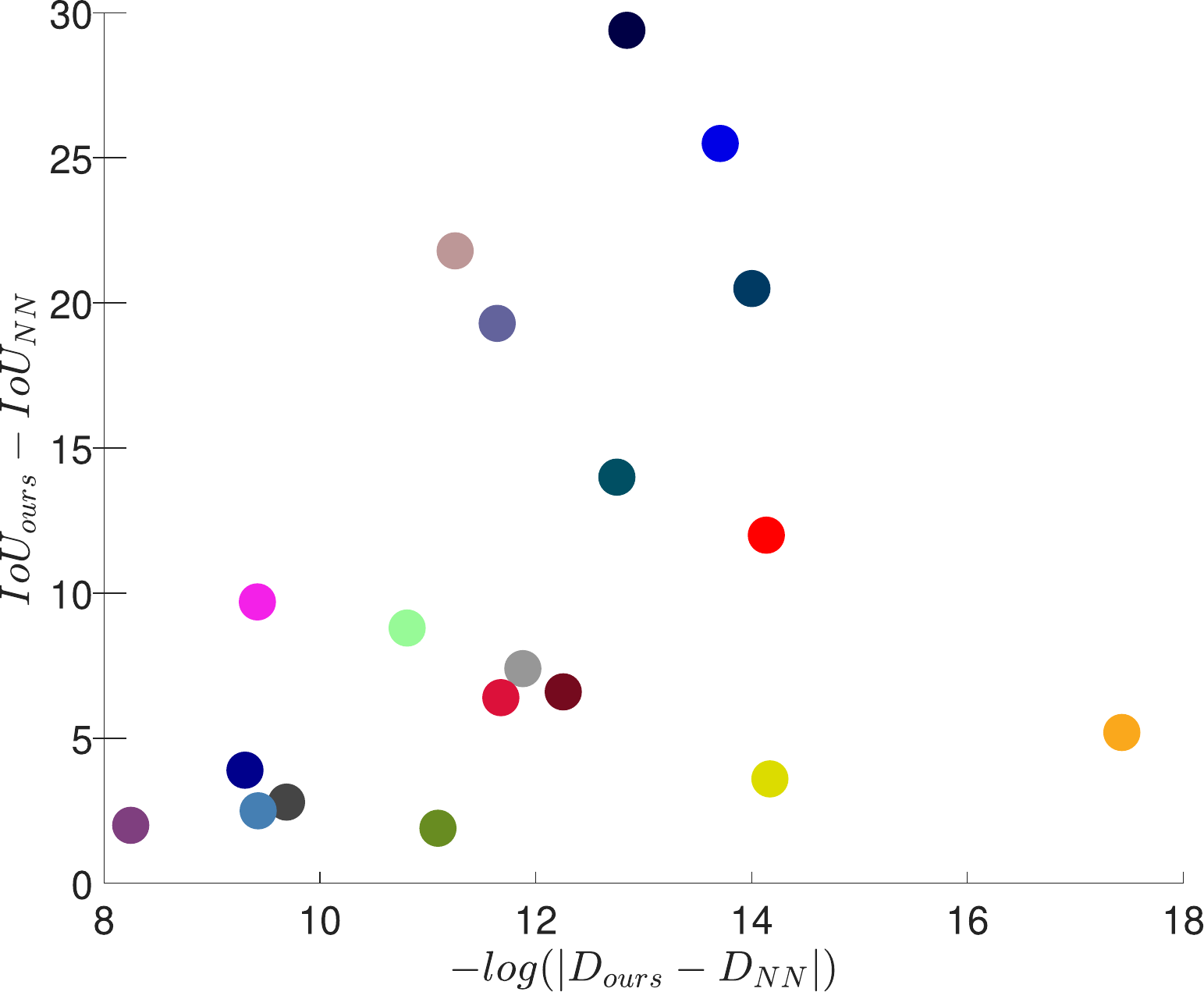}
        \caption{1/4 scale}
    \end{subfigure}\hfill
    \begin{subfigure}[b]{0.32\linewidth}
        \includegraphics[width=\linewidth]{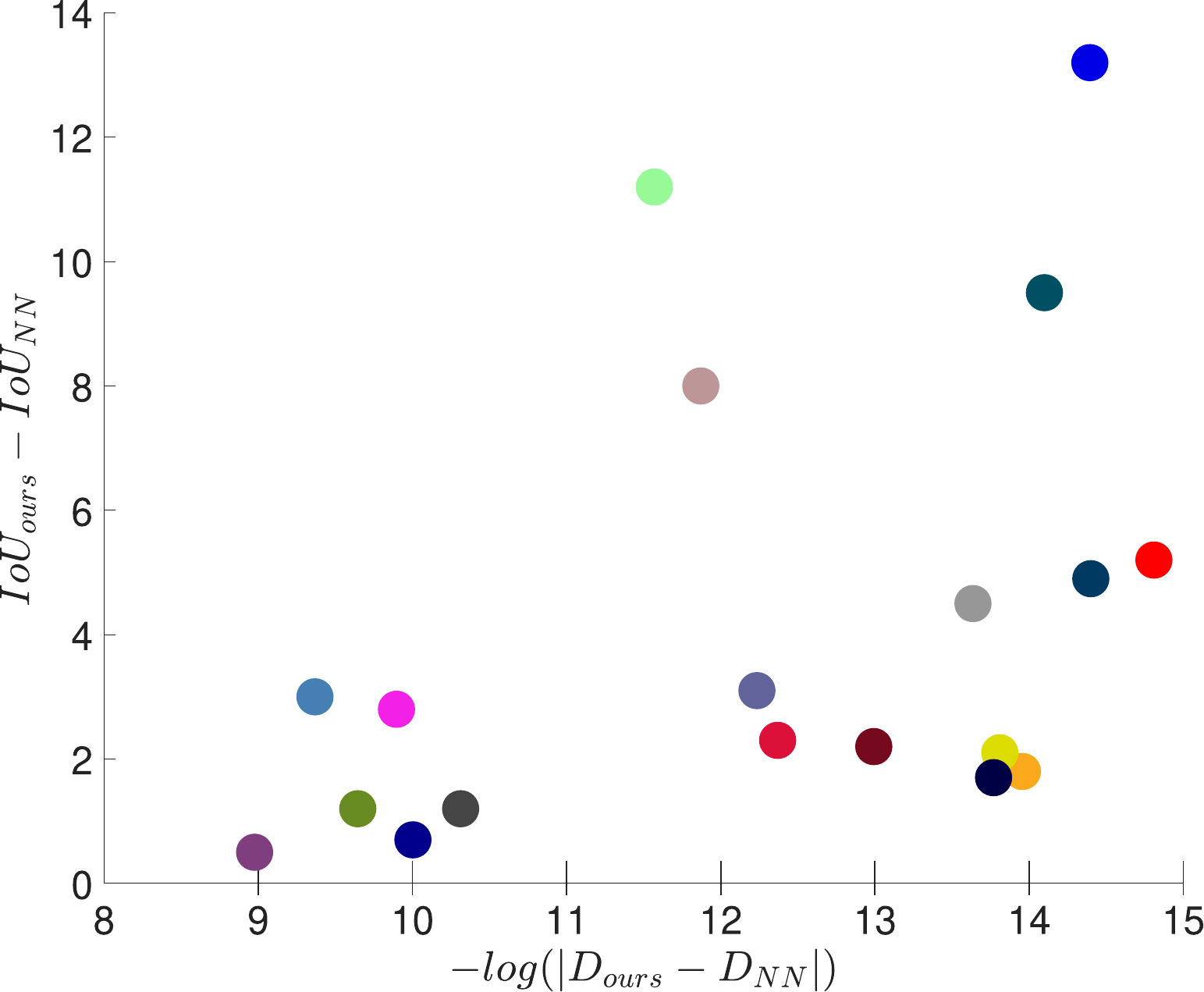}
        \caption{1/2 scale}
    \end{subfigure}
    
    \caption{Visual relationship of the per-class performance discrepancy and the per-class prior discrepancy for the analysed down-sampled resolutions on the Cityscapes dataset (1/8, 1/4 and 1/2). For visualisation purposes, we employ the logarithm as the discrepancies are in distinct scales, see Table \ref{tab:priors}. As the distance between the discrepancies is on the scale of $10^{-4}$, the logarithm yields large negative values for low discrepancies. Therefore, we employ the negative log so that large improvements in performance (top of the y-axis) correspond to large discrepancies (right side of the x-axis).}
    \label{fig:cor}
\end{figure*}

In the previous experiments, we have validated the benefits of using the soft-label encoding of information to pair the down-sampling strategy used for the label image with that used for the colour one. However, the use of soft-labels may allow to conserve all the information for any down-sample factor by considering the labels of all the pixels in the down-sampled region in the encoding. Thereby, the entropy of the class distribution in the down-sampled region and that of the resulting soft-label encoding must be the same. In our experiments, this is only the case for half-resolution down-sampling, as the bilineal sampling accounts for all the pixels in the down-sampled region. Thus, a model trained using half resolution yields equivalent performance to a full resolution one when employing equal batch size.

To assess the gain obtained when information is conserved, we analyse the distance between the original dataset and the down-sampled versions.  
For each class $c$ and down-sampling method $d$, we quantify the distance between the class priors of $d$ to the original dataset $D_{d,c}$ as the per-class probability of appearance between the original dataset $p_{original,c}$ and the down-sampled version $p_{d,c}$:
\begin{equation}
   D_{d,c}= p_{original,c} - p_{d,c}.
\end{equation}
Table \ref{tab:priors} presents the per-class priors discrepancies for each of the analysed resolutions. Note that for all resolutions our sampling presents smaller average discrepancies, especially for the 1/2 resolution, which presents the exact same priors and class distribution as the full-resolution dataset.
This discrepancy to the full-resolution priors is related to the improvements in performance. We argue that the closer a sampling method prior distribution is to the full-resolution distribution the better the performance is going to be.  Figure \ref{fig:cor} studies this relationship by measuring the difference between our sampling $D_{ours,c}$ and Nearest Neighbour sampling $D_{NN,c}$ with respect to the difference in performance $IoU_{ours,c}-IoU_{NN,c}$ (see Table \ref{tab:class_perf}).  It illustrates that classes with greatest improvement in performance, tend to present similar appearance rates to the full-resolution compared to the Nearest Neighbour down-sampled version.

From this relationship, it can be inferred that by enhancing information conservation at smaller input resolutions, the performance gap between low-resolution trained models and full-resolution ones can become even narrower. It is important to note that achieving this narrowing of the performance gap cannot be accomplished solely through the preservation of class distributions in the soft-labels. Instead, this objective also entails defining a down-sampling strategy for the colour image. This is a major challenge that needs to be explored in future work, as a soft-colour encoding would require a large input structure—although it is likely to be sparse. The size of such a structure may also impact the model parameters at its shallower layers, thus significantly increasing the memory required for training such a model.

%\subsection{The proposed down-sampling strategy exceeds state-of-the-art performance}
\subsection{The proposal exceeds state-of-the-art performance}
%\paragraph{Comparison with the state-of-the-art}
\begin{table*}[tp]
    \centering
    %\resizebox{.9\linewidth}{!}{%
    \begin{tabular}{c l l l l l l l c}
         &&&\multicolumn{3}{c}{Training Hardware}&\multicolumn{2}{c}{Training Hyperparameters}   \\
         Context method& Venue/Journal & Architecture& \#GPUs & GB/GPU &Total (GB)& Batch size & Resolution & mIoU \\ \midrule
         %&CVPR 2021\cite{m_Huynh-etal-CVPR21}&$128\times256$& $(16\times 8)$ &67.57\\
         Pyramid &ICCV 2019\cite{9008795}& U-Net& 1&12&12&32& $128\times256$ &   62.0 \\
         Pooling&CVPR 2020\cite{Wang_2020_CVPR}& Deeplab&1&12&12&12& $512\times1024$&  66.9\\
         &CVPR 2023\cite{Aakanksha_2023_CVPR} & Deeplab&2&24&48&16&$1024\times1024$&76.2\\
         &TIP 2022\cite{9745313}&Custom&4*&16*&64*&8&$769\times769$&78.5\\         
         &CVPR 2018\cite{rotabulo2017place}&Deeplab& 4& 24& 96 &12& $872\times872$  & 79.4\\
         &TIP 2023\cite{10112629}&Deeplab&8&32& 256&8&$769\times769$&76.5\\
         &TIP 2023\cite{10112629}&Deeplab&8&32&256&8&$769\times1537$&80.6\\
         &CVPR 2023\cite{Chen_2023_CVPR}&Custom& 8&48&384&16&$712\times712$&78.5\\
         %&CVPR 2020\cite{cheng2020panoptic} &Panoptic&32&32&32& $1024\times2048$  &  81.5\\
         \cdashline{2-9}
         
         %&ECCV 2018\cite{deeplabv3plus2018}&50&24&$>$16* &$513\times513$ & 79.6\\
         &Ours&Deeplab&1&12&12& 6&$256\times512$ &  77.7\\
         %&PanopticDL-Ours&1&24&3&$1024\times2048$ & 80.0\\
         &Ours&Deeplab&1&24&24& 3&$512\times1024$ &  81.3\\
         &Ours&Deeplab&1&48&48& 3&$1024\times2048$ & \textbf{81.5}\\
         &Ours&Deeplab&1&48&48& 6&$512\times1024$ & \textbf{81.5}\\\midrule
         Relational &ICCV 2023\cite{zhu2023good}&MobileNet&1*&32*&32*&4&$512\times512$& 79.7\\
         context& CVPR 2023\cite{lu2023cts}&Segmenter&1&40&40&8&$768\times768$& 76.5\\
         &CVPR 2023\cite{Aakanksha_2023_CVPR} & SegFormer&2&24&48&16&$1024\times1024$&81.1\\
         &ITSS 2022 \cite{hong2021deep}&DDRNet&4&11&44&12&$1024\times1024$&79.5\\
         & CVPR 2023\cite{xu2022pidnet}&PidNet&4&24&96&12&$1024\times1024$& 80.9\\
         &PAMI 2019\cite{9052469}&HRNet&4&32&128&12 &$512\times1024$ &  81.6\\
         &ECCV2020\cite{YuanCW20}&HRNet&4&32&128& 8 &$769\times769$ & 81.8\\
         &ICCV 2023\cite{liu2023boosting}&SegFormer&4&32&128&8&$1024\times1024$ &84.5\\
         
         &TIP 2022\cite{9745313}&Custom&4*&32*&128*&8&$769\times769$&82.6\\
         &TIP 2023\cite{10173725}&HRNet&4*&32*&128*&8&$768\times768$&83.2\\
         &CVPR 2022\cite{li2022deep}&HRNet&4&40&160& 8&$512\times1024$&  83.4\\
         %&CVPR 2023\cite{Yu_2023_CVPR}& &8&24&8&$512\times1028$&74.7\\
         &CVPR 2023\cite{Wang_2023_CVPR}& HRNet&8&32&256&16&$512\times1028$&79.9\\
         & CVPR 2023\cite{He_2023_CVPR}&Custom&8&32&256&16&$512\times1024$& 80.5\\
         &CVPR 2022\cite{zhou2022rethinking}&HRNet& 8&32&256&8& $768\times768$ & 81.1\\
         &NeurIPS 2021\cite{NEURIPS2021_950a4152}&MaskFormer &8&32&256&16&$512\times1024$ &  81.4\\
         &CVPR 2023\cite{Shi_2023_CVPR}&SwinTransformer&8&32&256&16&$512\times1028$&83.1\\
         &CVPR 2023\cite{Chen_2023_CVPR}&SwinTransformer&8&48&384&16&$712\times712$&80.1\\
         %&TMM 2023 \cite{10239324}& 8&48&32&$640\times640$&84.6\\
         &CVPR 2021\cite{borse2021inverseform}&GSNN&16&32&512& 16&$1024\times2048$ &  82.6\\\cdashline{2-9}
         %&SegNext\cite{https://doi.org/10.48550/arxiv.2209.08575}& 8&$1024\times2048$  & $(32\times 8)$GB& 83.9\\\cdashline{2-6}
         %Ours &$256\times512$ & 24GB&82.01 \\
         %&HRNetV2-Ours &1&12&20&  $128\times256$&75.3\\
         &Ours &HRNet&1&12&12&6&$256\times512$ &82.0 \\
         &Ours&HRNet&1&24&24& 3&$512\times1024$ & 84.4\\
         &Ours&HRNet&1&48&48& 3&$1024\times2048$ & 84.7\\%\hline
         &Ours&HRNet&1&48&48& 6&$512\times1024$ & \textbf{84.8}\\\bottomrule
         
         %&HRNetV2-Ours&1&48& 6&$512\times1024$ & 84.6\\\hline
    \end{tabular}%}
    \caption{Performance comparison of state-of-the-art semantic segmentation models on the Cityscapes validation set. The reported training setup considers the number of GPUs and memory per GPU employed for training. When a parameter is not explicitly reported `*', we provide an estimation given the known parameters. Bold indicates best results per context method.}
    \label{tab:SOTA_COMP_CS}
    
\end{table*}

Table \ref{tab:SOTA_COMP_CS} compares the models trained using label images down-sampled with the proposed strategy (Ours) with those for top-performing state-of-the-art methods in the Cityscapes validation set. Specifically, our best model trained in a constrained-training setup (HRNetV2) outperforms the current state-of-the-art constrained-training method \cite{zhu2023good} by a 6\%. 

\begin{figure}[tp]
    \hfill
    \begin{subfigure}[b]{0.5\linewidth}
    \includegraphics[width=\linewidth]{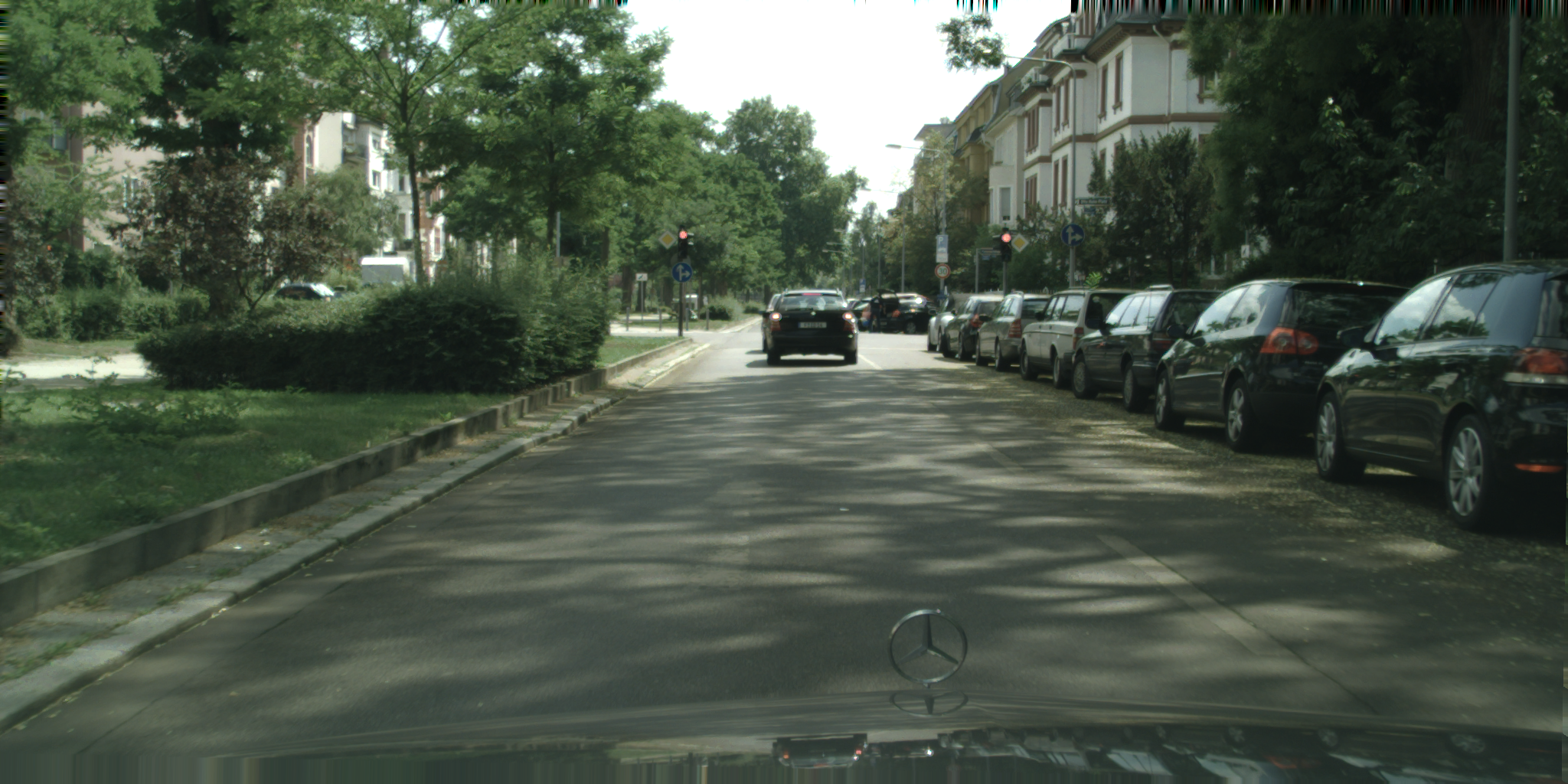}
    \end{subfigure}\hfill
    \begin{subfigure}[b]{0.5\linewidth}
    \includegraphics[width=\linewidth]{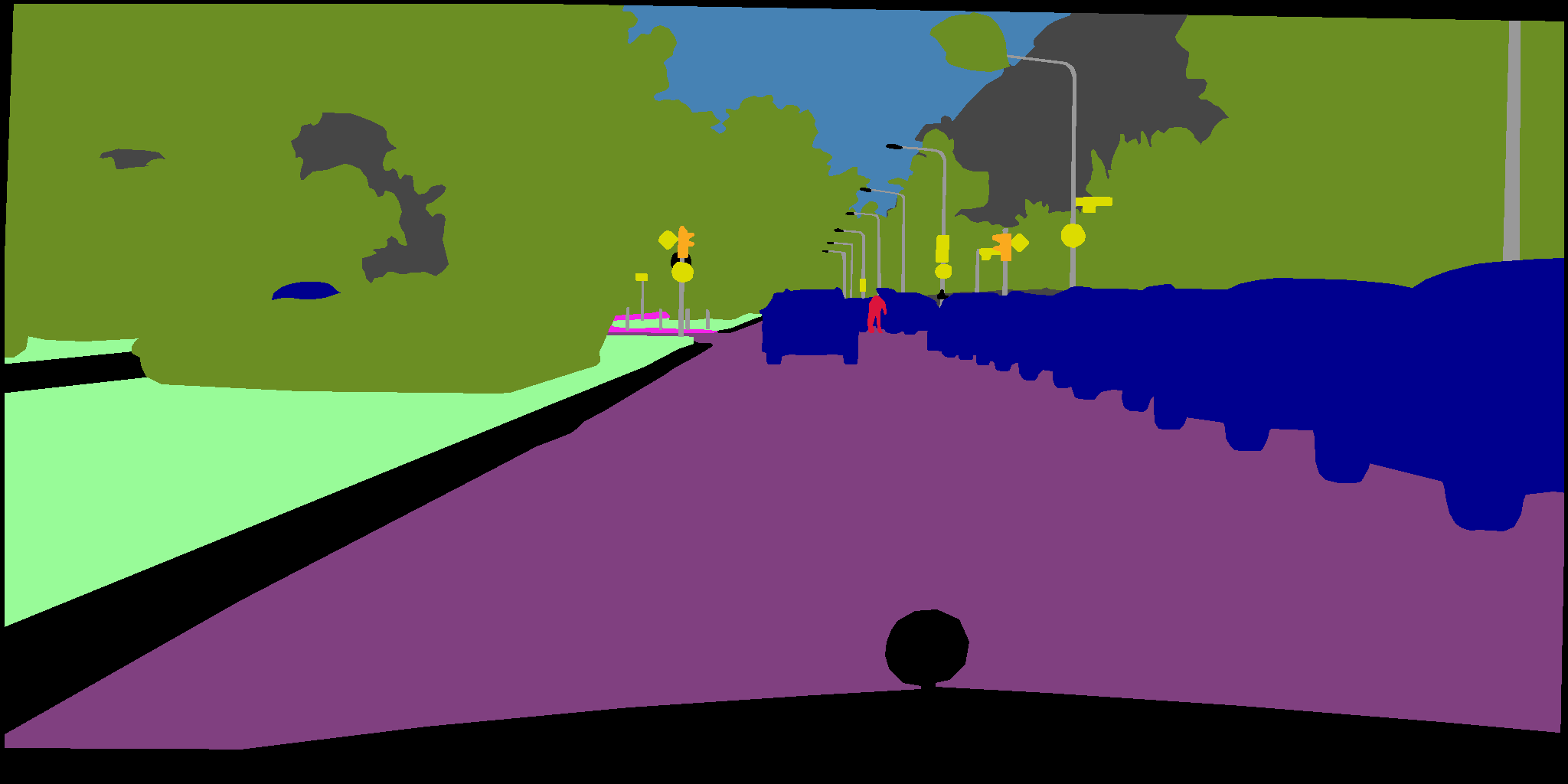}
    \end{subfigure}\hfill
    \begin{subfigure}[b]{0.5\linewidth}
    \includegraphics[width=\linewidth]{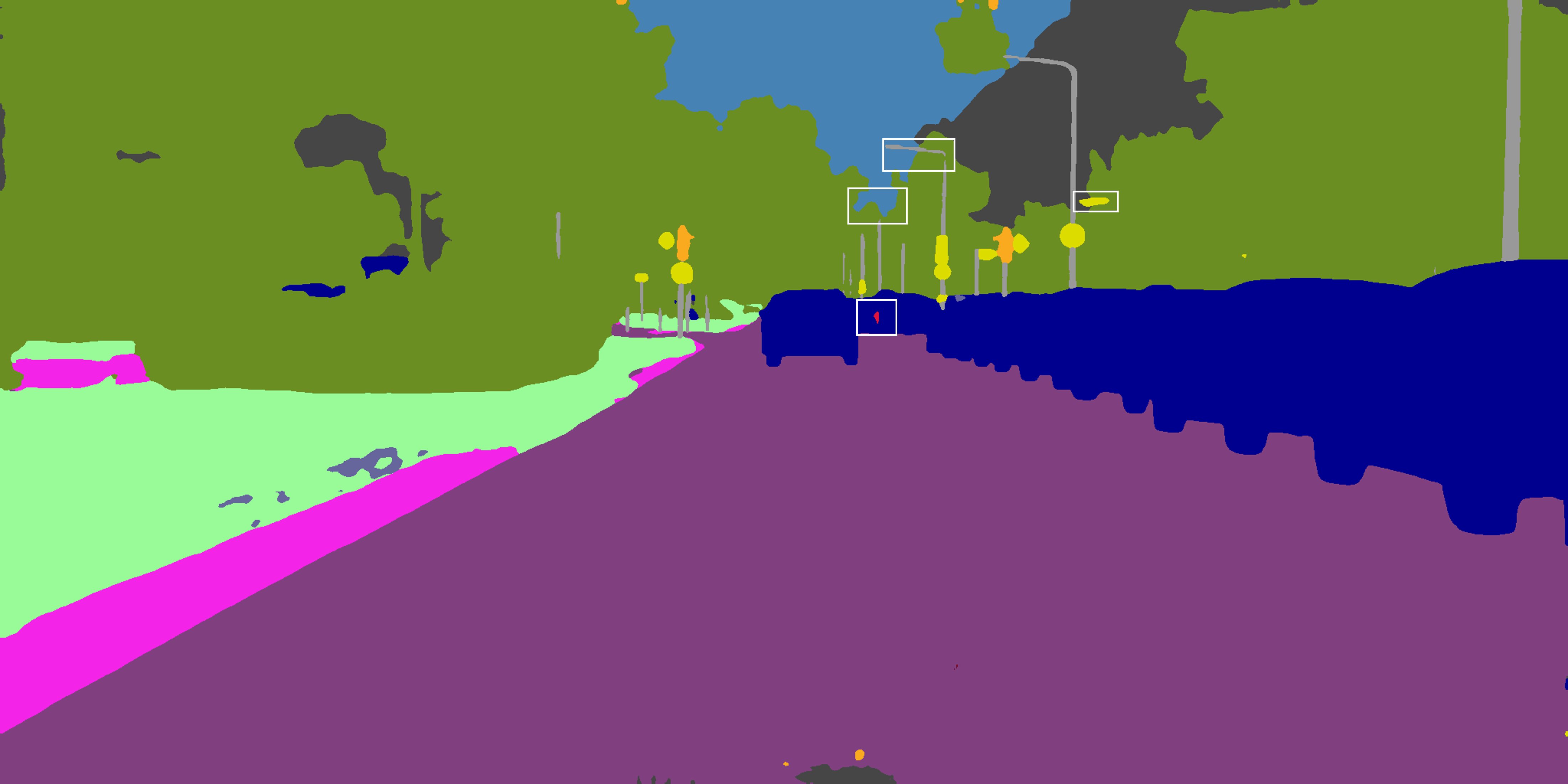}
    \end{subfigure}\hfill
    \begin{subfigure}[b]{0.5\linewidth}
    \includegraphics[width=\linewidth]{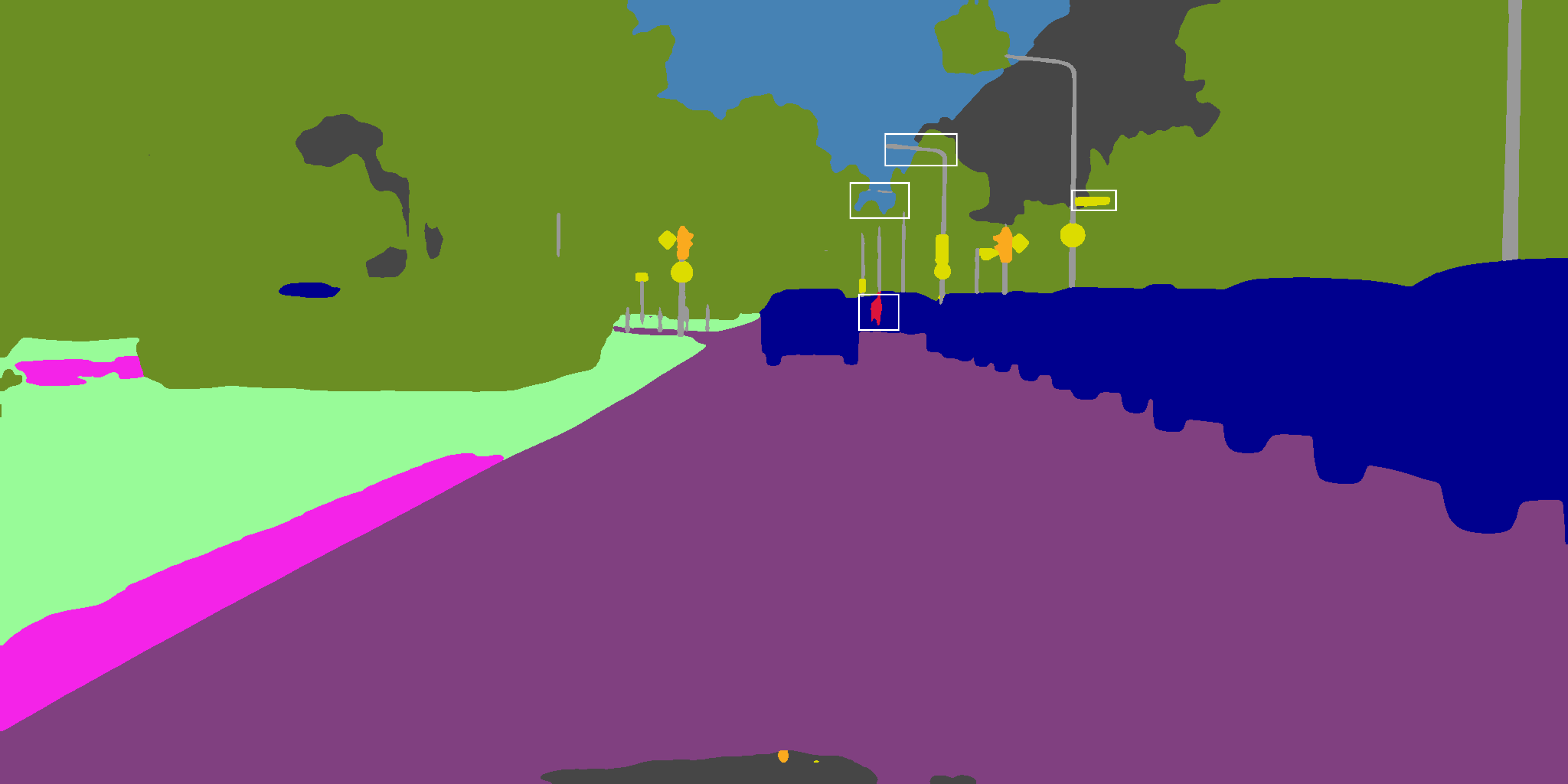}
    \end{subfigure}\hfill\hspace{2mm}\\
    \begin{subfigure}[b]{0.33\linewidth}
    \includegraphics[width=\linewidth, height=\linewidth]{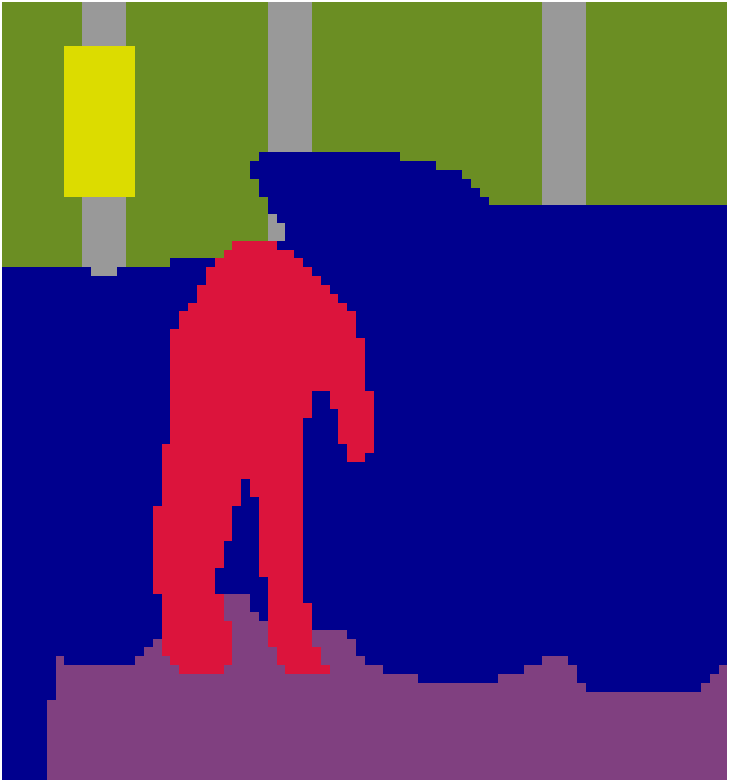}
    \end{subfigure}\hfill
    \begin{subfigure}[b]{0.33\linewidth}
    \includegraphics[width=\linewidth, height=\linewidth]{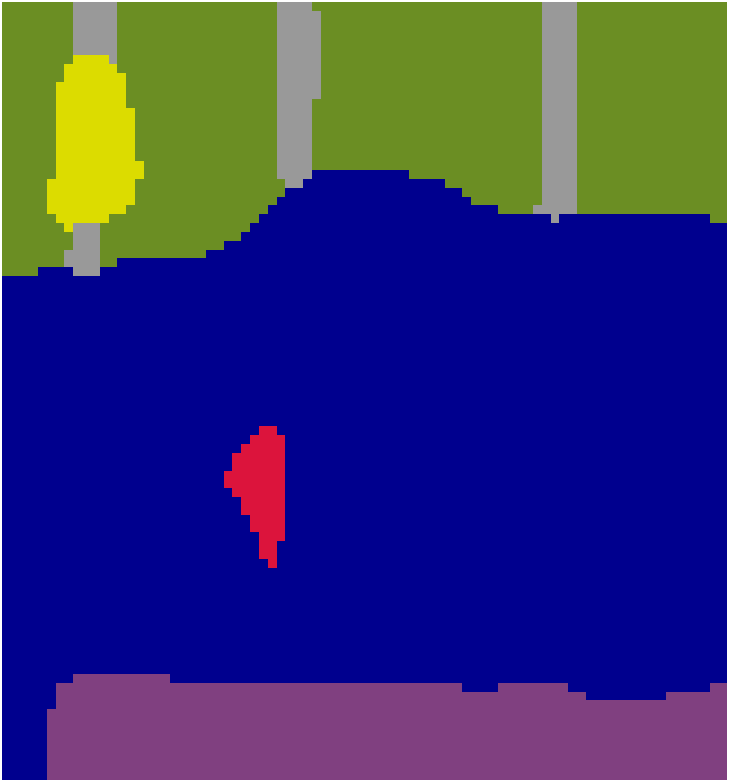}
    \end{subfigure}\hfill
    \begin{subfigure}[b]{0.33\linewidth}
    \includegraphics[width=\linewidth, height=\linewidth]{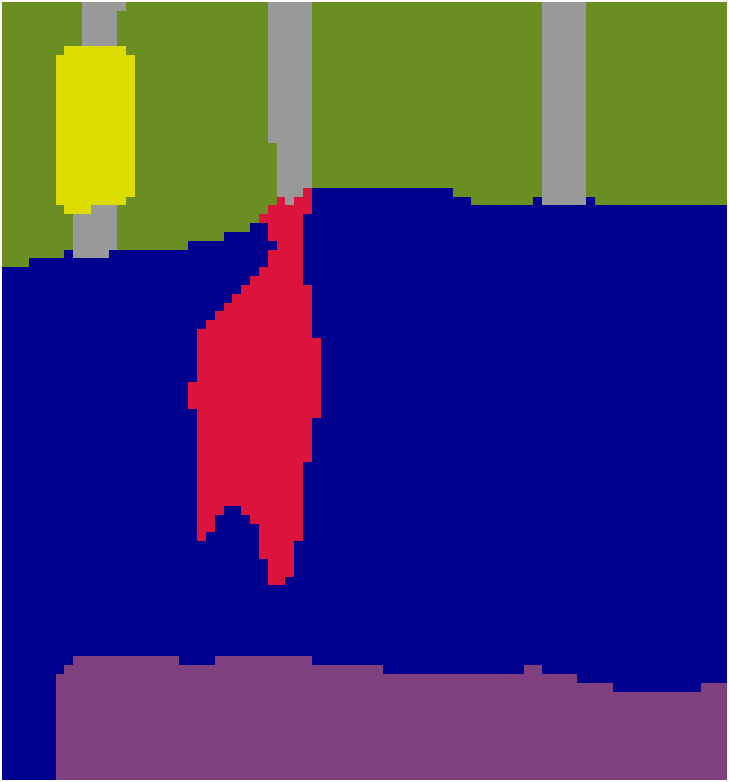}
    \end{subfigure}\\
    \begin{subfigure}[b]{0.33\linewidth}
    \includegraphics[width=\linewidth, height=\linewidth]{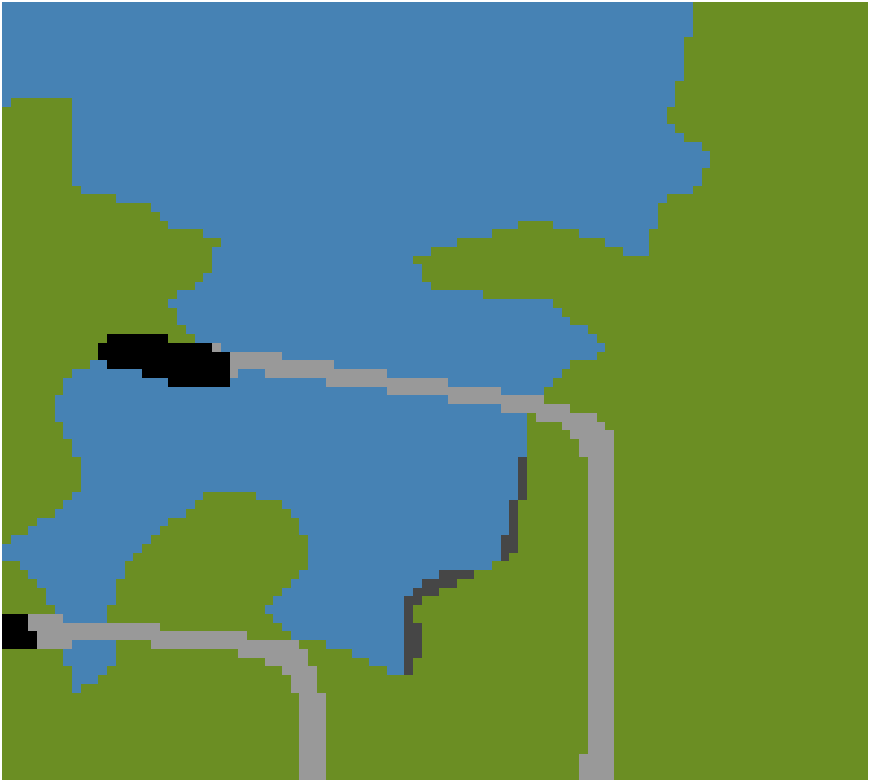}
    \end{subfigure}\hfill
    \begin{subfigure}[b]{0.33\linewidth}
    \includegraphics[width=\linewidth, height=\linewidth]{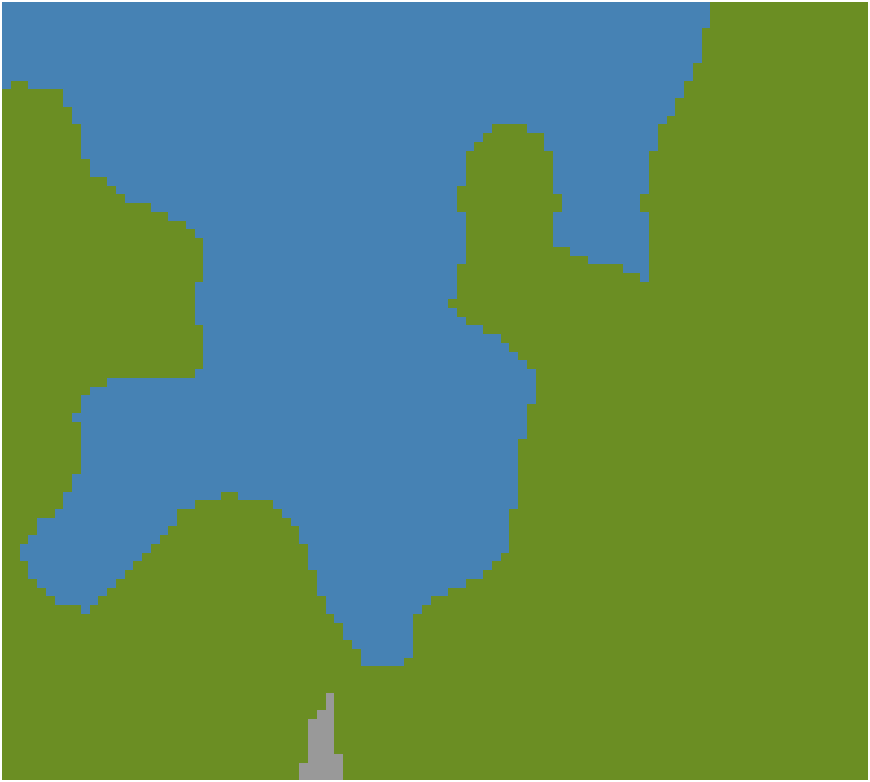}
    \end{subfigure}\hfill
    \begin{subfigure}[b]{0.33\linewidth}
    \includegraphics[width=\linewidth, height=\linewidth]{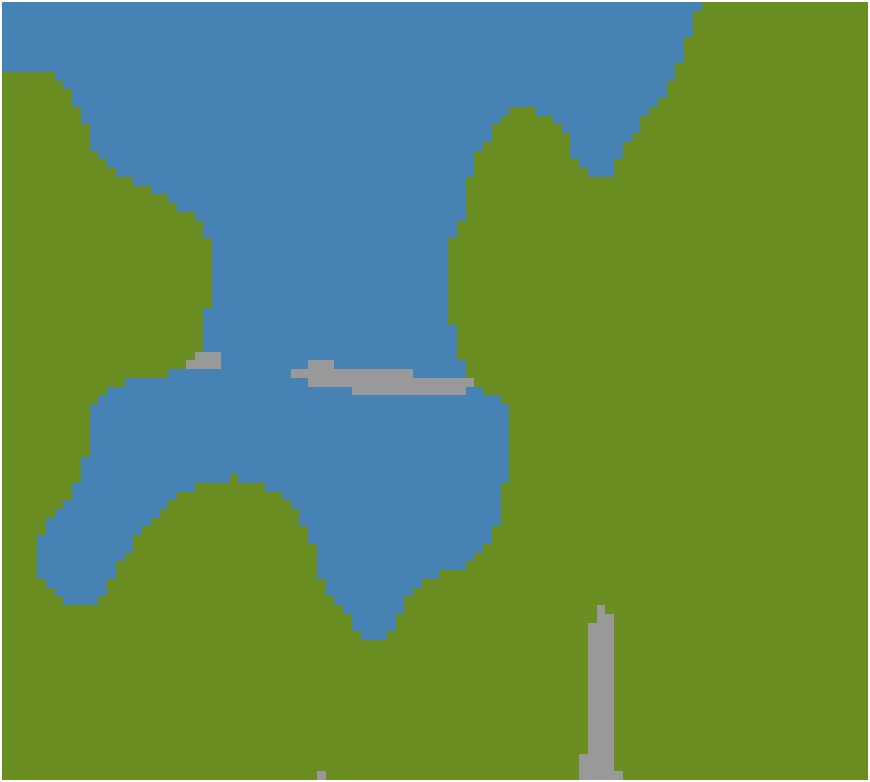}
    \end{subfigure}\\
    \begin{subfigure}[b]{0.33\linewidth}
    \includegraphics[width=\linewidth, height=\linewidth]{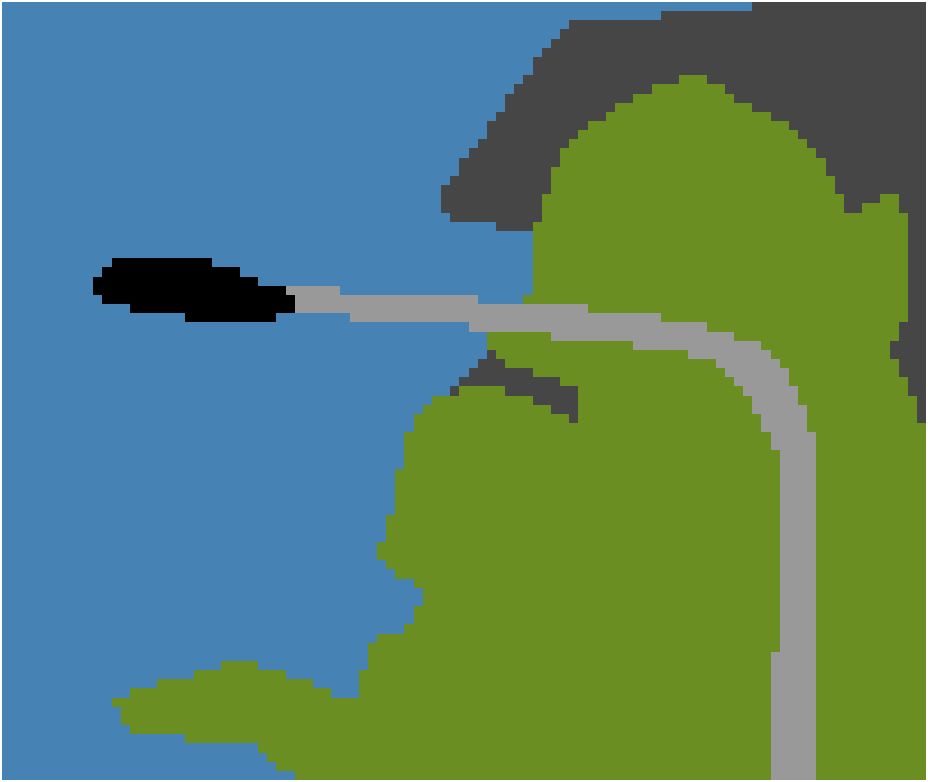}
    \end{subfigure}\hfill
    \begin{subfigure}[b]{0.33\linewidth}
    \includegraphics[width=\linewidth, height=\linewidth]{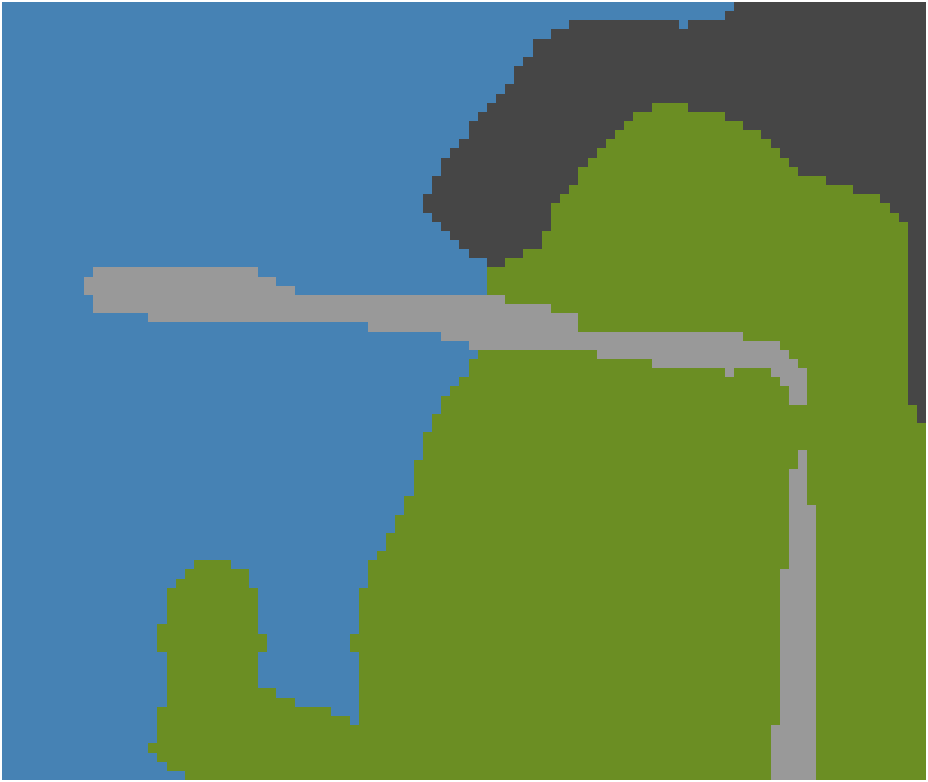}
    \end{subfigure}\hfill
    \begin{subfigure}[b]{0.33\linewidth}
    \includegraphics[width=\linewidth, height=\linewidth]{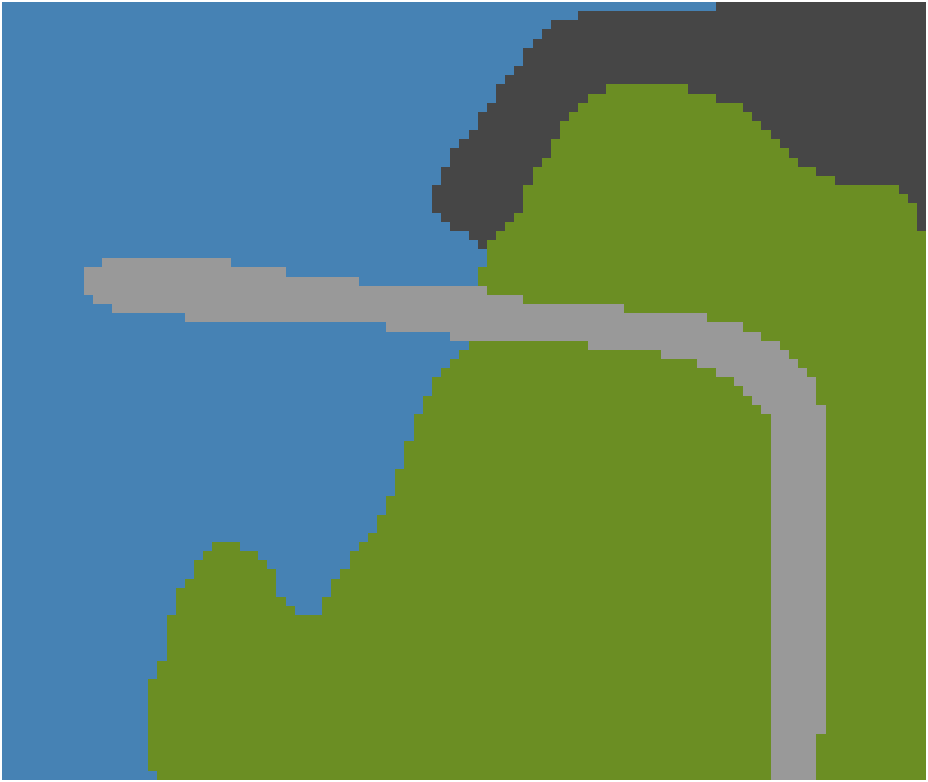}
    \end{subfigure}\\
    \begin{subfigure}[b]{0.33\linewidth}
    \includegraphics[width=\linewidth, height=\linewidth]{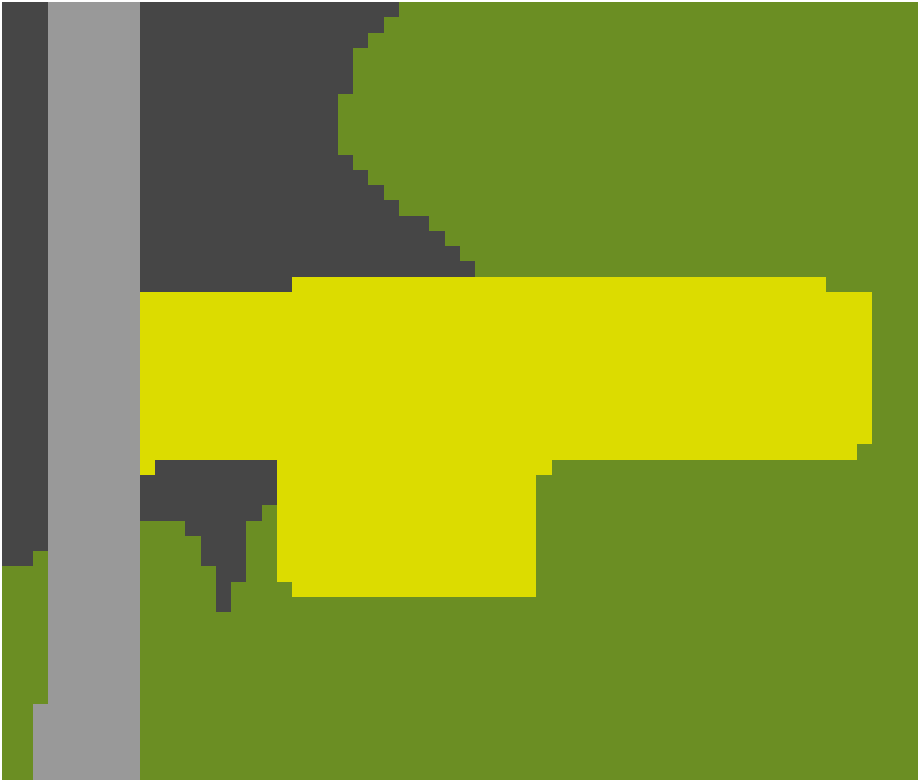}
    \end{subfigure}\hfill
    \begin{subfigure}[b]{0.33\linewidth}
    \includegraphics[width=\linewidth, height=\linewidth]{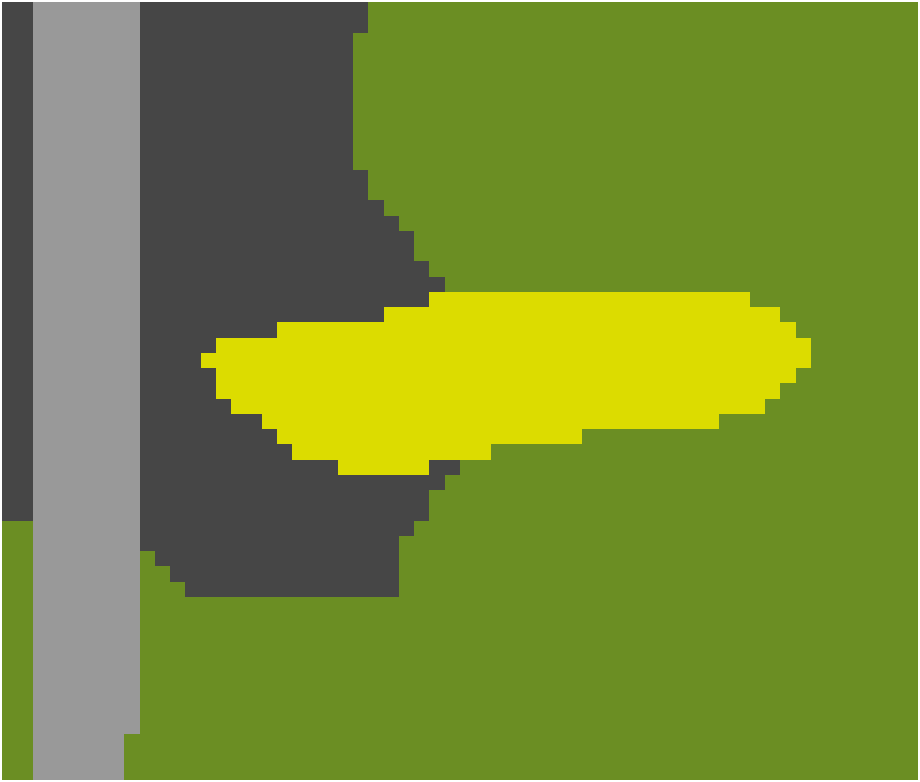}
    \end{subfigure}\hfill
    \begin{subfigure}[b]{0.33\linewidth}
    \includegraphics[width=\linewidth, height=\linewidth]{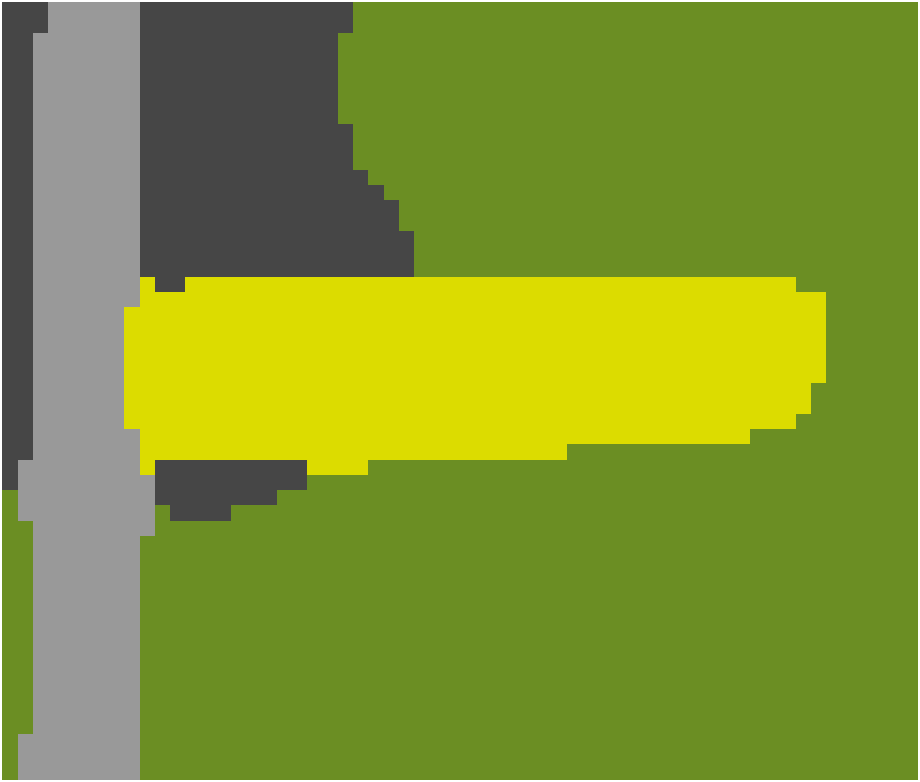}
    \end{subfigure}\hfill
    \caption{Detailed visual comparison of the baseline/reported HRNetV2 \cite{YuanCW20} and our HRNetV2 model trained on the Cityscapes validation set. First row represents the colour and label images respectively. The second row presents the reported HRNetV2 results and our segmentation maps respectively. The third to fifth row illustrates crops of the ground truth, reported HRNetV2 results, and our predictions.}
    \label{fig:tab4Crop}
    \vspace{-5mm}
\end{figure}

Compared to the selected state-of-the-art segmentation methods, our approach is capable of presenting similar/slightly superior performance with substantially lower memory requirements. Note that we employ and compare CNN- and transformer-based methods presenting similar improvements. Furthermore, our proposal has lower complexity\textemdash in terms of implementation effort and setup.  In contrast, most compared methods rely on additional losses to improve performance. As a result, our proposal stands out as a more efficient and effective solution. Figure \ref{fig:tab4Crop} visually compares the segmentation maps of the best-reported method employing HRNetV2 \cite{YuanCW20}  with our best model. Note how our model is less prone to generate semantic gaps in structures and can segment better the finer details.

%\subsection{The advantages of the proposal are transferred to other datasets}
\subsection{The proposal's benefits extend to other datasets}
\begin{table}[tbp]
    \centering
    %\resizebox{.8\linewidth}{!}{%
    \begin{tabular}{l l l l c}
        & \multicolumn{2}{c}{Training Hardware}&\\
         Venue/Journal & \#GPUs & GB/GPU & BS& mIoU \\\toprule
         CVPR 2023\cite{Yu_2023_CVPR}& 8&24& 128& 38.9\\
         TIP 2020\cite{8954873}&2*&24*&10&44.9\\%DeeplabV3+\cite{deeplabv3plus2018} & 50& 24& 44.1\\
         CVPR 2022\cite{li2022deep}& 8&32&16& 44.5\\
         %PSPNet & - &  44.94\\
         %SegFormer\cite{Xie2021SegFormerSA} &-& 45.13\\
         ECCV 2020\cite{YuanCW20} & 4&32 &16& 45.5\\
         TIP 2023\cite{9745313}&4*&16*&16&45.9\\
         CVPR 2023\cite{Chen_2023_CVPR}&  8&48&32&46.3\\
         NeurIPS 2021\cite{Xie2021SegFormerSA}&  8&32&16&46.5\\ \cdashline{1-4}
         Ours & 1& 24 & 4&48.3\\\bottomrule
    \end{tabular}%}
    \caption{Performance and training hardware comparison of state-of-the-art semantic segmentation methods on the ADE20K validation set. All compared methods are trained with an input resolution of $512\times512$. When a parameter is not explicitly reported `*', we provide an estimation given the known parameters. BS stands for training batch size.}% GPU stands for the amount of GPU memory required for training. The requirement of multiple GPUs by some methods is reported by: Memory of each GPU $\times$ Number of GPUs.}
    \label{tab:ADE20K}
    
\end{table}
Unlike the Cityscapes dataset, alternative semantic segmentation datasets (e.g., Mapilliary and ADE20K) are composed of images with variable resolutions. 

\begin{figure}[htp]
    \centering
    \begin{subfigure}[b]{0.25\linewidth}
    \includegraphics[width=\textwidth]{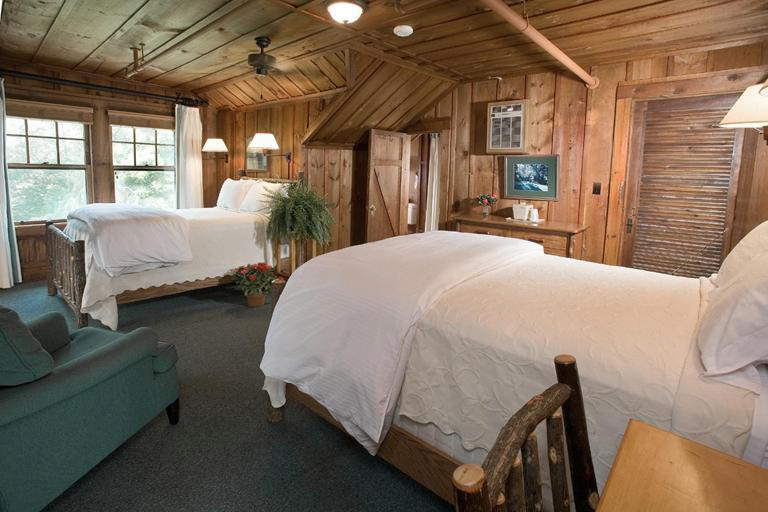}
    \end{subfigure}\hfill
    \begin{subfigure}[b]{0.25\linewidth}
    \includegraphics[width=\textwidth]{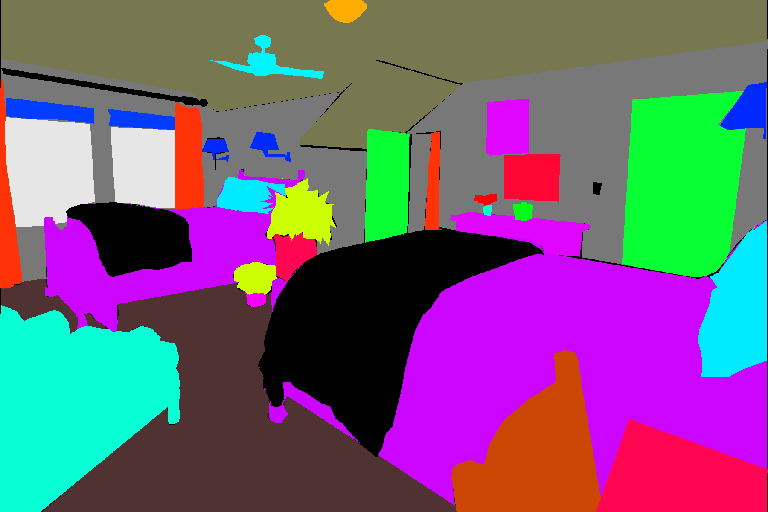}
    \end{subfigure}\hfill
    \begin{subfigure}[b]{0.25\linewidth}
    \includegraphics[width=\textwidth]{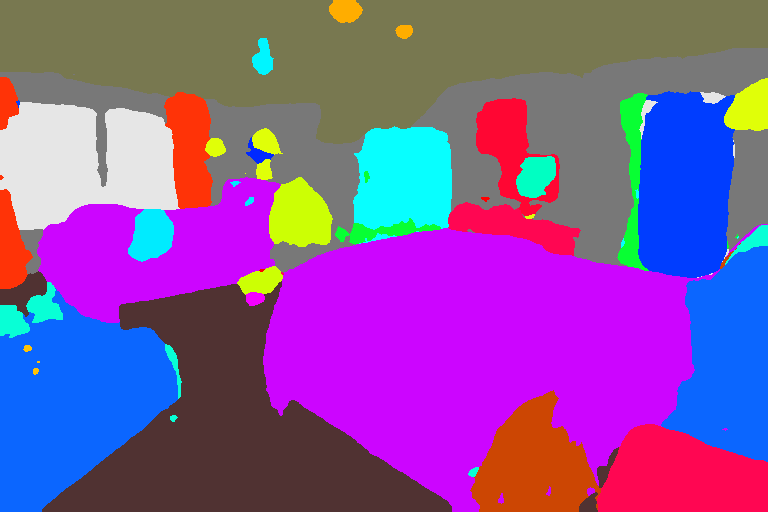}
    \end{subfigure}\hfill
    \begin{subfigure}[b]{0.25\linewidth}
    \includegraphics[width=\textwidth]{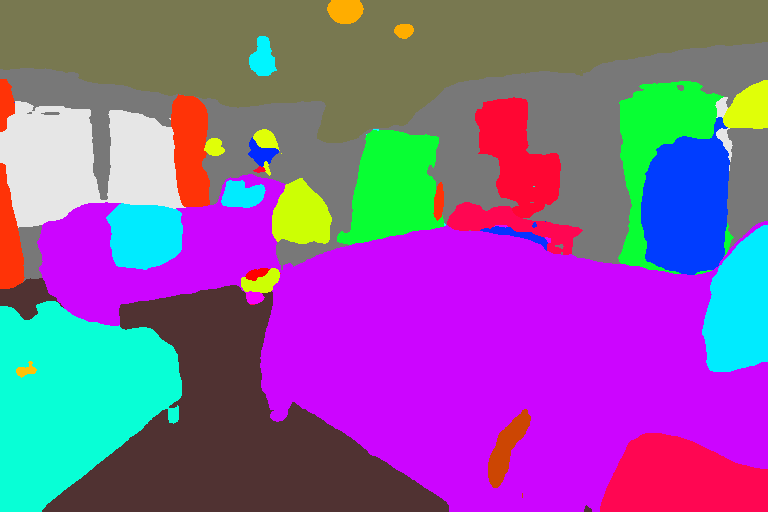}
    \end{subfigure}
     \begin{subfigure}[b]{0.25\linewidth}
    \includegraphics[width=\textwidth]{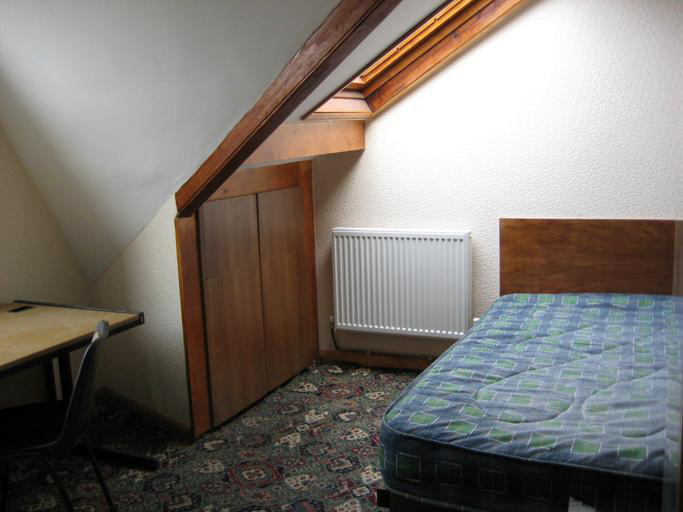}
    \end{subfigure}\hfill
    \begin{subfigure}[b]{0.25\linewidth}
    \includegraphics[width=\textwidth]{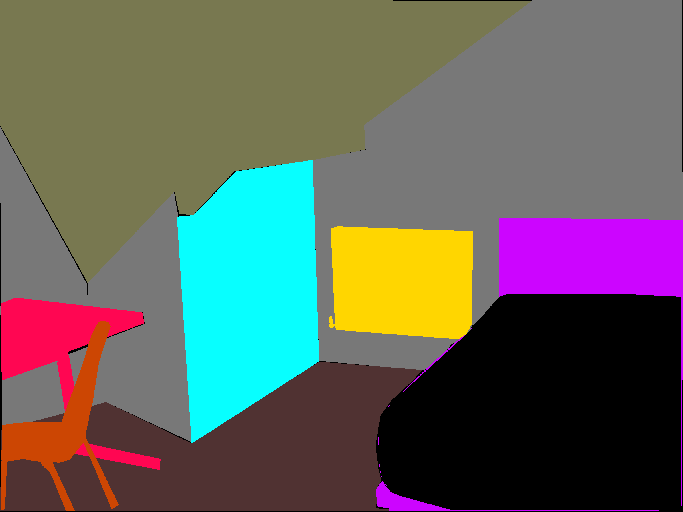}
    \end{subfigure}\hfill
     \begin{subfigure}[b]{0.25\linewidth}
    \includegraphics[width=\textwidth]{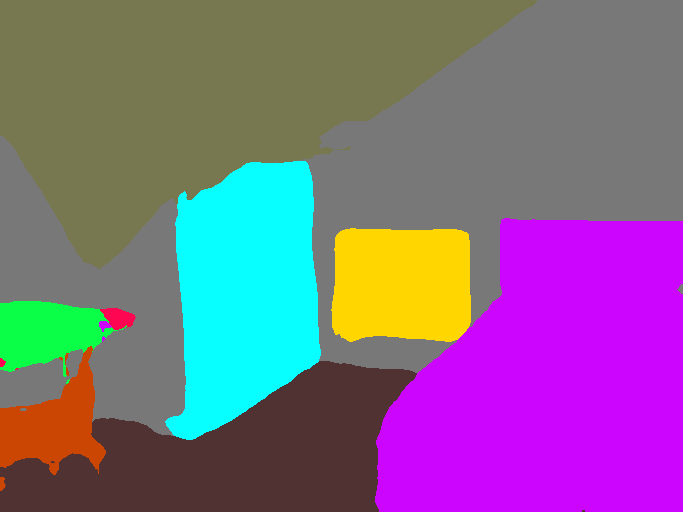}
    \end{subfigure}\hfill
    \begin{subfigure}[b]{0.25\linewidth}
    \includegraphics[width=\textwidth]{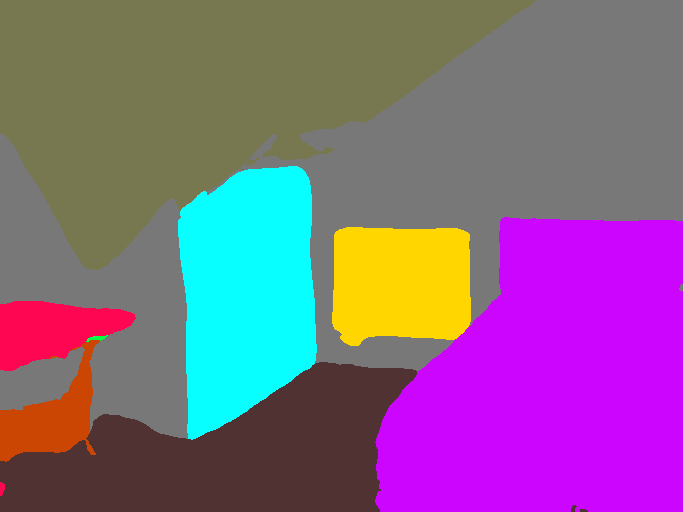}
    \end{subfigure}\hfill
    \caption{Visual comparison of segmentation maps from two validation images of the ADE20K generated by a SegFormer model trained with and without soft-labels. First and second columns present the colour and label images respectively. Third and fourth columns present the segmentation maps of the models trained without and with soft-labels. We employ the ADE20K colour code for visualization \cite{zhou2019semantic}.}
    \label{fig:ADE20k}
\end{figure}

For the ADE20K dataset, we here explore the effect of using the proposed sampling strategy in a more general sense than just to reduce input resolution, also performing up-sampling when the selected crop size is larger than the image. Including our strategy in the training of the architecture SegFormer \cite{Xie2021SegFormerSA}, notably benefits the test performance (see Table \ref{tab:ADE20K}). Figure \ref{fig:ADE20k} presents a visual comparison of segmentation maps of a SegFormer \cite{Xie2021SegFormerSA} trained with and without soft-labels. 

\begin{table}[tbp]
    \centering
    %\resizebox{.8\linewidth}{!}{%
    \begin{tabular}{l l l l c}
        & \multicolumn{2}{c}{Training Hardware}&\\
         Venue/Journal & \#GPUs & GB/GPU & BS& mIoU \\\toprule
%         PSPNet & - &  44.4\\
         IJCV 2023 \cite{Li2022SFNetFA} & 8& 11&16& 45.8\\
         ECCV 2018\cite{deeplabv3plus2018} &50&24&32& 47.7\\
         ECCV 2020\cite{YuanCW20} & 4&32 &8& 50.8\\
         NeurIPS 2021\cite{NEURIPS2021_950a4152} &  8&32 &8&53.1\\\cdashline{1-4}
         Ours & 1& 24 &2& 54.4\\\bottomrule
    \end{tabular}%}
    \caption{Performance and training requirements comparison of state-of-the-art semantic segmentation methods on the Mapilliary validation set with an HRNetV2 trained with a crop size of $512\times1024$. BS stands for training batch size.} %GPU stands for the amount of GPU memory required for training. The requirement of multiple GPUs by some methods is reported by: Memory of each GPU $\times$ Number of GPUs}
    \label{tab:Map}
    
\end{table}

\begin{figure}[htp]
    \centering
    \begin{subfigure}[b]{0.25\linewidth}
    \includegraphics[trim= 360 0 0 0, clip, height=.65\linewidth,width=\textwidth]{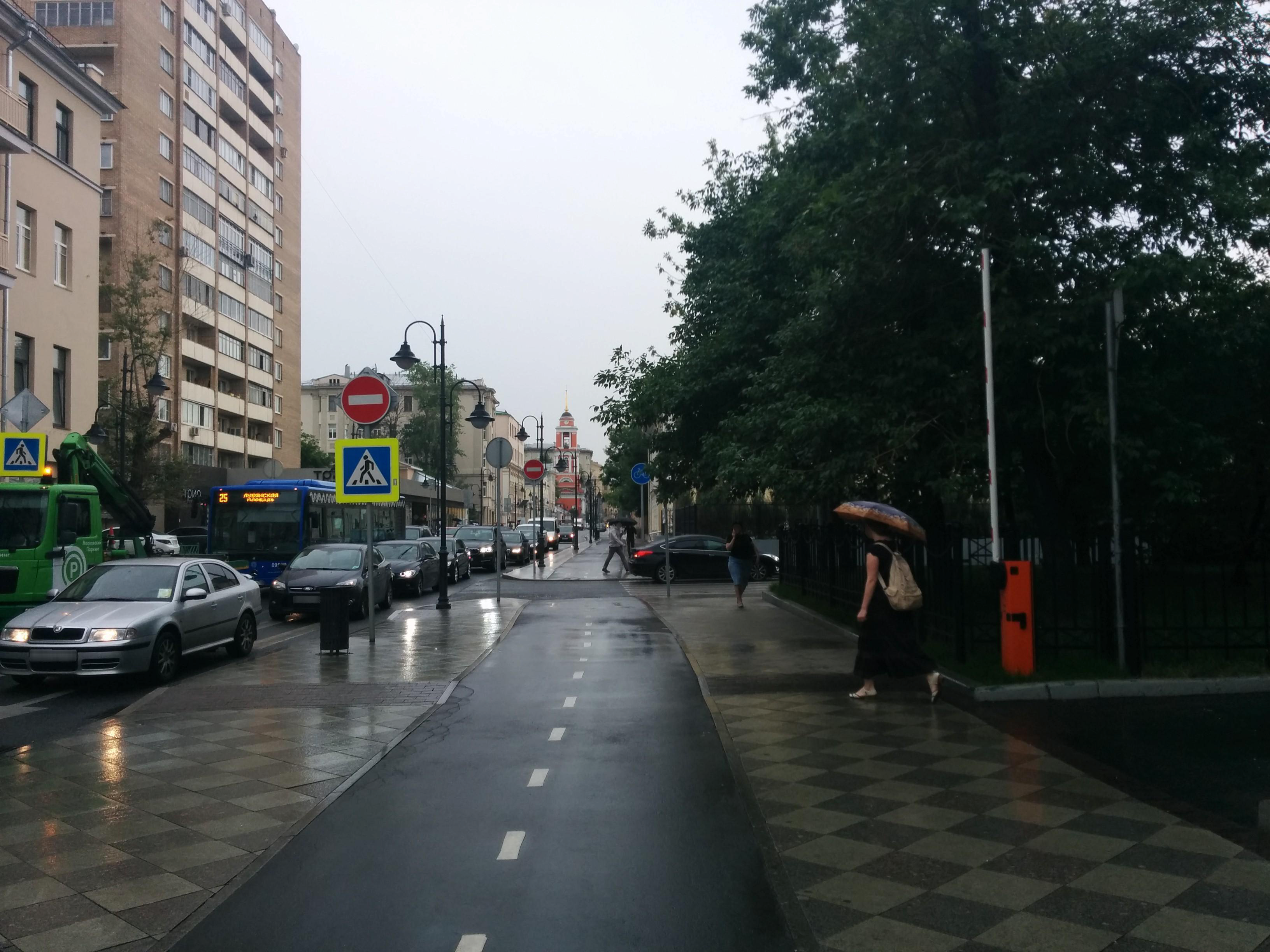}
    \end{subfigure}\hfill
    \begin{subfigure}[b]{0.25\linewidth}
    \includegraphics[trim= 360 0 0 0, clip,height=.65\linewidth,width=\textwidth]{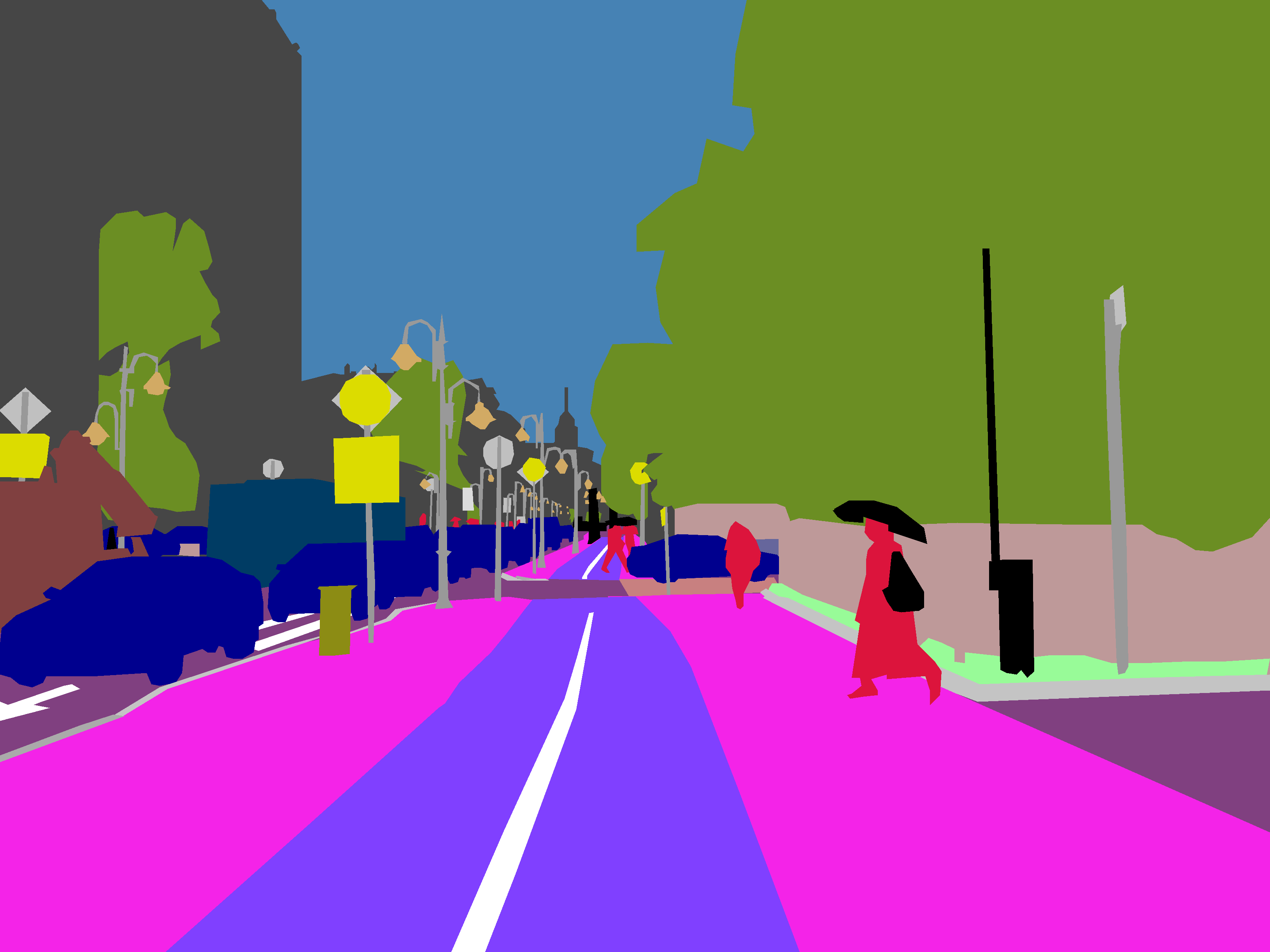}
    \end{subfigure}\hfill
    \begin{subfigure}[b]{0.25\linewidth}
    \includegraphics[trim= 360 0 0 0, clip,height=.65\linewidth,width=\textwidth]{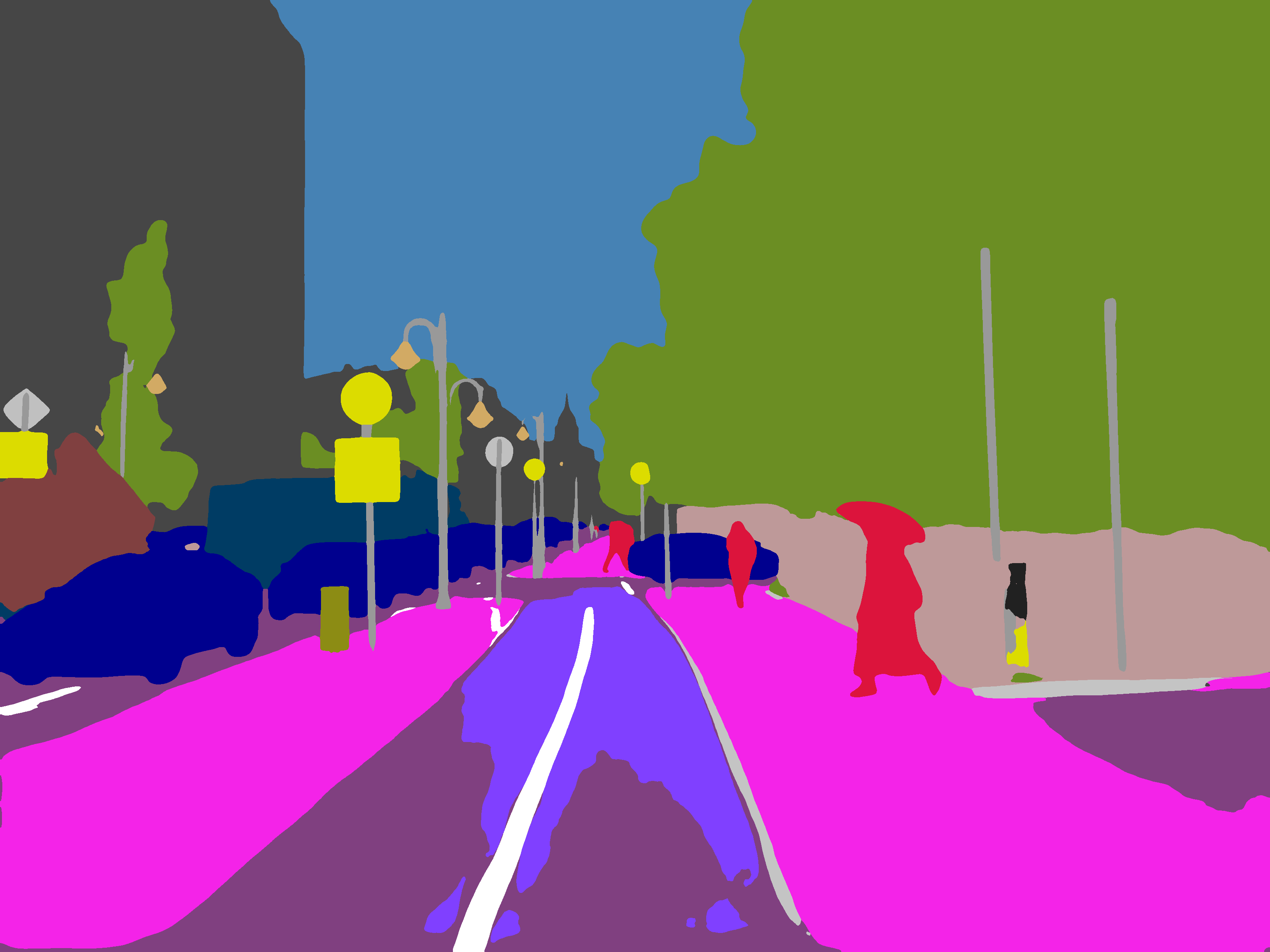}
    \end{subfigure}\hfill
    \begin{subfigure}[b]{0.25\linewidth}
    \includegraphics[trim= 250 0 0 0, clip,height=.65\linewidth,width=\textwidth]{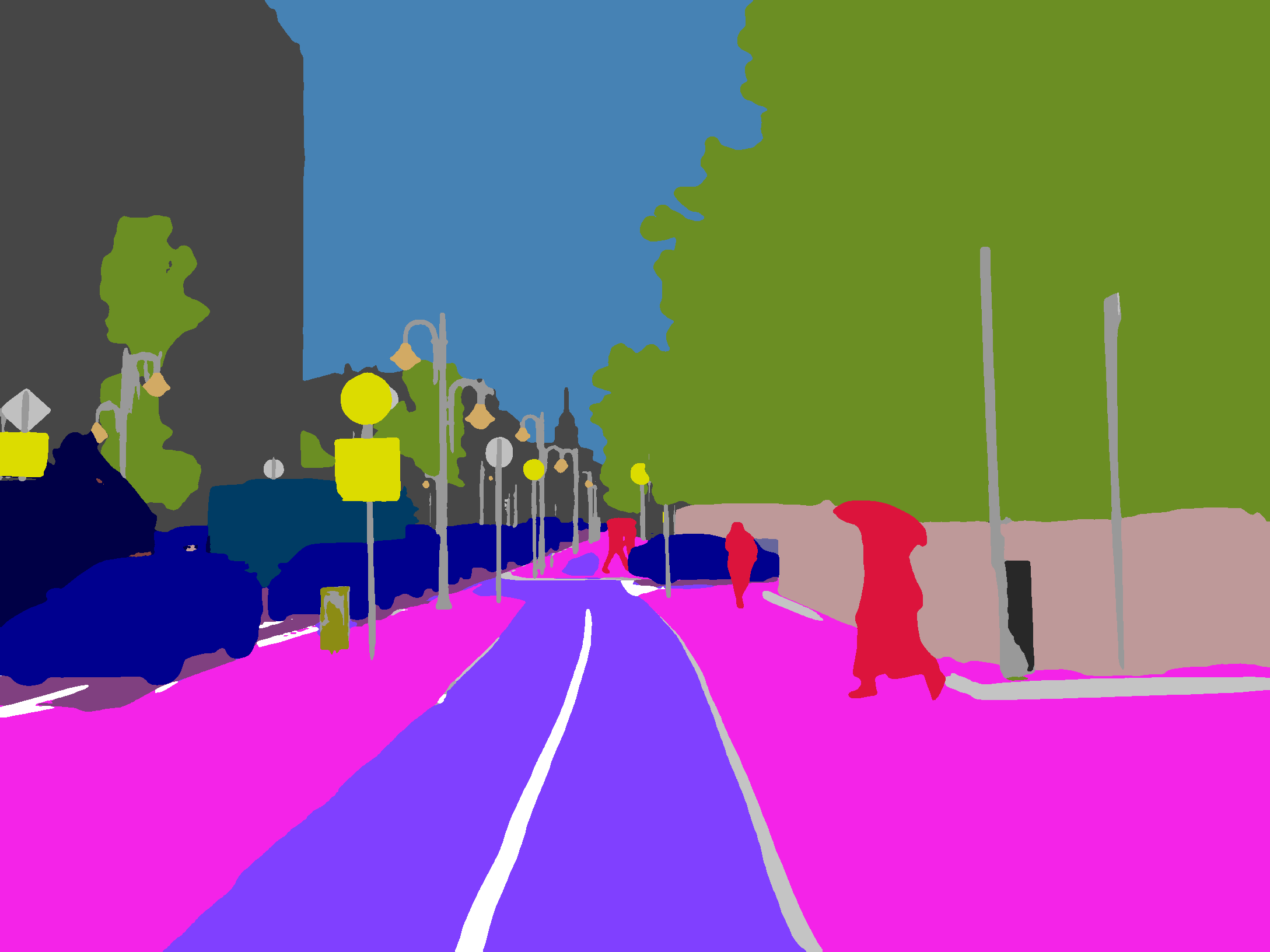}
    \end{subfigure}\\
    \begin{subfigure}[b]{0.25\linewidth}
    \includegraphics[trim=660 1060 1800 820,clip,height=.65\linewidth,width=\linewidth]{Qualitative/Mapillary/matatT_eRHwLZcJhUdE7CQ_input.png}
    \end{subfigure}\hfill
    \begin{subfigure}[b]{0.25\linewidth}
    \includegraphics[trim=660 1060 1800 820,clip,height=.65\linewidth,width=\linewidth,]{Qualitative/Mapillary/matatT_eRHwLZcJhUdE7CQ_gt.png}
    \end{subfigure}\hfill
    \begin{subfigure}[b]{0.25\linewidth}
    \includegraphics[trim=660 1060 1800 820,clip,height=.65\linewidth, width=\linewidth,]{Qualitative/Mapillary/matatT_eRHwLZcJhUdE7CQ_prediction.png}
    \end{subfigure}\hfill
    \begin{subfigure}[b]{0.25\linewidth}
    \includegraphics[trim=440 700 1200 550,clip,height=.65\linewidth, width=\linewidth,]{Qualitative/Mapillary/matatT_eRHwLZcJhUdE7CQ_prediction2.png}
    \end{subfigure}
    \caption{Detailed visual comparison of segmentation maps from a validation image of the Mapillary generated by a HRNetV2 model trained with and without soft-labels. First row presents the full-size images and the second row a detailed crop. The rows are divided into columns: First and second columns present the colour and label images. Third and Fourth columns present the segmentation maps from the model trained without and with soft-labels respectively. We employ the Mapilliary colour code for visualization \cite{MVD2017}.}
    \label{fig:MAP}
\end{figure}

Moreover, in Table \ref{tab:Map} we report performance results by the same setup used for our best-performing Cityscapes model with the HRNetV2 architecture when trained and evaluated on the Mapilliary dataset. Comparing this model with a reference one that uses the same architecture \cite{YuanCW20}, our down-sampling strategy provides a 7\% global increase in performance, requiring significantly less GPU memory. Figure \ref{fig:MAP} compares both models' segmentation. Both Figures \ref{fig:ADE20k} and \ref{fig:MAP} showcase the increased segmentation capabilities of edges and thin and small semantic structures of our model.

% \paragraph{Panoptic Segmentation}
% Our paired down-sampling can be easily transferred to related pixel-wise tasks such as panoptic segmentation. Table \ref{tab:Panoptic} compares the performance of employing our paired down-sampling and the reported  down-sampling with a PanopticDL model with different backbones \cite{cheng2020panoptic}. The results demonstrate consistent performance improvements across all metrics, underscoring the versatility of our proposal across various datasets and pixel-wise tasks.
% \begin{table}[htp]
%     \centering
%     \resizebox{1\linewidth}{!}{%
%     \begin{tabular}{c c c c c c}
%     %\toprule
%          Backbone & mIoU & AP & PQ&  SQ& RQ\\\hline
%          ResNet50\cite{cheng2020panoptic} & 78.6& 26.9& 59.8& 80.0& 73.5\\
%          ResNet50-Ours& 80.2& 29.9& 62.4& 80.7& 76.2\\\hdashline
%          ResNet101\cite{cheng2020panoptic}& 78.4&27.7&60.3&80.8&73.6\\
%          ResNet101-Ours&80.0&32.6&61.5&81.3&74.6\\\hline
%     \end{tabular}}
%     \vspace{-2mm}
%     \caption{Performance results for PanopticDL \cite{cheng2020panoptic} trained on the Cityscapes dataset using the proposed paired down-sampling with soft-labels (full resolution, $1024\times 2048$)}
%     \label{tab:Panoptic}
% \end{table}

\section{Conclusion}
\label{sec:CO}
In this paper, we propose a novel and effective approach for training semantic segmentation models under different resolutions, which can be easily included into any existing framework with minor modifications as shown in the experiments. By using a soft-label encoding of the labels after down-sampling, we manage to pair the down-sampling strategies used for the colour and label images, aligning their information and precluding the learning of discrepant information during the training of semantic segmentation models. Extensive experimentation is conducted to validate the effectiveness of our proposal. Experimental results suggest that the use of this strategy culminates in the learning of models that surpass by over a 88\% their respective baselines using the same architecture, data, and setup but trained adopting the default Nearest Neighbour down-sampling. Furthermore, in the explored setups, the proposed approach can be trained using one GPU, generating models whose performance is better than the state-of-the-art for all explored input resolutions. We believe our proposal has the potential to expand the effective validation of new research conducted at budget-constrained research facilities for semantic segmentation.

\FloatBarrier

\bibliographystyle{IEEEtran}
\bibliography{egbib}

\begin{thebibliography}{10}
\providecommand{\url}[1]{#1}
\csname url@samestyle\endcsname
\providecommand{\newblock}{\relax}
\providecommand{\bibinfo}[2]{#2}
\providecommand{\BIBentrySTDinterwordspacing}{\spaceskip=0pt\relax}
\providecommand{\BIBentryALTinterwordstretchfactor}{4}
\providecommand{\BIBentryALTinterwordspacing}{\spaceskip=\fontdimen2\font plus
\BIBentryALTinterwordstretchfactor\fontdimen3\font minus
  \fontdimen4\font\relax}
\providecommand{\BIBforeignlanguage}[2]{{%
\expandafter\ifx\csname l@#1\endcsname\relax
\typeout{** WARNING: IEEEtran.bst: No hyphenation pattern has been}%
\typeout{** loaded for the language `#1'. Using the pattern for}%
\typeout{** the default language instead.}%
\else
\language=\csname l@#1\endcsname
\fi
#2}}
\providecommand{\BIBdecl}{\relax}
\BIBdecl

\bibitem{Cordts2016Cityscapes}
M.~Cordts, M.~Omran, S.~Ramos, T.~Rehfeld, M.~Enzweiler, R.~Benenson,
  U.~Franke, S.~Roth, and B.~Schiele, ``The cityscapes dataset for semantic
  urban scene understanding,'' in \emph{IEEE Conf. Comput. Vis. Pattern
  Recognit. (CVPR)}, 2016, pp. 3212--3223.

\bibitem{9126262}
X.~Ren, S.~Ahmad, L.~Zhang, L.~Xiang, D.~Nie, F.~Yang, Q.~Wang, and D.~Shen,
  ``Task decomposition and synchronization for semantic biomedical image
  segmentation,'' \emph{IEEE Transactions on Image Processing}, vol.~29, pp.
  7497--7510, 2020.

\bibitem{NVIDIA_semseg}
A.~Tao, K.~Sapra, and B.~Catanzaro, ``Hierarchical multi-scale attention for
  semantic segmentation,'' \emph{CoRR}, vol. abs/2005.10821, 2020.

\bibitem{2019_zhao_rmi}
S.~Zhao, Y.~Wang, Z.~Yang, and D.~Cai, ``Region mutual information loss for
  semantic segmentation,'' \emph{Advances in Neural Information Processing
  Systems (NeurIPS)}, p. 11117–11127, 2019.

\bibitem{Wang_2023_CVPR}
Y.~Wang, J.~Fei, H.~Wang, W.~Li, T.~Bao, L.~Wu, R.~Zhao, and Y.~Shen,
  ``Balancing logit variation for long-tailed semantic segmentation,'' in
  \emph{IEEE Conf. Comput. Vis. Pattern Recognit. (CVPR)}, 2023, pp.
  19\,561--19\,573.

\bibitem{8803154}
Y.~Wang, Q.~Zhou, J.~Liu, J.~Xiong, G.~Gao, X.~Wu, and L.~J. Latecki, ``Lednet:
  A lightweight encoder-decoder network for real-time semantic segmentation,''
  in \emph{IEEE Int. Conf. Image Processing (ICIP)}, 2019, pp. 1860--1864.

\bibitem{zhou2019semantic}
B.~Zhou, H.~Zhao, X.~Puig, T.~Xiao, S.~Fidler, A.~Barriuso, and A.~Torralba,
  ``Semantic understanding of scenes through the ade20k dataset,'' \emph{Int.
  Journal of Computer Vision}, vol. 127, pp. 302--321, 2019.

\bibitem{liang2019winter}
C.~Liang, J.~Ge, W.~Zhang, K.~Gui, F.~A. Cheikh, and L.~Ye, ``Winter road
  surface status recognition using deep semantic segmentation network,'' in
  \emph{Int. Workshop on Atmospheric Icing of Structures (IWAIS)}, 2019, pp.
  23--28.

\bibitem{li2022deep}
L.~Li, T.~Zhou, W.~Wang, J.~Li, and Y.~Yang, ``Deep hierarchical semantic
  segmentation,'' in \emph{IEEE Conf. Comput. Vis. Pattern Recognit. (CVPR)},
  2022, pp. 1246--1257.

\bibitem{deeplabv3plus2018}
L.-C. Chen, Y.~Zhu, G.~Papandreou, F.~Schroff, and H.~Adam, ``Encoder-decoder
  with atrous separable convolution for semantic image segmentation,'' in
  \emph{IEEE Eur. Conf. Comput. Vis. (ECCV)}, 2018, p. 833–851.

\bibitem{Li2019DFANetDF}
H.~Li, P.~Xiong, H.~Fan, and J.~Sun, ``Dfanet: Deep feature aggregation for
  real-time semantic segmentation,'' \emph{IEEE Conf. Comput. Vis. Pattern
  Recognit. (CVPR)}, pp. 9514--9523, 2019.

\bibitem{Lin2020GraphGuidedAS}
P.~Lin, P.~Sun, G.~Cheng, S.~Xie, X.~Li, and J.~Shi, ``Graph-guided
  architecture search for real-time semantic segmentation,'' \emph{IEEE Conf.
  Comput. Vis. Pattern Recognit. (CVPR)}, pp. 4202--4211, 2020.

\bibitem{mehta2018espnetv2}
S.~Mehta, M.~Rastegari, L.~Shapiro, and H.~Hajishirzi, ``Espnetv2: A
  light-weight, power efficient, and general purpose convolutional neural
  network,'' in \emph{IEEE Conf. Comput. Vis. Pattern Recognit. (CVPR)}, 2019,
  pp. 9182--9192.

\bibitem{Yu_2018_ECCV}
C.~Yu, J.~Wang, C.~Peng, C.~Gao, G.~Yu, and N.~Sang, ``Bisenet: Bilateral
  segmentation network for real-time semantic segmentation,'' in \emph{IEEE
  Eur. Conf. Comput. Vis. (ECCV)}, 2018, pp. 334,349.

\bibitem{hong2021deep}
Y.~Hong, H.~Pan, W.~Sun, and Y.~Jia, ``Deep dual-resolution networks for
  real-time and accurate semantic segmentation of road scenes,'' \emph{IEEE
  Transactions on Intelligent Transportation Systems}, vol.~24, no.~3, pp.
  3448--3460, 2022.

\bibitem{xu2022pidnet}
J.~Xu, Z.~Xiong, and S.~P. Bhattacharyya, ``{PID}net: A real-time semantic
  segmentation network inspired from pid controller,'' \emph{IEEE Conf. Comput.
  Vis. Pattern Recognit. (CVPR)}, pp. 19\,529--19\,539, 2023.

\bibitem{zhou2022rethinking}
T.~Zhou, W.~Wang, E.~Konukoglu, and L.~Van~Gool, ``Rethinking semantic
  segmentation: A prototype view,'' in \emph{IEEE Conf. Comput. Vis. Pattern
  Recognit. (CVPR)}, 2022, pp. 2572--2583.

\bibitem{NN}
O.~Rukundo, ``Evaluation of rounding functions in nearest neighbor
  interpolation,'' \emph{Int. Journal of Computational Methods}, vol.~18,
  no.~8, p. 2150024, 2021.

\bibitem{9052469}
J.~Wang, K.~Sun, T.~Cheng, B.~Jiang, C.~Deng, Y.~Zhao, D.~Liu, Y.~Mu, M.~Tan,
  X.~Wang, W.~Liu, and B.~Xiao, ``Deep high-resolution representation learning
  for visual recognition,'' \emph{IEEE Transactions on Pattern Analysis and
  Machine Intelligence}, vol.~43, no.~10, pp. 3349--3364, 2019.

\bibitem{7298965}
J.~Long, E.~Shelhamer, and T.~Darrell, ``Fully convolutional networks for
  semantic segmentation,'' in \emph{IEEE Conf. Comput. Vis. Pattern Recognit.
  (CVPR)}, 2015, pp. 3431--3440.

\bibitem{Yang2021RealtimeSS}
M.~Y. Yang, S.~Kumaar, Y.~Lyu, and F.~Nex, ``Real-time semantic segmentation
  with context aggregation network,'' \emph{Journal of Photogrammetry and
  Remote Sensing (ISPRS)}, pp. 124--134, 2021.

\bibitem{borse2021inverseform}
S.~Borse, Y.~Wang, Y.~Zhang, and F.~Porikli, ``Inverseform: {A} loss function
  for structured boundary-aware segmentation,'' in \emph{IEEE Conf. Comput.
  Vis. Pattern Recognit. (CVPR)}, 2021, pp. 5897--5907.

\bibitem{YuanCW20}
Y.~Yuan, X.~Chen, and J.~Wang, ``Object-contextual representations for semantic
  segmentation,'' in \emph{IEEE Eur. Conf. Comput. Vis. (ECCV)}, 2020, pp.
  173--190.

\bibitem{10173725}
D.~Wu, Z.~Guo, A.~Li, C.~Yu, C.~Gao, and N.~Sang, ``Conditional boundary loss
  for semantic segmentation,'' \emph{IEEE Transactions on Image Processing},
  vol.~32, pp. 3717--3731, 2023.

\bibitem{m_Huynh-etal-CVPR21}
C.~Huynh, A.~Tran, K.~Luu, and M.~Hoai, ``Progressive semantic segmentation,''
  in \emph{IEEE Conf. Comput. Vis. Pattern Recognit. (CVPR)}, 2021, pp.
  16\,750--16\,759.

\bibitem{10112629}
Z.~Liang, X.~Dai, Y.~Wu, X.~Jin, and J.~Shen, ``Multi-granularity context
  network for efficient video semantic segmentation,'' \emph{IEEE Transactions
  on Image Processing}, vol.~32, pp. 3163--3175, 2023.

\bibitem{9008795}
D.~Marin, Z.~He, P.~Vajda, P.~Chatterjee, S.~Tsai, F.~Yang, and Y.~Boykov,
  ``Efficient segmentation: Learning downsampling near semantic boundaries,''
  in \emph{IEEE Int. Conf. Comput. Vis. (ICCV)}, 2019, pp. 2131--2141.

\bibitem{10.5555/3454287.3454709}
R.~M\"{u}ller, S.~Kornblith, and G.~Hinton, ``When does label smoothing help?''
  in \emph{Advances in Neural Information Processing Systems (NeurIPS)}, 2019,
  p. 4694–4703.

\bibitem{10.1007/978-3-319-54187-7_14}
Z.~Tan, S.~Zhou, J.~Wan, Z.~Lei, and S.~Z. Li, ``Age estimation based on a
  single network with soft softmax of aging modeling,'' in \emph{Asian Conf.
  Comput. Vis.(ACCV)}, 2017, pp. 203--216.

\bibitem{Diaz_2019_CVPR}
R.~Diaz and A.~Marathe, ``Soft labels for ordinal regression,'' in \emph{IEEE
  Conf. Comput. Vis. Pattern Recognit. (CVPR)}, 2019, pp. 4733--4742.

\bibitem{Han2013/03}
D.~Han, ``Comparison of commonly used image interpolation methods,'' in
  \emph{Int. Conf. Comput. Science and Electronics Engineering (ICCSEE)}, 2013,
  pp. 1556--1559.

\bibitem{OHE}
K.~Potdar, T.~Pardawala, and C.~Pai, ``A comparative study of categorical
  variable encoding techniques for neural network classifiers,'' \emph{Int.
  Journal of Computer Applications}, vol. 175, no.~4, pp. 7--9, 2017.

\bibitem{kullback1951information}
S.~Kullback and R.~A. Leibler, ``On information and sufficiency,'' \emph{The
  annals of mathematical statistics}, vol.~22, no.~1, pp. 79--86, 1951.

\bibitem{DBLP:conf/iccv/TzengHDS15}
E.~Tzeng, J.~Hoffman, T.~Darrell, and K.~Saenko, ``Simultaneous deep transfer
  across domains and tasks,'' in \emph{IEEE Int. Conf. Comput. Vis. (ICCV)},
  2015, pp. 4068--4076.

\bibitem{Wang2020}
H.~Wang, T.~Shen, W.~Zhang, L.~Duan, and T.~Mei, ``Classes matter: A
  fine-grained adversarial approach to cross-domain semantic segmentation,''
  \emph{IEEE Eur. Conf. Comput. Vis. (ECCV)}, pp. 642,659, 2020.

\bibitem{MVD2017}
G.~Neuhold, T.~Ollmann, S.~Rota~Bul\`o, and P.~Kontschieder, ``The mapillary
  vistas dataset for semantic understanding of street scenes,'' in \emph{IEEE
  Int. Conf. Comput. Vis. (ICCV)}, 2017, pp. 5000--5009.

\bibitem{Everingham10thepascal}
M.~Everingham, L.~Van~Gool, C.~K. Williams, J.~Winn, and A.~Zisserman, ``The
  pascal visual object classes (voc) challenge,'' \emph{International Journal
  of Computer Vision}, vol.~88, pp. 303--338, 2010.

\bibitem{rotabulo2017place}
S.~Rota~Bul\`o, L.~Porzi, and P.~Kontschieder, ``In-place activated batchnorm
  for memory-optimized training of dnns,'' in \emph{IEEE Conf. Comput. Vis.
  Pattern Recognit. (CVPR)}, 2018, pp. 5639--5647.

\bibitem{Xie2021SegFormerSA}
E.~Xie, W.~Wang, Z.~Yu, A.~Anandkumar, J.~M. {\'A}lvarez, and P.~Luo,
  ``Segformer: Simple and efficient design for semantic segmentation with
  transformers,'' in \emph{Advances in Neural Information Processing Systems
  (NeurIPS)}, 2021.

\bibitem{Wang_2020_CVPR}
L.~Wang, D.~Li, Y.~Zhu, L.~Tian, and Y.~Shan, ``Dual super-resolution learning
  for semantic segmentation,'' in \emph{IEEE Conf. Comput. Vis. Pattern
  Recognit. (CVPR)}, 2020, pp. 3773--3782.

\bibitem{Aakanksha_2023_CVPR}
Aakanksha and A.~N. Rajagopalan, ``Improving robustness of semantic
  segmentation to motion-blur using class-centric augmentation,'' in \emph{IEEE
  Conf. Comput. Vis. Pattern Recognit. (CVPR)}, 2023, pp. 10\,470--10\,479.

\bibitem{9745313}
X.~He, J.~Liu, W.~Wang, and H.~Lu, ``An efficient sampling-based attention
  network for semantic segmentation,'' \emph{IEEE Transactions on Image
  Processing}, vol.~31, pp. 2850--2863, 2022.

\bibitem{Chen_2023_CVPR}
J.~Chen, J.~Lu, X.~Zhu, and L.~Zhang, ``Generative semantic segmentation,'' in
  \emph{IEEE Conf. Comput. Vis. Pattern Recognit. (CVPR)}, 2023, pp.
  7111--7120.

\bibitem{zhu2023good}
J.~Zhu, Y.~Luo, X.~Zheng, H.~Wang, and L.~Wang, ``A good student is cooperative
  and reliable: Cnn-transformer collaborative learning for semantic
  segmentation,'' \emph{IEEE Int. Conf. Comput. Vis. (ICCV)}, 2023.

\bibitem{lu2023cts}
C.~Lu, D.~{de Geus}, and G.~Dubbelman, ``{Content-aware Token Sharing for
  Efficient Semantic Segmentation with Vision Transformers},'' in \emph{IEEE
  Conf. Comput. Vis. Pattern Recognit. (CVPR)}, 2023, pp. 23\,631--23\,640.

\bibitem{liu2023boosting}
Y.~Liu, C.~Liu, K.~Han, Q.~Tang, and Z.~Qin, ``Boosting semantic segmentation
  from the perspective of explicit class embeddings,'' \emph{IEEE Int. Conf.
  Comput. Vis. (ICCV)}, 2023.

\bibitem{He_2023_CVPR}
H.~He, J.~Cai, Z.~Pan, J.~Liu, J.~Zhang, D.~Tao, and B.~Zhuang, ``Dynamic
  focus-aware positional queries for semantic segmentation,'' in \emph{IEEE
  Conf. Comput. Vis. Pattern Recognit. (CVPR)}, 2023, pp. 11\,299--11\,308.

\bibitem{NEURIPS2021_950a4152}
B.~Cheng, A.~Schwing, and A.~Kirillov, ``Per-pixel classification is not all
  you need for semantic segmentation,'' in \emph{Advances in Neural Information
  Processing Systems (NeurIPS)}, 2021, pp. 17\,864--17\,875.

\bibitem{Shi_2023_CVPR}
H.~Shi, M.~Hayat, and J.~Cai, ``Transformer scale gate for semantic
  segmentation,'' in \emph{IEEE Conf. Comput. Vis. Pattern Recognit. (CVPR)},
  2023, pp. 3051--3060.

\bibitem{Yu_2023_CVPR}
L.~Yu and W.~Xiang, ``X-pruner: explainable pruning for vision transformers,''
  in \emph{IEEE Conf. Comput. Vis. Pattern Recognit. (CVPR)}, 2023, pp.
  24\,355--24\,363.

\bibitem{8954873}
H.~Ding, X.~Jiang, B.~Shuai, A.~Q. Liu, and G.~Wang, ``Semantic segmentation
  with context encoding and multi-path decoding,'' \emph{IEEE Transactions on
  Image Processing}, vol.~29, pp. 3520--3533, 2020.

\bibitem{Li2022SFNetFA}
X.~Li, J.~Zhang, Y.~Yang, G.~Cheng, K.~Yang, Y.~Tong, and D.~Tao, ``Sfnet:
  Faster, accurate, and domain agnostic semantic segmentation via semantic
  flow,'' \emph{International Journal of Computer Vision}, 2023.

\end{thebibliography}
\begin{IEEEbiography}[{\includegraphics[width=1in,height=1.25in,clip,keepaspectratio]{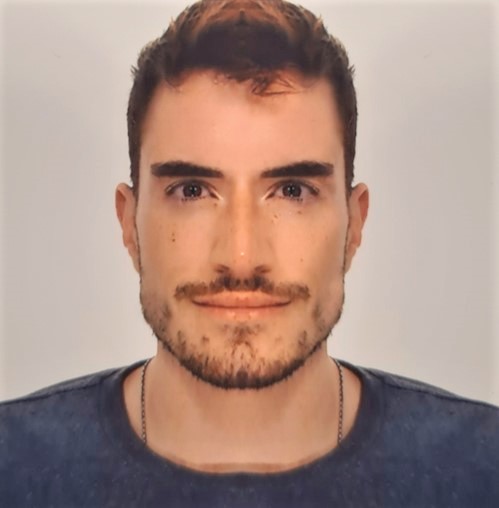}}]{Roberto Alcover-Couso} was born in Madrid Spain in 1996. He received the B.S. and M.S. degrees in computer science from the Autonomous University of Madrid. Currently pursuing a Ph.D. degree in domain adaptation for semantic segmentation. From 2019 to 2021, he worked as a data scientist in Magiquo. Since 2021 he is a reseach intern in the Video Processing and Understanding lab (VPULab) in the Autonomous University of Madrid (UAM). His research focuses on domain adaptation, semantic segmentation, and synthetic data
\end{IEEEbiography}

\begin{IEEEbiography}[{\includegraphics[width=1in,height=1.25in,clip,keepaspectratio]{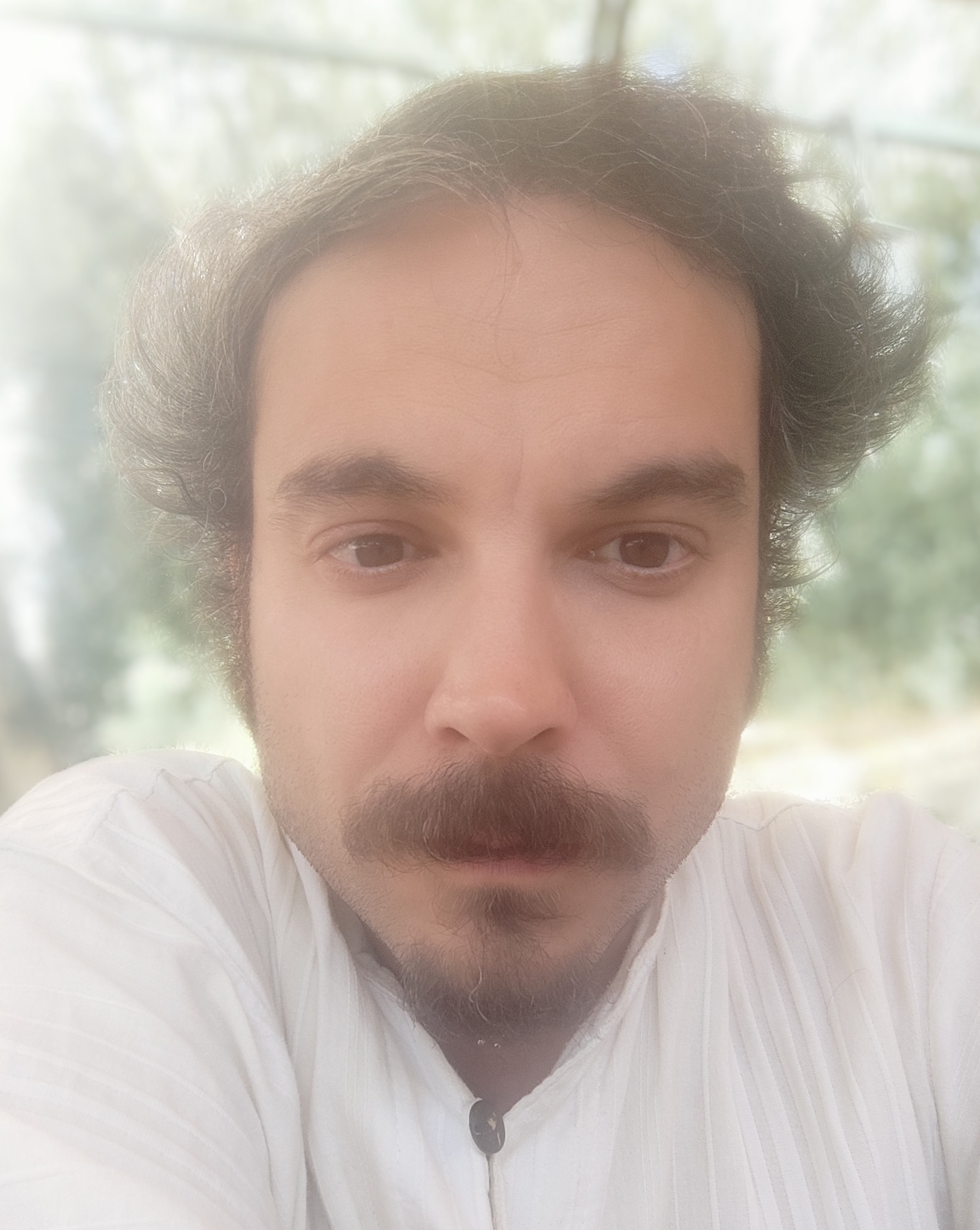}}]{Marcos Escudero-Viñolo} is a researcher and university teacher co-author of more than 25 papers published in
international high-quality peer-reviewed journals and international conferences. His research his funded by national grants and European and
National competitive projects from the public
and the private sectors, including four projects
in which he has been the Principal Investigator. Regarding research topics, he has defined
strategies for driving the analysis of vision signals based on regional and contextual constrains, especially for semantic segmentation, scene
recognition and medical image analysis. He is current research deals with the creation of strategies to provide interpretability, assessment
and profiling gates to the knowledge encoded by deep-learning visual models. These are used to untap the reasons that preclude these models from being reliable, trustworthy, and fair. He is a recurrent reviewer of top-journals (e.g., TCSVT, TIP) and conferences (e.g., CVPR) and has recently accepted to be evaluator of the Spanish AEI.

\end{IEEEbiography}

\begin{IEEEbiography}[{\includegraphics[width=1in,height=1.25in,clip,keepaspectratio]{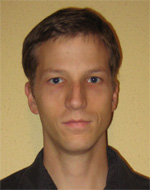}}]{Juan C. SanMiguel} received the Ph.D. degree
in computer science and telecommunication from the Autonomous University of Madrid, Madrid,Spain, in 2011. He was a Postdoctoral Researcher with the Queen Mary University of London, London, U.K., from 2013 to 2014, under a Marie Curie IAPP Fellowship. He is currently an Associate Professor at the Autonomous University of
Madrid and a Researcher with the Video Processing and Understanding Laboratory. He has authored over 55 journals and conference papers. His research interests
include computer vision with a focus on reliability estimation and multi-camera activity understanding for video segmentation and tracking.
\end{IEEEbiography}
\begin{IEEEbiography}
[{\includegraphics[width=1in,height=1.25in,clip,keepaspectratio]{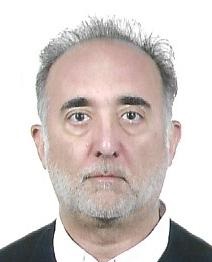}}]{José M. Martinez}    (Senior Member, IEEE) received
the Ph.D. degree in computer science and telecommunication from the Universidad Politécnica de Madrid, Madrid, Spain, in 1998. He is currently a Full Professor with the Escuela Politécnica Superior,
Universidad Autónoma de Madrid, Madrid. He is the author or coauthor of more than 100 papers in international journals and conferences and a coauthor of
the first book about the MPEG-7 standard published in 2002. His professional interests cover different aspects of advanced video surveillance systems and
multimedia information systems. He has acted as an Auditor and a Reviewer for the EC for projects of the frameworks program for research in Information
Society and Technology (IST).
\end{IEEEbiography}
\end{document}